\documentclass[journal]{IEEEtran}

\ifCLASSINFOpdf
  \usepackage[pdftex]{graphicx}
\else
\fi

\usepackage{cite}

\usepackage{ragged2e}
\usepackage{tikz}
\usepackage{tabularx}
\usepackage{adjustbox}
\usepackage{graphicx}
\usepackage{subfigure}
\usepackage{multirow}
\usepackage{colortbl}
\usepackage{enumitem}
\usepackage{comment}
\usepackage{enumitem}

\usepackage{hyperref}
\hypersetup{
}
\begin{document}
\begin{comment}{\noindent\onecolumn{
\noindent\textcopyright 2019 IEEE. Personal use of this material is permitted. Permission from IEEE must be obtained for all other uses, in any current or future media, including reprinting/republishing this material for advertising or promotional purposes, creating new collective works, for resale or redistribution to servers or lists, or reuse of any copyrighted component of this work in other works.}}
%----------------------------------------------------------------%
\newpage
\twocolumn
\end{comment}
% Do not put math or special symbols in the title.
\title{Deep Learning for Visual Tracking: A Comprehensive Survey}
%----------------------------------------------------------------%
\author{Seyed~Mojtaba~Marvasti-Zadeh,~\IEEEmembership{Student Member,~IEEE,}
        Li~Cheng,~\IEEEmembership{Senior Member,~IEEE,}
        Hossein~Ghanei-Yakhdan,
        and~Shohreh~Kasaei,~\IEEEmembership{Senior Member,~IEEE}
\IEEEcompsocitemizethanks{\IEEEcompsocthanksitem S.~M.~Marvasti-Zadeh is with \textit{Digital Image \& Video Processing Lab} (DIVPL), Department of Electrical Engineering, Yazd University, Iran. He is also a member of Vision and Learning Lab, University of Alberta, Canada, and \textit{Image Processing Lab} (IPL), Sharif University of Technology, Iran. E-mail: \href{mailto:mojtaba.marvasti@ualberta.ca}{mojtaba.marvasti@ualberta.ca} 
\IEEEcompsocthanksitem L.~Cheng is with \textit{Vision and Learning Lab}, Department of Electrical and Computer Engineering, University of Alberta, Edmonton, Canada. E-mail: \href{mailto:lcheng5@ualberta.ca}{lcheng5@ualberta.ca}
\IEEEcompsocthanksitem H.~Ghanei-Yakhdan is with \textit{Digital Image \& Video Processing Lab} (DIVPL), Department of Electrical Engineering, Yazd University, Yazd, Iran. E-mail: \href{mailto:hghaneiy@yazd.ac.ir}{hghaneiy@yazd.ac.ir}
\IEEEcompsocthanksitem S.~Kasaei is with \textit{Image Processing Lab} (IPL), Department of Computer Engineering, Sharif University of Technology, Tehran, Iran. E-mail: \href{mailto:kasaei@sharif.edu}{kasaei@sharif.edu}
}% <-this % stops an unwanted space
\thanks{Manuscript received ...; revised ...}}
%\thanks{Manuscript received April 19, 2005; revised August 26, 2015.}}
% The paper headers
\markboth{Journal of \LaTeX\ Class Files,~Vol.~., No.~., Month~.}%
% The paper headers
{Marvasti-Zadeh \MakeLowercase{\textit{et~al.}}: Deep Learning for Visual Tracking: A Comprehensive Survey}
\maketitle
%----------------------------------------------------------------%
\begin{abstract}
Visual target tracking is one of the most sought-after yet challenging research topics in computer vision. Given the ill-posed nature of the problem and its popularity in a broad range of real-world scenarios, a number of large-scale benchmark datasets have been established, on which considerable methods have been developed and demonstrated with significant progress in recent years -- predominantly by recent \textit{deep learning} (DL)-based methods. This survey aims to systematically investigate the current DL-based visual tracking methods, benchmark datasets, and evaluation metrics. It also extensively evaluates and analyzes the leading visual tracking methods. First, the fundamental characteristics, primary motivations, and contributions of DL-based methods are summarized from nine key aspects of: network architecture, network exploitation, network training for visual tracking, network objective, network output, exploitation of correlation filter advantages, aerial-view tracking, long-term tracking, and online tracking. Second, popular visual tracking benchmarks and their respective properties are compared, and their evaluation metrics are summarized. Third, the state-of-the-art DL-based methods are comprehensively examined on a set of well-established benchmarks of OTB2013, OTB2015, VOT2018, LaSOT, UAV123, UAVDT, and VisDrone2019. Finally, by conducting critical analyses of these state-of-the-art trackers quantitatively and qualitatively, their pros and cons under various common scenarios are investigated. It may serve as a gentle use guide for practitioners to weigh when and under what conditions to choose which method(s). It also facilitates a discussion on ongoing issues and sheds light on promising research directions.
\end{abstract}
%----------------------------------------------------------------%
% Note that keywords are not normally used for peerreview papers.

\IEEEpeerreviewmaketitle
%----------------------------------------------------------------%
%----------------------------------------------------------------%
%----------------------------------------------------------------%
\section{Introduction}\label{sec:1}
\begin{figure}[bt!]
\vspace{-5mm}
\centering
\includegraphics[width=0.8\linewidth]{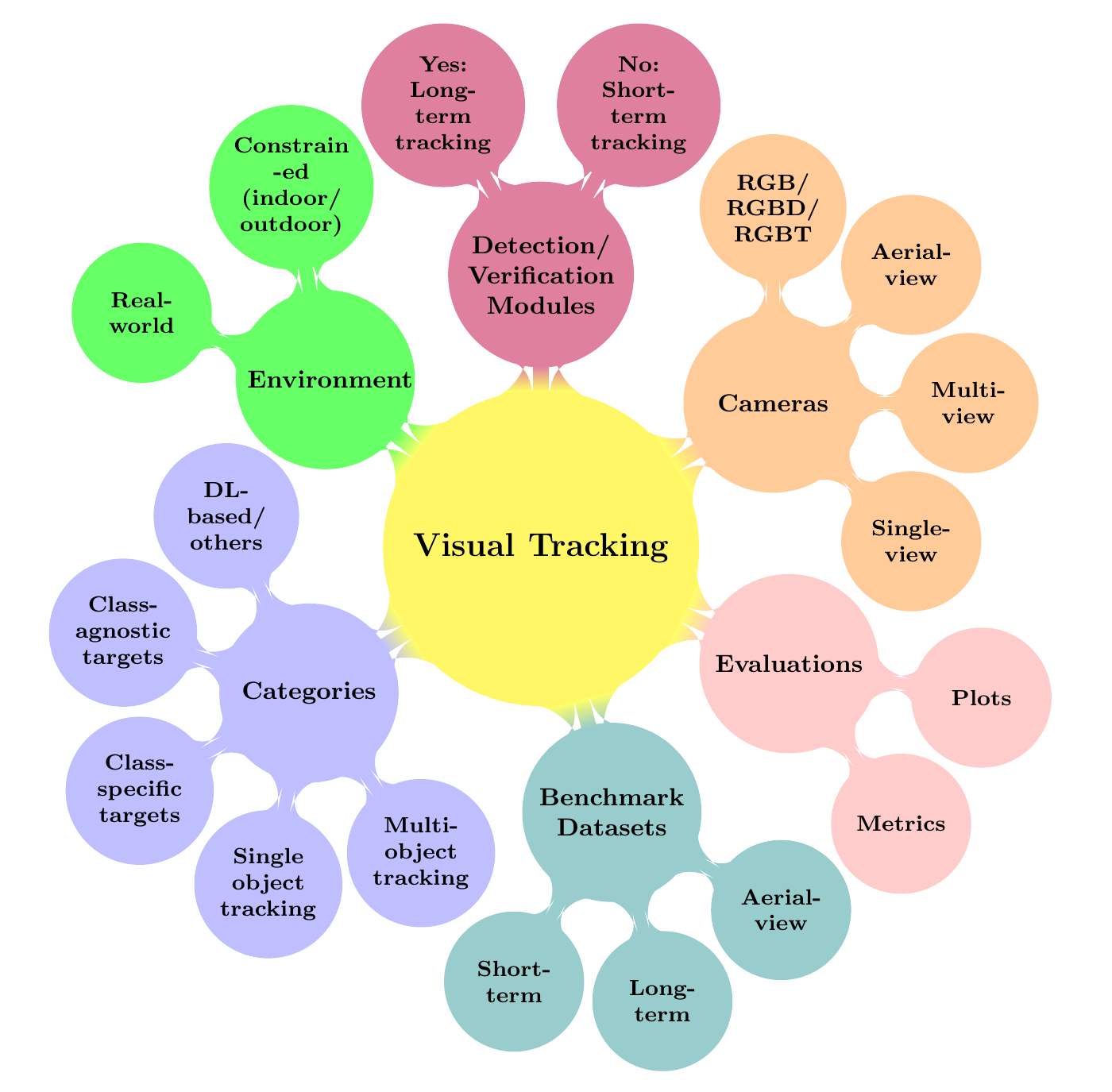}
\vspace{-.35cm}
\caption{An overview of visual target tracking.}
\label{fig:Overview} 
\vspace{-6mm}
\end{figure}

\begin{figure}[t!]
\justify
\includegraphics[width=.98\linewidth]{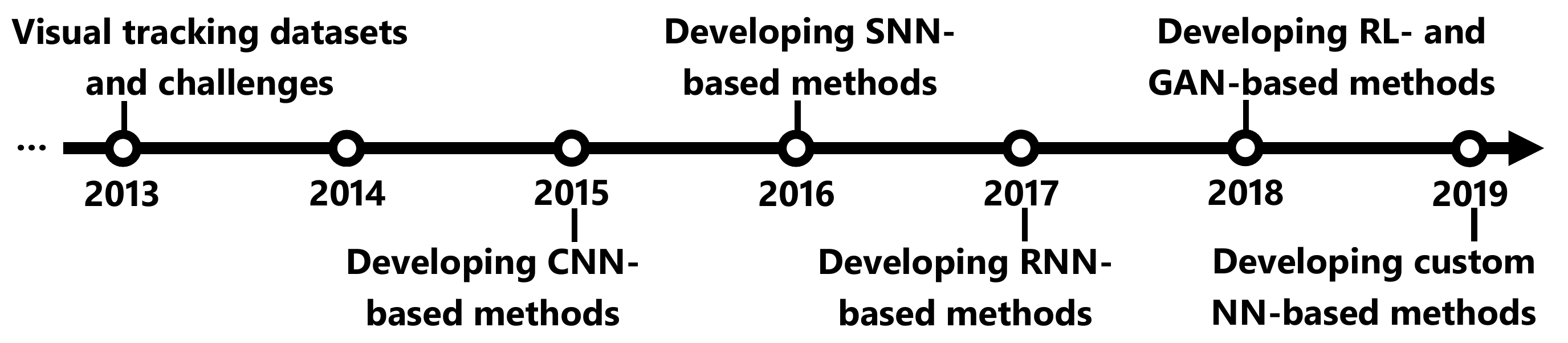}
\vspace{-.3cm}
\caption{Timeline of deep visual tracking methods.\protect\\ 2015: Exploring/studying deep features to exploit the traditional methods.\protect\\ 
2016: Offline training/fine-tuning of DNNs, employing Siamese networks for real-time tracking, and integrating DNNs into traditional frameworks.\protect\\ 
2017: Incorporating temporal \& contextual information, and investigating various offline training on large-scale datasets.\protect\\ 
2018: Studying different learning \& search strategies, and designing more sophisticated architectures for visual tracking task.\protect\\ 
2019: Investigating deep detection \& segmentation approaches, and taking advantages of deeper backbone networks.}
\label{fig:Timeline} 
\vspace{-.5cm}
\end{figure}

\IEEEPARstart{G}{eneric} visual tracking aims to estimate an unknown visual target trajectory when only an initial state of the target (in a video frame) is available. Visual tracking is an open and attractive research field with a broad extent of categories (see Fig.~\ref{fig:Overview}) and applications, including self-driving cars \cite{ITS-AutVehicle}, autonomous robots \cite{SurveyMultiRobot}, surveillance \cite{ITS-HumanTracking}, augmented reality \cite{Ababsa2008}, aerial-view tracking \cite{ReviewUAV}, sports \cite{SurveySoccer}, surgery \cite{ReviewSurgical}, biology \cite{Ulman2017}, ocean exploration \cite{ReviewUnderwater}, to name a few. 
%----------------------------------------------------------------%
%----------------------------------------------------------------%
The ill-posed definition of the visual tracking (i.e., model-free tracking, on-the-fly learning, single-camera, 2D information) is more challenging in complicated real-world scenarios which may include arbitrary classes of targets (e.g., human, drone, animal, vehicle) and motion models, various imaging characteristics (e.g., static/moving camera, smooth/fast movement, camera resolution), and changes in environmental conditions (e.g., illumination variation, background clutter, crowded scenes). Traditional methods employ various visual tracking frameworks, such as \textit{discriminative correlation filters} (DCF) \cite{KCF,ITS-KCF,RADSST,DCF_Mar1,DCF_Mar2,DCF_Mar3,DCF_Mar4}, silhouette tracking \cite{Xiao2016}, Kernel tracking \cite{Bruni2014}, point tracking \cite{Lychkov2018} -- for appearance \& motion modeling of a target. In general, traditional trackers have inflexible assumptions about target structures \& their motion in real-world scenarios. These trackers exploit handcrafted features (e.g., the \textit{histogram of oriented gradients} (HOG) \cite{HOG} and \textit{Color-Names} (CN) \cite{CN}), so they cannot interpret semantic target information and handle significant appearance changes. However, some tracking-by-detection methods (e.g., DCF-based trackers) provide an appealing trade-off of competitive tracking performance and efficient computations \cite{SRDCF,SRDCFdecon,BACF}. For instance, aerial-view trackers \cite{UAV-AutoTrack,UAV-Aberrance,UAV-Distillation} extensively use these CPU-based algorithms considering limited on-board computational power \& embedded hardware. \\
\indent Inspired by \textit{deep learning} (DL) breakthroughs \cite{AlexNet,VGGM,VGGNet,GoogLeNet,ResNet} in \textit{ImageNet large-scale visual recognition competition} (ILSVRC) \cite{ImageNet} and also \textit{visual object tracking} (VOT) challenge \cite{VOT-2013,VOT-2014,VOT-2015,VOT-2016,VOT-2017,VOT-2018,VOT-2019}, DL-based methods have attracted considerable interest in the visual tracking community to provide robust trackers. Although \textit{convolutional neural networks} (CNNs) have been dominant networks initially, a broad range of architectures such as \textit{recurrent neural networks} (RNNs), \textit{auto-encoders} (AEs), \textit{generative adversarial networks} (GANs), and especially \textit{Siamese neural networks} (SNNs) \& custom neural networks are currently investigated. Fig.~\ref{fig:Timeline} presents a brief history of the development of deep visual trackers in recent years. The state-of-the-art DL-based visual trackers have distinct characteristics such as exploitation of various architectures, backbone networks, learning procedures, training datasets, network objectives, network outputs, types of exploited deep features, CPU/GPU implementations, programming languages \& frameworks, speed, and so forth. Therefore, this work provides a comparative study of DL-based trackers, benchmark datasets, and evaluation metrics to investigate proposed trackers in detail and facilitate developing advanced trackers.\\
%----------------------------------------------------------------%
%----------------------------------------------------------------%
% \subsection{Comparison with Previous Surveys}
Visual target trackers can be roughly classified into two main categories, before and after the revolution of DL in computer vision. The first category is primarily reviewed by \cite{SurveyYilmaz,SurveySmeulders,SurveyYang,SurveyLi-AppearanceModels}, which include traditional trackers based on classical appearance \& motion models, and then examine their pros and cons systematically, experimentally, or both. These trackers employ manually-designed features for target modeling to alleviate appearance variations and to provide efficient computational complexity. For instance, although these trackers are suitable to implement on the flying robots \cite{UAV-Aberrance,UAV-AutoTrack,UAV-Boundary,UAV-Distillation,UAV-KeyFilter} due to the restrictions of using advanced GPUs, they do not have enough robustness to handle the challenges of in-the-wild videos. Typically, these trackers try to ensemble multiple features to construct a complementary set of visual cues. But, tuning an optimal trade-off that also maintains efficiency for real-world scenarios is tricky. Considering DL-based trackers' significant progress in recent years, the reviewed methods by the mentioned works are outdated. \\
\indent The second category includes DL-based trackers that employ either deep off-the-shelf features or end-to-end networks. A straightforward approach is integrating pre-trained deep features into the traditional frameworks. However, such trackers result in inconsistency problems considering task differences. But, end-to-end trained visual trackers have been investigated regarding existing tracking challenges. Recently, \cite{TrackingNoisyTargets,HandcraftedDeepTrackers,SurveyDeepTracking} review limited DL-based visual trackers. For instance, \cite{TrackingNoisyTargets,HandcraftedDeepTrackers} categorize some handcrafted \& deep methods into the correlation filter trackers \& non-correlation filter ones. Then, a further classification based on architectures \& tracking mechanisms has been applied. The work \cite{SurveySiamese} particularly investigates some SNN-based trackers based on their network branches, layers, and training aspects. However, it does not include state-of-the-art trackers and custom networks with \& without partial exploitation of SNNs. At last, the work \cite{SurveyDeepTracking} categorizes the DL-based trackers according to their structure, function, and training. Then, the evaluations are performed to conclude the categorizations based on the observations. From the structure perspective, the trackers are categorized into the CNN, RNN, and others, while they are classified into the \textit{feature extraction network} (FEN) or \textit{end-to-end network} (EEN) according to their functionality in visual tracking. The EENs are also classified in terms of the outputs, including object score, confidence map, and \textit{bounding box} (BB). Finally, DL-based methods are categorized according to pre-training \& online learning based on the network training perspective. \\
\indent According to the previous efforts, the motivations of this work are presented as follows. \\
\indent 1) Despite all efforts, existing review papers do not include state-of-the-art visual trackers that roughly employ Siamese or customized networks. \\
\indent 2) Notwithstanding significant progress in recent years, long-term trackers and tracking from aerial-views have not yet been studied. Hence, investigating the current issues and proposed solutions are necessary. \\
\indent 3) Many details are ignored in previous works studying the DL-based trackers (e.g., backbone networks, training details, exploited features, implementations, etc.). \\
\indent 4) State-of-the-art benchmark datasets (short-term, long-term, aerial-view) are not compared completely. \\
\indent 5) Finally, exhaustive comparisons of DL-based trackers on a wide variety of benchmarks have not been previously performed. These analyzes can demonstrate the advantages and limitations of existing trackers. \\
\indent Motivated by the aforementioned concerns, this work's primary goals are filling the gaps, investigating the present issues, and studying potential future directions. Thus, we focus merely on extensive state-of-the-art DL-based trackers, namely: HCFT \cite{HCFT}, DeepSRDCF \cite{DeepSRDCF}, FCNT \cite{FCNT}, CNN-SVM \cite{CNN-SVM}, DPST \cite{DPST}, CCOT \cite{CCOT}, GOTURN \cite{GOTURN}, SiamFC \cite{SiamFC}, SINT \cite{SINT}, MDNet \cite{MDNet}, HDT \cite{HDT}, STCT \cite{STCT}, RPNT \cite{RPNT}, DeepTrack \cite{DeepTrack,DeepTrack_BMVC}, CNT \cite{CNT}, CF-CNN \cite{CF-CNN}, TCNN \cite{TCNN}, RDLT \cite{RDLT}, PTAV \cite{PTAV-ICCV,PTAV}, CREST \cite{CREST}, UCT/UCT-Lite \cite{UCT}, DSiam/DSiamM \cite{DSiam}, TSN \cite{TSN}, WECO \cite{WECO}, RFL \cite{RFL}, IBCCF \cite{IBCCF}, DTO \cite{DTO}], SRT \cite{SRT}, R-FCSN \cite{R-FCSN}, GNET \cite{GNet}, LST \cite{LST}, VRCPF \cite{VRCPF}, DCPF \cite{DCPF}, CFNet \cite{CFNet}, ECO \cite{ECO}, DeepCSRDCF \cite{DeepCSRDCF}, MCPF \cite{MCPF}, BranchOut \cite{BranchOut}, DeepLMCF \cite{DeepLMCF}, Obli-RaFT \cite{Obli-RaFT}, ACFN \cite{ACFN}, SANet \cite{SANet}, DCFNet/DCFNet2 \cite{DCFNet}, DET \cite{DET}, DRN \cite{DRN}, DNT \cite{DNT}, STSGS \cite{STSGS}, TripletLoss \cite{Tripletloss}, DSLT \cite{DSLT}, UPDT \cite{UPDT}, ACT \cite{ACT}, DaSiamRPN \cite{DaSiamRPN}, RT-MDNet \cite{RT-MDNet}, StructSiam \cite{StructSiam}, MMLT \cite{MMLT}, CPT \cite{CPT}, STP \cite{STP}, Siam-MCF \cite{Siam-MCF}, Siam-BM \cite{Siam-BM}, WAEF \cite{WAEF}, TRACA \cite{TRACA}, VITAL \cite{VITAL}, DeepSTRCF \cite{STRCF}, SiamRPN \cite{SiamRPN}, SA-Siam \cite{SA-Siam}, FlowTrack \cite{FlowTrack}, DRT \cite{DRT}, LSART \cite{LSART}, RASNet \cite{RASNet}, MCCT \cite{MCCT}, DCPF2 \cite{DCPF2}, VDSR-SRT \cite{VDSR-SRT}, FCSFN \cite{FCSFN}, FRPN2T-Siam \cite{FRPN2T-siam}, FMFT \cite{FMFT}, IMLCF \cite{IMLCF}, TGGAN \cite{TGGAN}, DAT \cite{DAT}, DCTN \cite{DCTN}, FPRNet \cite{FPRNet}, HCFTs \cite{HCFTs}, adaDDCF \cite{adaDDCF}, YCNN \cite{YCNN}, DeepHPFT \cite{DeepHPFT}, CFCF \cite{CFCF}, CFSRL \cite{CFSRL}, P2T \cite{P2T}, DCDCF \cite{DCDCF}, FICFNet \cite{FICFNet}, LCTdeep \cite{LCTdeep}, HSTC \cite{HSTC}, DeepFWDCF \cite{DeepFWDCF}, CF-FCSiam \cite{CF-FCSiam}, MGNet \cite{MGNet}, ORHF \cite{ORHF}, ASRCF \cite{ASRCF}, ATOM \cite{ATOM}, C-RPN \cite{CRPN}, GCT \cite{GCT}, RPCF \cite{RPCF}, SPM \cite{SPM-Tracker}, SiamDW \cite{SiamDW}, SiamMask \cite{SiamMask}, SiamRPN++ \cite{SiamRPN++}, TADT \cite{TADT}, UDT \cite{UDT}, DiMP \cite{DiMP}, ADT \cite{ADT}, CODA \cite{CODA}, DRRL \cite{DRRL}, SMART \cite{SMART}, MRCNN \cite{MRCNN}, MM \cite{MM}, MTHCF \cite{MTHCF}, AEPCF \cite{AEPCF}, IMM-DFT \cite{IMM-DFT}, TAAT \cite{TAAT}, DeepTACF \cite{DeepTACF}, MAM \cite{MAM}, ADNet \cite{ADNet-CVPR,ADNet-TNNLS}, C2FT \cite{C2FT}, DRL-IS \cite{DRL-IS}, DRLT \cite{DRLT}, EAST \cite{EAST}, HP \cite{HP}, P-Track \cite{P-Track}, RDT \cite{RDT}, SINT++ \cite{SINT++}, Meta-Tracker \cite{Meta-Tracker}, CRVFL \cite{CRVFL}, VTCNN \cite{VTCNN}, BGBDT \cite{BGBDT}, GFS-DCF \cite{GFS-DCF}, GradNet \cite{GradNet}, MLT \cite{MLT}, UpdateNet \cite{UpdateNet}, CGACD \cite{CGACD}, CSA \cite{CSA}, D3S \cite{D3S}, OSAA \cite{OSAA}, PrDiMP \cite{PrDiMP}, RLS \cite{RLS}, ROAM \cite{ROAM}, SiamAttn \cite{SiamAttn}, SiamBAN \cite{SiamBAN}, SiamCAR \cite{SiamCAR}, SiamRCNN \cite{SiamRCNN}, TMAML \cite{TMAML}, FGTrack \cite{FGTrack}, DHT \cite{DHT}, MLCFT \cite{MLCFT}, DSNet \cite{DSNet}, BEVT \cite{BEVT}, CRAC \cite{CRAC}, KAOT \cite{KAOT_ICRA,KAOT_TMM}, MKCT \cite{MKCT}, SASR \cite{SASR}, COMET \cite{COMET}, FGLT \cite{FGLT}, GlobalTrack \cite{GlobalTrack}, i-Siam \cite{i-Siam}, LRVN \cite{LRVN}, MetaUpdater \cite{MetaUpdater}, SPLT \cite{SPLT}.\\
\indent According to the network architecture, these trackers are classified into CNN-, SNN-, RNN-, GAN-, and custom-based (i.e., AE- \& \textit{reinforcement learning} (RL)-based, and combined) networks. It indicates the popularity of different approaches, which also their problems and proposed solutions are studied in this work. From the network exploitation, the methods are categorized into the exploitation of deep off-the-shelf features and deep features for visual tracking (similar to FENs \& EENs in \cite{SurveyDeepTracking}). However, in this work, the detailed characteristics of various trackers are investigated, such as backbone networks, exploited layers, training datasets, objective functions, tracking speeds, extracted features, network outputs, CPU/GPU implementations, programming languages, and DL framework. From the network training perspective, this work separately studies deep off-the-shelf features and deep features for visual tracking since deep off-the-shelf features (extracted from FENs) are mostly pre-trained on the ImageNet for object recognition tasks. Besides, the end-to-end training for visual tracking purposes is categorized into exploiting offline training, online training, or both. Furthermore, meta-learning based visual trackers are investigated, which are recently employed to adapt visual trackers to unseen targets fast. Moreover, this work presents all the details about backbone networks, offline \& online training datasets, strategies to avoid over-fitting, data augmentations, and many more. From exploiting the advantages of correlation filters, the trackers are also classified as the methods based on DCF and the ones that employ end-to-end networks, which take advantage of the online learning efficiency of DCFs \& the discriminative power of CNN features. Next, DL-based trackers are classified based on their application for aerial-view tracking, long-term tracking, or online tracking. Finally, this work comprehensively analyses different aspects of extensive state-of-the-art trackers on seven benchmark datasets, namely OTB2013 \cite{OTB2013}, OTB2015 \cite{OTB2015}, VOT2018 \cite{VOT-2018}, LaSOT \cite{LaSOT}, UAV123 \cite{UAV123}, UAVDT \cite{UAVDT2018}, and VisDrone2019-test-dev \cite{VisDrone2019}.
\vspace{-.2cm}
%----------------------------------------------------------------%
\subsection{Contributions} \label{Sec1.1}
\indent The main contributions are summarized as follows.\\
\indent 1) State-of-the-art DL-based visual trackers are categorized based on the architecture (i.e., CNN, SNN, RNN, GAN, and custom networks), network exploitation (i.e., off-the-shelf deep features and deep features for visual tracking), network training for visual tracking (i.e., only offline training, only online training, both offline \& online training, meta-learning), network objective (i.e., regression-based, classification-based, and both classification \& regression-based), exploitation of correlation filter advantages (i.e., DCF framework and utilizing correlation filter/layer/function), aerial-view tracking, long-term tracking, and online tracking. Such a study covering all of these aspects in the detailed categorization of visual tracking methods has not been previously presented.\\
\indent 2) The main issues and proposed solutions of DL-based trackers to tackle visual tracking challenges are presented. This classification provides proper insight into designing visual trackers.\\
\indent 3) The well-known single-object visual tracking datasets (i.e., short-term, long-term, aerial view) are completely compared based on their fundamental characteristics (e.g., the number of videos, frames, classes/clusters, sequence attributes, absent labels, and overlap with other datasets). These benchmark datasets include OTB2013 \cite{OTB2013}, OTB2015 \cite{OTB2015}, VOT \cite{VOT-2013,VOT-2014,VOT-2015,VOT-2016,VOT-2017,VOT-2018,VOT-2019}, ALOV \cite{SurveySmeulders}, TC128 \cite{TC128}, UAV123 \cite{UAV123}, NUS-PRO \cite{NUS-PRO}, NfS \cite{NfS}, DTB \cite{DTB}, TrackingNet \cite{TrackingNet},  OxUvA \cite{OxUvA}, BUAA-PRO \cite{BUAA-PRO}, GOT10k \cite{GOT-10k}, LaSOT \cite{LaSOT}, UAV20L \cite{UAV123}, TinyTLP/TLPattr \cite{TLP}, TLP \cite{TLP}, TracKlinic \cite{TracKlinic}, UAVDT \cite{UAVDT2018}, LTB35 \cite{Long-term_NYSM}, VisDrone \cite{VisDrone2018,VisDrone2019}, VisDrone2019L \cite{VisDrone2019}, and Small-90/Small-112 \cite{Small90Dataset}.\\
\indent 4) Finally, extensive experimental evaluations are performed on a wide variety of tracking datasets, namely OTB2013 \cite{OTB2013}, OTB2015 \cite{OTB2015}, VOT2018 \cite{VOT-2018}, LaSOT \cite{LaSOT}, UAV123 \cite{UAV123}, UAVDT \cite{UAVDT2018}, and VisDrone2019 \cite{VisDrone2019}, and the state-of-the-art visual trackers are analyzed based on different aspects. Moreover, this work specifies the most challenging visual attributes for the VOT2018 dataset and OTB2015, LaSOT, UAV123, UAVDT, and VisDrone2019 datasets. By doing so, the most primary challenges of each dataset for recent trackers are specified. \\
\indent According to the comparisons, the following remarks are concluded. \\
\indent 1) The Siamese-based networks are the most promising deep architectures due to their satisfactory balance between performance and efficiency for visual tracking. Moreover, some methods recently attempt to exploit the advantages of RL \& GAN approaches to refine their decision-making and alleviate the lack of training data. Based on these advantages, recent trackers aim to design custom neural networks to fully exploit scene information. \\
\indent 2) The offline end-to-end learning of deep features appropriately transfers pre-trained generic features to visual tracking task. Although conventional online training of DNN increases the computational complexity such that most of these methods are not suitable for real-time applications, it considerably helps visual trackers to adapt with significant appearance variation, prevent visual distractors, and improve the performance of visual trackers. Exploiting meta-learning approaches have provided significant advances to the online adaptation of visual trackers. Therefore, both offline and (efficient) online training procedures result in promising tracking performances. \\
\indent 3) Leveraging deeper and wider backbone networks improves the discriminative power of distinguishing the target from its background. Pre-trained networks (e.g., ResNet \cite{ResNet}) are sub-optimal, and tracking performance can be remarkably improved by training backbone networks for visual tracking.  \\
\indent 4) The best trackers exploit both regression \& classification objective functions to distinguish the target from the background and find the tightest BB for target localization. These objectives are complementary such that the regression function has the role of auxiliary supervision on the classification one. Recently, \textit{video object segmentation} (VOS) approaches are integrated into visual trackers for representing targets by segmentation masks. \\
\indent 5) The exploitation of different features enhances the robustness of the target model. For instance, most of the DCF-based methods fuse the deep off-the-shelf features and hand-crafted features (e.g., HOG \& CN) for this reason. Also, the exploitation of complementary features, such as temporal or contextual information, has led to more robust features in challenging scenarios. \\
\indent 6) The most challenging attributes for DL-based visual tracking methods are occlusion, out-of-view, fast motion, aspect ratio change, and similar objects. Moreover, visual distractors with similar semantics may result in the drift problem.\\
\indent The rest of this paper is as follows. Section \ref{sec:2} introduces our taxonomy of deep visual trackers. The visual tracking benchmark datasets and evaluation metrics are compared in Section \ref{sec:3}. Experimental comparisons of the state-of-the-art visual tracking methods are performed in Section \ref{sec:4}. Finally, Section \ref{sec:5} summarizes the conclusions and future directions.
%----------------------------------------------------------------%
\section{Deep Visual Tracking Taxonomy}\label{sec:2}
Generally, three major components of: i) target representation/information, ii) training process, and iii) learning procedure play important roles in designing visual tracking methods. Most DL-based trackers aim to improve a target representation by utilizing/fusing deep hierarchical features, exploiting contextual/motion information, and select more discriminative/robust deep features. To effectively train DNNs for visual tracking systems, general motivations can be classified into employing various training schemes (e.g., network pre-training, online training (also meta-learning), or both) and handling training problems (e.g., lacking training samples, over-fitting, or computational complexity). Unsupervised training is another recent scheme to use abundant unlabeled samples, which can be performed by clustering the samples according to contextual information, mapping training data to a manifold space, or exploiting consistency-based objective function. Finally, the primary motivations regarding learning procedures are online update schemes, scale/aspect ratio estimation, search strategies, and long-term memory. \\
% Then, the proposed comprehensive taxonomy of DL-based methods is presented. \\
%----------------------------------------------------------------%
\begin{figure*}
\vspace{-5mm}
\centering
\includegraphics[width=0.98\linewidth]{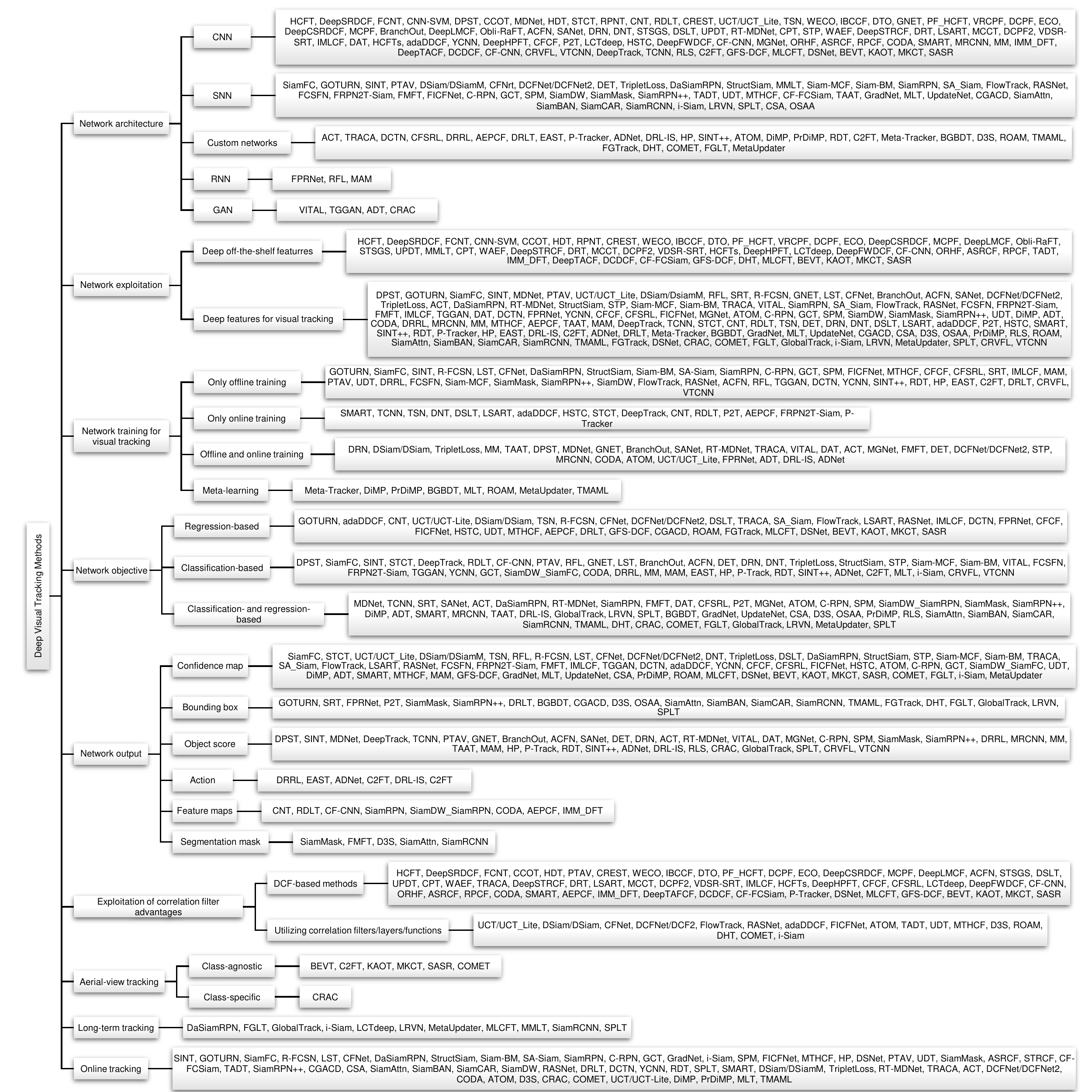}
\vspace{-2mm}
\caption{Taxonomy of DL-based visual tracking methods.}
\label{fig:Taxonomy} 
\vspace{-4mm}
\end{figure*}
%----------------------------------------------------------------%
\indent In the following, DL-based methods are comprehensively categorized based on nine aspects, and the main motivations and contributions of trackers are classified. Fig.~\ref{fig:Taxonomy} presents the proposed taxonomy of DL-based visual trackers, including network architecture, network exploitation, network training for visual tracking purposes, network objective, network output, exploitation of correlation filter advantages, aerial-view tracking, long-term tracking, and online tracking. Moreover, DL-based trackers are compared in detail regarding the pre-trained networks, backbone networks, exploited layers, types of deep features, the fusions of hand-crafted \& deep features, training datasets, tracking outputs, tracking speeds, hardware implementation details, programming languages, and DL frameworks.
\vspace{-.2cm}
%----------------------------------------------------------------%
\subsection{Network Architecture} \label{Sec2.1}
Although CNNs have been extensively adopted for visual tracking, other architectures also have been mainly developed to improve the efficiency and robustness of visual trackers in recent years. Accordingly, the proposed taxonomy consists of the CNN-, SNN-, GAN-, RNN-, and custom network-based visual trackers. \\
\indent CNN-based trackers were the first to provide powerful representations of a target by hierarchical processing of two-dimensional frames, independently.
However, conventional CNNs have inherent limitations, such as training on large supervised datasets, ignoring temporal dependencies, and computational complexities for online adaptation. As an alternative approach, SNN-based trackers measure the similarity between the target exemplar and the search region to overcome the limitations. Generally, SNNs employ CNN layers/blocks/networks in two or more branches for similarity learning purposes and run (near/over) real-time speed. However, online adaptation and handling challenges like occlusion are still under investigation. Architectures such as RNNs and GANs have been limitedly studied for visual tracking. In general, RNNs are used to capture temporal information among video frames, but they have limitations in their stability and long-term learning dependencies. GANs comprise generator \& discriminator sub-networks, which can provide the possibility to address some limitations. For instance, competing for these networks can help trackers handle scarce positive samples, although there are some barriers to training and generalization of GANs. Finally, recent custom networks include various architectures to strengthen learning features and reduce computational complexity. In the following, the primary contributions of DL-based visual trackers are summarized.
%----------------------------------------------------------------%
\subsubsection{\textbf{Convolutional Neural Network (CNN)}} \label{Sec2.1.1}
Motivated by CNN breakthroughs in computer vision and their attractive advantages (e.g., parameter sharing, sparse interactions, and dominant representations), a wide range of CNN-based trackers have been proposed. The main motivations are presented as follows.
%----------------------------------------------------------------%
\begin{itemize}[wide = 0pt]
\item \textbf{Robust target representation:} Providing powerful representations of targets is the primary advantage of employing CNNs for visual tracking. To learn robust target models, the contributions can be classified into:
i) offline training of CNNs on large-scale visual tracking datasets \cite{DPST,MDNet,UCT,GNet,BranchOut,ACFN,SANet,DRN,RT-MDNet,STP,IMLCF,DAT,YCNN,CFCF,MGNet,CODA,MRCNN,MM,TAAT}, 
ii) designing specific CNNs instead of employing pre-trained models \cite{DPST,MDNet,STCT,DeepTrack,CNT,TCNN,RDLT,UCT,TSN,GNet,BranchOut,ACFN,SANet,DRN,DNT,DSLT,RT-MDNet,STP,LSART,IMLCF,DAT,adaDDCF,YCNN,CFCF,P2T,HSTC,MGNet,CODA,SMART,MRCNN,MM,AEPCF,TAAT,DSNet}, 
iii) constructing multiple target models to capture varieties of target appearances \cite{TCNN,STP,LSART,MCCT,DCPF2,DeepHPFT,P2T,IMM-DFT,MKCT},
iv) incorporating spatial and temporal information to improve model generalization \cite{CREST,TSN,STSGS,WAEF,STRCF,DAT,DeepFWDCF,MGNet,KAOT_ICRA,KAOT_TMM},
v) fusion of different deep features to exploit complementary spatial and semantic information \cite{CCOT,SANet,DSLT,UPDT,IMLCF,DSNet,MLCFT,BEVT},
vi) learning particular target models such as relative model \cite{DRN} or part-based models \cite{STP,LSART,P2T} to handle partial occlusion and deformation, 
vii) utilizing a two-stream network \cite{LSART} to prevent over-fitting and to learn rotation information, and accurately estimating target aspect ratio to avoid contaminating target model with non-relevant information \cite{C2FT},
and viii) group feature selection through channel \& spatial dimensions to learn the structural relevance of features.
\item \textbf{Balancing training data:} Based on problem definition, there is just one positive sample in the first frame that increases the risk of over-fitting during tracking. Although the background information arbitrary can be considered negative in each frame, target sampling based on imperfect target estimations may also lead to noisy/unreliable training samples. These issues dramatically affect the performance of visual tracking methods. To alleviate them, CNN-based trackers propose:
i) domain adaption mechanism (i.e., transferring learned knowledge from the source domain to target one with insufficient samples) \cite{GNet,CODA},
ii) various update mechanisms (e.g., periodic, stochastic, short-term, \& long-term updates) \cite{DNT,MCCT,DeepHPFT,LCTdeep,MM},
iii) convolutional \textit{Fisher discriminative analysis} (FDA) for positive and negative sample mining \cite{adaDDCF},
iv) multiple-branch CNN for online ensemble learning \cite{BranchOut}, v) efficient sampling strategies to increase the number of training samples \cite{AEPCF}, and vi) a recursive least-square estimation algorithm to provide a compromise between the discrimination power \& update iterations during online learning \cite{RLS}.
\item \textbf{Computational complexity problem:} Despite the significant progress of CNNs in appearance representation, the CNN-based methods still suffer from high computational complexity. To reduce this limitation, CNN-based visual tracking methods exploit different solutions, namely:
i) employing a straightforward CNN architecture \cite{VTCNN},
ii) disassembling a CNN into several shrunken networks \cite{RDLT},
iii) compressing or pruning training sample space \cite{ECO,CPT,adaDDCF,MGNet,MRCNN} or feature selection \cite{FCNT,ORHF},
iv) feature computation via RoIAlign operation \cite{RT-MDNet} (i.e., feature approximation via bilinear interpolation) or oblique random forest \cite{Obli-RaFT} for better data capturing,
v) corrective domain adaption method \cite{CODA},
vi) lightweight structure \cite{DeepTrack,CNT,SMART},
vii) efficient optimization processes \cite{DeepLMCF,ASRCF},
viii) particle sampling strategy \cite{MCPF},
ix) utilizing attentional mechanism \cite{ACFN}, x) extending the random vector functional link (RVFL) network to a convolutional structure \cite{CRVFL}, and xi) exploiting advantages of correlation filters \cite{HCFT,DeepSRDCF,FCNT,CCOT,HDT,CF-CNN,PTAV-ICCV,PTAV,CREST,UCT,WECO,IBCCF,DTO,DCPF,ECO,DeepCSRDCF,MCPF,DeepLMCF,ACFN,STSGS,DSLT,UPDT,CPT,WAEF,STRCF,DRT,LSART,MCCT,DCPF2,VDSR-SRT,IMLCF,HCFTs,adaDDCF,DeepHPFT,CFCF,LCTdeep,HSTC,DeepFWDCF,ASRCF,RPCF,CODA,SMART,AEPCF,IMM-DFT,DeepTACF} for efficient computations. Exploiting the advantages of correlation filters refers to either applying DCFs on pre-trained networks or combining correlation filters/layers/functions with end-to-end networks.
\end{itemize}
%----------------------------------------------------------------%
\subsubsection{\textbf{Siamese Neural Network (SNN)}}  \label{Sec2.1.2}
SNNs are widely employed for visual trackers in the past few years. Given the pairs of target and search regions, these two-stream networks compute the same function to produce a similarity map. They mainly aim to overcome the limitations of pre-trained deep CNNs and take full advantage of end-to-end learning for real-time applications.
%----------------------------------------------------------------%
\begin{itemize}[wide = 0pt]
\item \textbf{Discriminative target representation:} The ability to construct a robust target model majorly relies on target representation. For achieving more discriminative deep features and improving target modeling, SNN-based methods propose:
i) learning distractor-aware \cite{DaSiamRPN} or target-aware features \cite{TADT},
ii) fusing deep multi-level features \cite{FCSFN,CRPN} or combining confidence maps \cite{R-FCSN,LST,SA-Siam},
iii) utilizing different loss functions in Siamese formulation  to train more effective filters \cite{Tripletloss,SiamMask,TADT,UDT,DiMP},
iv) leveraging different types of deep features such as context information \cite{Siam-MCF,SA-Siam,GCT} or temporal features/models \cite{GOTURN,DSiam,FlowTrack,FRPN2T-siam,GCT,MAM},
v) full exploring of low-level spatial features \cite{FCSFN,CRPN},
vi) considering angle estimation of a target to prevent salient background objects \cite{Siam-BM},
vii) utilizing multi-stage regression  to refine target representation \cite{CRPN}, vii) using the deeper and wider deep network as the backbone to increase the receptive field of neurons, which is equivalent to capturing the structure of the target \cite{SiamDW}, viii) employing correlation-guided attention modules to exploit the relationship between the template \& RoI feature maps \cite{CGACD}, ix) computing correlations between attentional features \cite{SiamAttn}, x) accurately estimating the scale and aspect ratio of the target \cite{SiamBAN}, xi) simultaneously learning classification \& regression models \cite{SiamCAR}, xii) mining hard samples in training \cite{SiamRCNN}, and xiii) employing skimming \& perusal modules for inferring optimal target candidates \cite{SPLT}. Finally, \cite{CSA,OSAA} are performed adversarial attacks on SNN-based trackers to evaluate misbehaving SNN models for visual tracking scenarios. These methods generate slight perturbations for deceiving the trackers to finally investigate DL models and improve their robustness.
\item \textbf{Adapting target appearance variation:} Using only offline training of the first generation of SNN-based trackers caused a poor generalization to unseen targets. To solve it, recent SNN-based trackers propose:
i) online update strategies considering strategies to reduce the risk of over-fitting \cite{DSiam,LST,CFNet,DET,DaSiamRPN,CF-FCSiam,GradNet,MLT,UpdateNet,LRVN},
ii) background suppression \cite{DSiam,DaSiamRPN},
iii) formulating tracking task as a one-shot local detection task \cite{DaSiamRPN,SiamRPN}, 
iv) and giving higher weights to important feature channels or score maps \cite{R-FCSN,SA-Siam,RASNet,FICFNet}, and v) modeling all potential distractors considering their motion and interaction \cite{SiamRCNN}.
Alternatively, the DaSiamRPN \cite{DaSiamRPN} and MMLT \cite{MMLT} use a local-to-global search region strategy and memory exploitation to handle critical challenges such as full occlusion and out-of-view and enhance local search strategy.
\item \textbf{Balancing training data:} The same as CNN-based methods, some efforts by SNN-based methods have been performed to address the imbalance distribution of training samples. The main contributions of the SNN-based methods are:
i) exploiting multi-stage Siamese framework to stimulate hard negative sampling \cite{CRPN}, 
ii) adopting sampling heuristics such as fixed foreground-to-background ratio \cite{CRPN} or sampling strategies such as random sampling \cite{DaSiamRPN} or flow-guided sampling \cite{FRPN2T-siam}, and 
iii) taking advantages of correlation filter/layer into Siamese framework \cite{PTAV-ICCV,PTAV,DSiam,CFNet,DCFNet,DaSiamRPN,SiamRPN,FlowTrack,RASNet,FICFNet,ORHF,ATOM,TADT,UDT,MTHCF,CF-FCSiam}.
\end{itemize}
%----------------------------------------------------------------%
\subsubsection{\textbf{Recurrent Neural Network (RNN)}} \label{Sec2.1.3}
Since visual tracking is related to both spatial and temporal information of video frames, RNN-based methods also consider target motion/movement. Because of arduous training and a considerable number of parameters, the number of RNN-based methods is limited. Almost all these methods try to exploit additional information and memory to improve target modeling. Also, the second aim of using RNN-based methods is to avoid fine-tuning of pre-trained CNN models, which takes a lot of time and is prone to over-fitting. The primary purposes of these methods can be classified to the spatio-temporal representation capturing \cite{RFL,FPRNet,MAM}, leveraging contextual information to handle background clutters \cite{FPRNet}, exploiting multi-level visual attention to highlight target, background suppression \cite{MAM}, and using convolutional \textit{long short-term memory} (LSTM) as the memory unit of previous target appearances \cite{RFL}. Moreover, RNN-based methods exploit pyramid multi-directional recurrent network \cite{FPRNet} or incorporate LSTM into different networks \cite{RFL} to memorize target appearance and investigate time dependencies. Finally, the \cite{FPRNet} encodes the self-structure of a target to reduce tracking sensitivity related to similar distractors.
%----------------------------------------------------------------%
\subsubsection{\textbf{Generative Adversarial Network (GAN)}} \label{Sec2.1.4}
Based on some attractive advantages, such as capturing statistical distribution and generating desired training samples without extensive annotated data, GANs have been intensively utilized in many research areas. Although GANs are usually hard to train and evaluate, some DL-based trackers employ GANs to enrich training samples and target modeling. These networks can augment positive samples in feature space to address the imbalance distribution of training samples \cite{VITAL}. Also, the GAN-based methods can learn general appearance distribution to deal with visual tracking's self-learning problem \cite{TGGAN}. Furthermore, the joint optimization of regression \& discriminative networks will take advantage of both these two tasks \cite{ADT}. Lastly, these networks can explore the relations between a target with its contextual information for searching interested regions and transfer this information to videos with different inherits such as transferring from ground-view to drone-view \cite{CRAC}. 
%----------------------------------------------------------------%
\subsubsection{\textbf{Custom Networks}} \label{Sec2.1.5}
Inspired by particular deep architectures and network layers, modern DL-based methods have combined a wide range of networks such as AE, CNN, RNN, SNN, detection networks \& also deep RL for visual tracking. The main motivation of custom networks is to compensate for ordinary trackers' deficiencies by exploiting the advantages of other networks. Furthermore, meta-learning (or learning to learn) has been recently attracted by the visual tracking community. It aims to address few-shot learning problems and fast adaptation of a learner to a new task by leveraging accumulated experiences from similar tasks. By employing the meta-learning framework, different networks can learn unseen target appearances during online tracking. The primary motivations and contributions of custom networks are classified as follows.
\begin{itemize}[wide = 0pt] 
\item\textbf{Robust and accurate tracking:} Recent networks seek general \& effective frameworks for better localization and BB estimation. For instance, aggregating several online trackers \cite{DHT} is a way to improve tracking performance. An alternative is a better understanding of the target's pose by exclusively designed target estimation and classification networks \cite{ATOM}. Also, meta-learning based networks \cite{DiMP,PrDiMP} can predict powerful target models inspiring by discriminative learning procedures. However, some other works \cite{BGBDT,TMAML} consider the tracking task as an instance detection and aim to convert modern object detectors directly to a visual tracker. These trackers exploit class-agnostic networks, which can: i) differentiate intra-class samples, ii) quickly adapt to different targets by meta-learners, and iii) consider temporal cues. The D3S \cite{D3S} method models a visual target from its segmentation mask with complementary geometric properties to improve the robustness of template-based trackers. The COMET \cite{COMET} bridges the gap between advanced visual trackers and aerial-view ones in detecting small/tiny objects. It employs multi-scale feature learning and attention modules to compensate for the inferior performance of generic trackers in medium-/high-attitude aerial views.
\item\textbf{Computational complexity problem:} As stated before, this problem limits the performance of online trackers in real-time applications. To control computational complexity, the TRACA \cite{TRACA} and AEPCF \cite{AEPCF} methods employ AEs to compress raw conventional deep features. The EAST \cite{EAST} adaptively takes either shallow features for simple frames for tracking or expensive deep features for challenging ones \cite{EAST}, and the TRACA \cite{TRACA}, CFSRL \cite{CFSRL}, \& AEPCF \cite{AEPCF} exploit the DCF computation efficiency. An effective way to avoid high computational burden is exploiting meta-learning that quickly adapts pre-trained trackers on unseen targets. The target model of the meta-learning based trackers can be optimized in a few iterations \cite{Meta-Tracker,DiMP,PrDiMP,BGBDT,MLT,ROAM,TMAML,MetaUpdater}.
\item\textbf{Model update:} To maintain the stability of the target model during the tracking process, different update strategies have been proposed; for instance, the CFSRL \cite{CFSRL} updates multiple models in parallel, the DRRL \cite{DRRL} incorporates an LSTM to exploit long-range time dependencies, and the AEPCF \cite{AEPCF} utilizes long-term and short-term update schemes to increase tracking speed. To prevent the erroneous model update and drift problem, the RDT \cite{RDT} has revised the visual tracking formulation to a consecutive decision-making process about the best target template for the next localization. Moreover, efficient learning of good decision policies using RL \cite{P-Track} is another technique to take either model update or ignore the decision. A recent alternative is employing meta-learning approaches for quick model adaptation. For instance, the works \cite{ROAM,DiMP,PrDiMP} use recurrent optimization processes that update the target model in a few gradient steps. Finally, \cite{MetaUpdater} integrates sequential information (e.g., geometric, discriminative, and appearance cues) and exploits a meta-updater to effectively update on reliable frames.
\item\textbf{Limited training data:} The soft and non-representative training samples can disturb visual tracking if occlusion, blurring, and large deformation happen. The AEPCF \cite{AEPCF} exploits a dense circular sampling scheme to prevent the over-fitting problem caused by limited training data. The SINT++ \cite{SINT++} generates positive and hard training samples by \textit{positive sample generation network} (PSGN) and \textit{hard positive transformation network} (HPTN) to make diverse and challenging training data. To efficiently train DNNs without a large amount of training data, partially labeled training samples are utilized by an action-driven deep tracker \cite{ADNet-CVPR,ADNet-TNNLS}. The P-Track \cite{P-Track} also uses active decision-making to label videos interactively while learning a tracker when limited annotated data are available. Meta-Tracker \cite{Meta-Tracker} was the first attempt to exploit an offline meta-learning-based method for better online adaptation of visual trackers. This method can generalize the target model and avoid over-fitting to distractors. Besides, various pioneer trackers \cite{DiMP,PrDiMP,MetaUpdater,TMAML,BGBDT,MLT,ROAM} enjoy the advantages of meta-learners in one/few-shot learning tasks. 
\item\textbf{Search strategy:} From the definition, visual tracking methods estimate the new target state in the next frame's search region, given an initial target state in the first frame. The best search region selection depends on the iterative search strategies that usually are independent of video content and are heuristic, brute-force, and hand-engineered. Despite classical search strategies based on sliding windows, mean shift, or particle filter, the state-of-the-art DL-based visual trackers exploit RL-based methods to learn data-driven searching policies. To exhaustively explore a region of interest and select the best target candidate, action-driven tracking mechanisms \cite{ADNet-CVPR,ADNet-TNNLS} consider the target context variation and actively pursues the target movement. Furthermore, the ACT and DRRL have proposed practical RL-based search strategies for real-time requirements by dynamic search \cite{ACT} and coarse-to-fine verification \cite{DRRL}. Lastly, the full-image visual tracker \cite{GlobalTrack} exploits a two-stage detector for localizing the target without any assumptions (e.g., temporal consistency of target regions).
\item\textbf{Exploiting additional information:} To enhance the target model by utilizing motion or contextual information, the DCTN \cite{DCTN} establishes a two-stream network, and the SRT \cite{SRT} adopts multi-directional RNN to learn further dependencies of a target during visual tracking. Also, the FGTrack \cite{FGTrack} estimates the scale \& rotation of the target and its displacement by the finer-grained motion information provided by optical-flow. A recurrent convolutional network \cite{DRLT} models previous semantic information and tracking proposals to encode relevant information for better localization. At last, DRL-IS \cite{DRL-IS} has introduced an Actor-Critic network to estimate target motion parameters efficiently. 
\item\textbf{Decision making:} Online decision making has principal effects on the performance of DL-based visual tracking methods. The state-of-the-art methods attempt to learn online decision making by incorporating RL into the DL-based methods instead of hand-designed techniques. To gain effective decision policies, the P-Track \cite{P-Track} ultimately exploits data-driven techniques in an active agent to decide about tracking, re-initializing, or updating processes. Also, the DRL-IS \cite{DRL-IS} utilizes a principled RL-based method to select sensible action based on target status. Also, an action-prediction network has been proposed to adjust a visual tracker's continuous actions to determine the optimal hyper-parameters for learning the best action policies and make satisfactory decisions \cite{HP}. On the other hand, the work \cite{PrDiMP} considers the uncertainty in estimating target states. By predicting a conditional probability density of the visual target, direct interpretations can be provided for deciding about an update procedure or lost target. Also, a result judgment module \cite{FGLT} can help short-term trackers in occlusion/out-of-view situations.
\end{itemize}
%----------------------------------------------------------------%
\subsection{Network Exploitation} \label{Sec2.2}
Roughly speaking, there are two main exploitations of DNNs for visual tracking, including reusing a pre-trained model on partially related datasets or exploiting deep features for visual tracking, which is equivalent to train DNNs for visual tracking tasks.
%----------------------------------------------------------------%
\subsubsection{\textbf{Model Reuse or Deep Off-the-Shelf Features}} \label{Sec2.2.1}
Exploiting deep off-the-shelf features is the simplest way to transfer the power of deep features into the traditional visual tracking methods. These features provide a generic representation of visual targets and help visual tracking methods to construct more robust target models. Regarding topologies, DNNs include either a simple multi-layer stack of non-linear layers (e.g., AlexNet \cite{AlexNet}, VGGNet \cite{VGGM,VGGNet}) or a directed acyclic graph topology (e.g., GoogLeNet \cite{GoogLeNet}, ResNet \cite{RASNet}, SSD \cite{SSD}, Siamese convolutional neural network \cite{Koch2015}), which allows designing more complex deep architectures that include layers with multiple input/output. The main challenge of these trackers is how to benefit the generic representations effectively. Different methods employ various feature maps and models that have been pre-trained majorly on large-scale still images of the ImageNet dataset \cite{ImageNet} for the object recognition task. Numerous methods have studied the properties of pre-trained models and explored the impact of deep features in traditional frameworks (see Table~\ref{tab1}). As a result, the DL-based methods have preferred simultaneous exploitation of both semantic and fine-grained deep features \cite{HCFT,FCNT,CCOT,HCFTs,CRPN,RefineNet,Lin2017,MLCFT,BEVT}. 
The fusion of deep features is also another motivation of these methods, which is performed by different techniques to utilize multi-resolution deep features \cite{HCFT,DeepSRDCF,FCNT,CCOT,HDT,WECO,DCPF2,DeepHPFT,IMM-DFT,UPDT,CF-FCSiam,MCCT,MLCFT,BEVT} and independent fusion of deep features with shallow ones at a later stage \cite{UPDT}. Exploiting motion information \cite{DCPF,STSGS,IMM-DFT,DeepMotionFeatures} and selecting appropriate deep features for visual tracking tasks \cite{FCNT,ORHF,GFS-DCF} are two other interesting motivations for DL-based methods. The detailed characteristics of DL-based visual trackers based on deep off-the-shelf features are shown in Table~\ref{tab1}. Needless to say, the network output for these methods are deep feature maps.
%----------------------------------------------------------------%
\begin{table*}
\caption{Deep off-the-shelf features for visual tracking. The abbreviations are denoted as: confidence map (CM), saliency map (SM), bounding box (BB), votes (vt), deep appearance features (DAF), deep motion features (DMF).} % title of Table
\vspace{-2mm}
\centering % used for centering table
\resizebox{\textwidth}{!}{
\begin{tabular}{c c c c c c c c c c c} % centered columns (4 columns)
\hline\hline %inserts double horizontal lines
Method & Pre-trained models & Exploited layers & Pre-training data & Pre-training dataset(s) & Exploited features & PC (CPU, RAM, Nvidia GPU) & Language & Framework & Speed (fps) & Tracking output \protect\\ 
% Method & Value & Parameter & Value \\ [0.5ex] % inserts table
%heading
\hline\hline 
DeepSRDCF \cite{DeepSRDCF} & VGG-M & Conv5 & Still images & ImageNet & HOG, DAF & N/A, GPU & Matlab & MatConvNet & N/A & CM \protect\\ 
CCOT \cite{CCOT} & VGG-M & Conv1, Conv5 & Still images & ImageNet & HOG, CN, DAF & N/A, GPU & Matlab & MatConvNet & ~1 & CM \protect\\
ECO \cite{ECO} & VGG-M & Conv1, Conv5 & Still images & ImageNet & HOG, CN, DAF & N/A, GPU & Matlab & MatConvNet & 8 & CM \protect\\
DeepCSRDCF \cite{DeepCSRDCF} & VGG-M & N/A & Still images & ImageNet & HOG, CN, DAF & Intel I7 3.4GHz CPU, GPU & Matlab & MatConvNet & 13 & CM \protect\\ 
SASR \cite{SASR} & VGG-M & Conv4 & Still images & ImageNet & DAF & Intel-8700K 3.7GHz CPU, 32GB RAM, Quadro P2000 GPU & Matlab & MatConvNet & 3.84 & CM \protect\\
KAOT \cite{KAOT_ICRA,KAOT_TMM} & VGG-M & Conv3 & Still images & ImageNet & DAF & Intel I7-8700K 3.7GHz CPU, 32GB RAM, RTX 2080 GPU & Matlab & MatConvNet & 14.1 & CM \protect\\ 
MLCFT \cite{MLCFT} & VGG-M & Conv-1, Conv-3, Conv-5 & Still images & ImageNet & DAF & Intel I7 3770K 3.5 CPU, 8GB RAM, GTX 960 GPU & Matlab & MatConvNet & 16.1 & CM \protect\\
UPDT \cite{UPDT} & VGG-M/ GoogLeNet/ ResNet-50 & N/A & Still images & ImageNet & HOG, CN, DAF & N/A & Matlab & MatConvNet & N/A & CM \protect\\ 
WAEF \cite{WAEF} & VGG-M & Conv1, Conv5 & Still images & ImageNet & HOG, CN, DAF & Intel Xeon(R) 3.20 GHz CPU, 44GB RAM, GTX 1080Ti & Matlab & MatConvNet & 0.62 & CM \protect\\ 
DeepSTRCF \cite{STRCF} & VGG-M & Conv3 & Still images & ImageNet & HOG, CN, DAF & Intel I7-7700 CPU, 32GB RAM, GTX 1070 GPU & Matlab & MatConvNet & 24.3 & CM \protect\\ 
DRT \cite{DRT} & VGG-M, VGG-16 & Conv1, Conv4-3 & Still images & ImageNet & HOG, CN, DAF & N/A, GPU & Matlab & Caffe & N/A & CM \protect\\ 
WECO \cite{WECO} & VGG-M & Conv1, Conv5 & Still images & ImageNet & DAF & Intel Xeon(R) 2.60GHz CPU, GTX 1080 GPU & Matlab & MatConvNet & 4 & CM \protect\\ 
VDSR-SRT \cite{VDSR-SRT} & VGG-M & Conv1, Conv5 & Still images & ImageNet & HOG, DAF & Intel I7-6700k 4.00GHz CPU, 16GB RAM, GTX 1070 GPU & Matlab & MatConvNet & 13.5 & CM \protect\\ 
ASRCF \cite{ASRCF} & VGG-M, VGG-16 & Norm1, Conv4-3 & Still images & ImageNet & HOG, DAF & Intel I7-8700 CPU, 32GB RAM, GTX 1080Ti GPU & Matlab & MatConvNet & 28 & CM \protect\\ 
RPCF \cite{RPCF} & VGG-M & Conv1, Conv5 & Still images & ImageNet & HOG, CN, DAF & Intel I7-4790K CPU, GTX 1080 GPU & Matlab & MatConvNet & 5 & CM \protect\\ 
DeepTACF \cite{DeepTACF} & VGG-M & Conv1 & Still images & ImageNet & HOG, DAF & Intel I7-6700 3.40GHz CPU, GTX Titan GPU & Matlab & MatConvNet & N/A & CM \protect\\ 
FCNT \cite{FCNT} & VGG-16 & Conv4-3, Conv5-3 & Still images & ImageNet & DAF & 3.4GHz CPU, GTX Titan GPU & Matlab & Caffe & 3 & CM \protect\\ 
CREST \cite{CREST} & VGG-16 & Conv4-3 & Still images & ImageNet & DAF & Intel I7 3.4GHz CPU, GTX Titan Black GPU & Matlab & MatConvNet & N/A & CM \protect\\ 
DTO \cite{DTO} & VGG-16, SSD & Conv3-3, Conv4-3, Conv5-3 & Still images & ImageNet & DAF & Intel I7-4770K CPU, 32G RAM, GTX 1070 GPU & Matlab & Caffe & N/A & CM, BB \protect\\ 
VRCPF \cite{VRCPF} & VGG-16, Faster R-CNN & N/A & Still images & ImageNet, COCO & DAF & N/A & N/A & N/A & N/A & BB \protect\\ 
Obli-RaFT \cite{Obli-RaFT} & VGG-16 & Conv4-3, Conv5-3 & Still images & ImageNet & DAF & Intel I7-3770 3.40GHz CPU, 2 GTX Titan X GPUs & Matlab & Caffe & 2 & VT \protect\\ 
CPT \cite{CPT} & VGG-16 & Conv5-1, Conv5-3 & Still images & ImageNet & HOG, CN, DAF & Intel I7-7800X CPU, 16GB RAM, GTX 1080Ti GPU & Matlab & MatConvNet & 14 & CM \protect\\ 
DeepHPFT \cite{DeepHPFT} & VGG-16, VGG-19, and GoogLeNet & Conv5-3, Conv5-4, and icp6-out & Still images & ImageNet & HOG, CN, DAF & Intel Xeon 2.4GHz CPU, 256 GB RAM, GTX Titan XP GPU & Matlab & MatConvNet & 4 & CM \protect\\ 
DeepFWDCF \cite{DeepFWDCF} & VGG-16 & Conv4-3 & Still images & ImageNet & DAF & N/A, GTX 1080Ti GPU & Matlab & MatConvNet & 2.7 & CM \protect\\ 
MMLT \cite{MMLT} & VGGNet, Fully-convolutional Siamese network & Conv5 & Still images, Video frames & ImageNet, ILSVRC-VID & DAF & Intel I7-4770 3.40GHz CPU, 11GB RAM, GTX 1080Ti & Matlab & N/A & 6.15 & CM \protect\\ 
MKCT \cite{MKCT} & VGGNet & Conv3-4 & Still images & ImageNet & DAF & Intel I7 3.7GHz CPU, 32GB RAM, Quadro 2000 GPU & Matlab & MatConvNet & 9.4 & CM \protect\\ 
BEVT \cite{BEVT} & VGGNet & Conv3-4, Conv4-4, Conv5-4 & Still images & ImageNet & DAF & Intel I7-8700K 3.7GHz CPU, 48GB RAM, Quadro P2000 GPU & Matlab & MatConvNet & 0.6 & CM \protect\\ 
HCFT \cite{HCFT} & VGG-19 & Conv3-4, Conv4-4, Conv5-4 & Still images & ImageNet & DAF & Intel I7-4770 3.40GHz CPU, 32 GB RAM, GTX Titan GPU & Matlab & MatConvNet & 10.4 & CM \protect\\ 
HDT \cite{HDT} & VGG-19 & Conv4-2, Conv4-3, Conv4-4, Conv5-2, Conv5-3, Conv5-4 & Still images & ImageNet & DAF & Intel I7-4790K 4.00GHz CPU, 16GB RAM, GTX 780Ti GPU & Matlab & MatConvNet & ~1 & CM \protect\\ 
IBCCF \cite{IBCCF} & VGG-19 & Conv3-4, Conv4-4, Conv5-4 & Still images & ImageNet & DAF & Intel Xeon(R) 3.3GHz CPU, 32GB RAM, GTX 1080 GPU & Matlab & MatConvNet & 1.25 & CM \protect\\ 
DCPF \cite{DCPF} & VGG-19 & Conv3-4, Conv4-4, Conv5-4 & Still images & ImageNet & DAF & N/A & N/A & N/A & N/A & CM \protect\\ 
MCPF \cite{MCPF} & 	VGG-19 & Conv3-4, Conv4-4, Conv5-4 & Still images & ImageNet & DAF & Intel 3.10GHz CPU, 256 GB RAM, GTX Titan X GPU & Matlab & MatConvNet & 0.5 & CM \protect\\ 
DeepLMCF \cite{DeepLMCF} & VGG-19 & Conv3-4, Conv4-4, Conv5-4 & Still images & ImageNet & DAF & Intel 3.60GHz CPU, Tesla K40 GPU & Matlab & MatConvNet & 10 & CM \protect\\   
STSGS \cite{STSGS} & VGG-19 & Conv3-4, Conv4-4, Conv5-4 & Still images & ImageNet & DAF, DMF & Intel I7 3.20GHz CPU, 8 GB RAM & Matlab & Caffe & 4~5 & CM \protect\\ 
MCCT \cite{MCCT} & VGG-19 & Conv4-4, Conv5-4 & Still images & ImageNet & DAF & Intel I7-4790K 4.00GHz CPU, 16GB RAM, GTX 1080Ti GPU & Matlab & MatConvNet & 8 & CM \protect\\ 
DCPF2 \cite{DCPF2} & VGG-19 & Conv3-4, Conv4-4, Conv5-4 & Still images & ImageNet & DAF & N/A & N/A & N/A & N/A & CM \protect\\
HCFTs \cite{HCFTs} & VGG-19 & Conv3-4, Conv4-4, Conv5-4 & Still images & ImageNet & DAF & Intel I7-4770 3.40GHz CPU, 32GB RAM, GTX Titan GPU & Matlab & MatConvNet & 6.7 & CM \protect\\ 
LCTdeep \cite{LCTdeep} & VGG-19 & Conv5-4 & Still images & ImageNet & DAF & Intel I7-4770 3.40GHz CPU, 32GB RAM, GPU & Matlab & MatConvNet & 13.8 & CM \protect\\ 
CF-CNN \cite{CF-CNN} & VGG-19 & Conv3-4, Conv4-4, Conv5-4 & Still images & ImageNet & DAF & Intel I7-4770 3.40GHz CPU, 32GB RAM, GTX Titan GPU & Matlab & MatConvNet & 12.3 & CM \protect\\ 
ORHF \cite{ORHF} & VGG-19 & Conv3-4, Conv4-4, Conv5-4 & Still images & ImageNet & HOG, DAF & Intel I7-4770K 3.50GHz CPU, 24GB RAM, N/A & Matlab & N/A & N/A & CM \protect\\ 
IMM-DFT \cite{IMM-DFT} & VGG-19 & Conv3-4, Conv4-4, Conv5-4 & Still images & ImageNet & DAF & Intel I5-4590 3.30GHz CPU, 16GB RAM, GTX Titan X GPU & Matlab & MatConvNet & 10 & CM \protect\\ 
CNN-SVM \cite{CNN-SVM} & R-CNN & First fully-connected layer & Still images & ImageNet & DAF & N/A & N/A & Caffe & N/A & SM \protect\\ 
RPNT \cite{RPNT} & Object proposal network & N/A & Still images & ImageNet, PASCAL VOC & DAF & N/A & C/C++ & N/A & 3.8 & BB \protect\\ 
CF-FCSiam \cite{CF-FCSiam} & Fully-convolutional Siamese network & N/A & Video frames & ILSVRC-VID & HOG, DAF & Intel I7-6700K 4.00GHz CPU, GTX Titan GPU & Matlab & MatConvNet & ~33 & CM \protect\\ 
TADT \cite{TADT} & Siamese matching network & Conv4-1, Conv4-3 & Still images & ImageNet & DAF & Intel I7 3.60GHz CPU, 32GB RAM, GTX 1080 GPU & Matlab & MatConvNet & 33.7 & CM \protect\\ 
%-----------------------------
GFS-DCF \cite{GFS-DCF} & ResNet-50 & Res4x & Still images & ImageNet & DAF & Intel Xeon E5-2637v3 CPU, N/A, GTX Titan X GPU & Matlab & MatConvNet & 8 & CM \protect\\ 
DHT \cite{DHT} & Various networks & Various layers & Still images, Video frames & Various datasets & DAF & N/A & N/A & N/A & 12 & BB \protect\\ 
\hline %inserts single line
  \end{tabular}
  \label{tab1}
  }
\vspace{-4mm}
\end{table*}
%----------------------------------------------------------------%
\subsubsection{\textbf{Deep Features for Visual Tracking Purposes}} \label{Sec2.2.2}
One trending part of recent trackers is how to design and train DNNs for visual tracking. Using deep off-the-shelf features limits the tracking performance due to inconsistency among the objectives of different tasks. Also, offline learned deep features may not capture target variations and tend to over-fit on initial target templates. Hence, DNNs are trained on large-scale datasets to specialize the networks for visual tracking purposes. Besides, applying a fine-tuning process during visual tracking can adjust some network parameters and produce more refined target representations. However, the fine-tuning process is time-consuming and prone to over-fitting because of a heuristically fixed iteration number and limited available training data. As shown in Table~\ref{tab2} to Table~\ref{tab4}, these DL-based methods usually train a pre-trained network (i.e., backbone network) by offline training, online training, or both.
\vspace{-.3cm}
%----------------------------------------------------------------%
\subsection{Network Training} \label{Sec2.3}
The state-of-the-art DL-based visual tracking methods mostly exploit end-to-end learning with train/re-train a DNN by applying gradient-based optimization algorithms. However, these methods have differences according to their offline network training, online fine-tuning, computational complexity, dealing with lack of training data, addressing the overfitting problem, and exploiting unlabeled samples by unsupervised training. The network training sections in the previous review papers \cite{TrackingNoisyTargets,HandcraftedDeepTrackers,SurveyDeepTracking} consider both FENs and EENs, although the FENs were only pre-trained for other tasks, and there is no training procedure for visual tracking. In this survey, DL-based methods are categorized into only offline pre-training, only online training, and both offline and online training for visual tracking purposes. The training details of these methods are shown in Table~\ref{tab2} to Table~\ref{tab4}.
%----------------------------------------------------------------%
\subsubsection{\textbf{Training Datasets}} \label{Sec2.3.0}
Visual trackers employ diverse datasets to train their networks. These datasets are generally categorized into general-purpose \& tracking datasets (see Table~\ref{tab1} to Table~\ref{tab4}). A general-purpose dataset refers to a dataset from other tasks that provide desirable representations of different targets, e.g., object recognition or segmentation. It can include numerous datasets such as ImageNet \cite{ImageNet}, YouTube-VOS \cite{YT-VOS}, YouTube-BoundingBoxes \cite{YouTube-BB}, KITTI \cite{KITTI}, etc. That is, these datasets are used as auxiliary datasets in training procedures. However, tracking datasets are also utilized for training visual tracking networks. For instance, large-scale tracking datasets such as LaSOT \cite{LaSOT} \& TrackingNet \cite{TrackingNet} are explored in recent years. By exploring tracking datasets, the networks are trained on task-specific scenarios in the presence of challenging tracking attributes.
%----------------------------------------------------------------%
\subsubsection{\textbf{Only Offline Training}} \label{Sec2.3.1}
Most of the DL-based visual tracking methods only pre-train their networks to provide a generic target representation and reduce the high risk of over-fitting due to imbalanced training data and fixed assumptions. To adjust the learned filter weights for visual tracking task, the specialized networks are trained on large-scale data to exploit better representation and achieve acceptable tracking speed by preventing from training during visual tracking (see Table~\ref{tab2}).
%----------------------------------------------------------------%
\begin{table*}
\caption{Only offline training for visual tracking. The abbreviations are denoted as: confidence map (CM), saliency map (SM), bounding box (BB), object score (OS), feature maps (FM), segmentation mask (SGM), rotated bounding box (RBB), action (AC), deep appearance features (DAF), deep motion features (DMF), deeo optical flow (DOF).} % title of Table
\vspace{-2mm}
\centering % used for centering table
\resizebox{\textwidth}{!}{
\begin{tabular}{c c c c c c c c c} % centered columns (4 columns)
\hline\hline %inserts double horizontal lines
Method & Backbone network & Offline training dataset(s) & Exploited features & PC (CPU, RAM, Nvidia GPU) & Language & Framework & Speed (fps) & Tracking output \protect\\ 
% Method & Value & Parameter & Value \\ [0.5ex] % inserts table
%heading \protect\\ 
\hline\hline 
GOTURN \cite{GOTURN} & 	AlexNet	 & ILSVRC-DET, ALOV & 	DAF	 & N/A, GTX Titan X GPU	 & C/C++  & Caffe & 166	 & BB \protect\\ 
SiamFC \cite{SiamFC} & 	AlexNet & 	ImageNet, ILSVRC-VID & 	DAF	 & Intel I7-4790K 4.00GHz CPU, GTX Titan X GPU & 	Matlab &  MatConvNet & 	58	 & CM \protect\\ 
SINT \cite{SINT}	 & AlexNet, VGG-16 & 	ImageNet, ALOV & 	DAF	 & N/A & 	Matlab & 	 Caffe  & 	N/A	 & OS \protect\\ 
R-FCSN \cite{R-FCSN} & 	AlexNet & ImageNet, ILSVRC-VID & 	DAF	 & N/A, GTX Titan X GPU & 	Matlab &  MatConvNet & 	50.25 & 	CM \protect\\ 
LST \cite{LST} & 	AlexNet & 	ImageNet, ILSVRC-VID & 	DAF	 & Intel Xeon 3.50GHz CPU, GTX Titan X GPU & 	Matlab &  MatConvNet & 	~24	 & CM \protect\\ 
CFNet \cite{CFNet} & 	AlexNet	 & ImageNet, ILSVRC-VID	 & DAF & 	Intel I7 4.00GHz CPU, GTX Titan X GPU & 	Matlab &  MatConvNet & 	75 & 	CM \protect\\ 
DaSiamRPN \cite{DaSiamRPN} & 	AlexNet	 & ILSVRC, YTBB, Augmented ILSVRC-DET, Augmented MSCOCO-DET & 	DAF & 	Intel I7 CPU, 48GB RAM, GTX Titan X GPU & 	Python & 	PyTorch & 	160	 & CM \protect\\ 
StructSiam \cite{StructSiam} & 	AlexNet	 & ILSVRC-VID, ALOV	 & DAF & 	Intel I7-4790 3.60GHz, GTX 1080 GPU	 & Python &  TensorFlow & 	45 & 	CM \protect\\ 
Siam-BM \cite{Siam-BM}	 & AlexNet & 	ImageNet, ILSVRC-VID & 	DAF	 & Intel Xeon 2.60GHz CPU, Tesla P100 GPU & 	Python &  TensorFlow & 	48 & 	CM \protect\\ 
SA-Siam \cite{SA-Siam} & 	AlexNet	 & ImageNet, TC128, ILSVRC-VID & 	DAF & 	Intel Xeon 2.40GHz CPU, GTX Titan X GPU & 	Python & 	TensorFlow	 & 50 & 	CM \protect\\ 
SiamRPN \cite{SiamRPN} & 	AlexNet & 	ILSVRC, YTBB & 	DAF	 & Intel I7 CPU, 12GB RAM, GTX 1060 GPU	 & Python & 	PyTorch & 160	 & FM \protect\\ 
C-RPN \cite{CRPN} & 	AlexNet	 & ImageNet, ILSVRC-VID, YTBB & 	DAF & 	N/A, GTX 1080 GPU & 	Matlab &  MatConvNet & 	~36	 & CM \protect\\ 
GCT \cite{GCT}	 & AlexNet	 & ImageNet, ILSVRC-VID	 & DAF	 & Intel Xeon 3.00GHz CPU, 256GB RAM, GTX 1080Ti GPU & 	Python	 & TensorFlow	 & 49.8	 & CM \protect\\ 
GradNet \cite{GradNet} & AlexNet & ILSVRC-2014 & DAF	 & Intel I7 3.2GHz CPU, 32GB RAM, GTX 1080Ti GPU	 & Python  & TensorFlow & 80 & CM \protect\\ 
i-Siam \cite{i-Siam} & AlexNet & GOT-10k & DAF & Intel I7-7700K 4.20GHz CPU, Titan Xp GPU & N/A & N/A & 43 & CM \protect\\ 
UpdateNet \cite{UpdateNet} & AlexNet & LaSOT & 	DAF	& N/A & Python & PyTorch & N/A & CM \protect\\ 
SPM \cite{SPM-Tracker}	 & AlexNet, SiameseRPN, RelationNet	 & ImageNet, ILSVRC-VID, YTBB, ILSVRC-DET, MSCOCO, CityPerson, WiderFace	 & DAF & 	N/A, P100 GPU & 	N/A	 & N/A & 	120 & 	OS \protect\\ 
FICFNet \cite{FICFNet} & AlexNet	 & ImageNet, ILSVRC-VID	 & DAF & 	Intel I7 4.00GHz CPU, GTX Titan X GPU	 & Matlab &  MatConvNet & 	28	 & CM \protect\\ 
MTHCF \cite{MTHCF}	 & AlexNet	 & ImageNet, ILSVRC-VID	 & DAF & 	Intel 6700 3.40GHz CPU, GTX Titan GPU & 	Matlab &  MatConvNet & 	33	 & CM \protect\\ 
HP \cite{HP}	 & AlexNet	 & ImageNet, ILSVRC-VID	 & DAF & 	N/A	 & Python &  Keras & 	69 & 	CM \protect\\ 
EAST \cite{EAST} & 	AlexNet	 & ImageNet, ILSVRC-VID & 	DAF	 & Intel I7 4.00GHz CPU, GTX Titan X GPU & 	Matlab &  MatConvNet & 	23.2 & 	AC \protect\\ 
CFCF \cite{CFCF} & 	VGG-M & 	ImageNet, ILSVRC-VID, VOT2015 & 	HOG, DAF & 	Intel Xeon 3.00GHz CPU, Tesla K40 GPU	 & Matlab	& MatConvNet & 	~1.7 & 	CM \protect\\ 
CFSRL \cite{CFSRL} & 	VGG-M	 & ILSVRC-VID & 	DAF	 & Intel Xeon 2.40GHz CPU, 32GB RAM, GTX Titan X GPU	 & Matlab, Python & PyTorch & 	N/A	 & CM \protect\\ 
C2FT \cite{C2FT} & 	VGG-M & 	ImageNet, N/A	 & DAF & 	Intel Xeon 2.60GHz CPU, GTX 1080Ti GPU & 	N/A & 	N/A &	N/A	 & AC \protect\\ 
DSNet \cite{DSNet} & VGG-M & ILSVRC-2015 & 	DAF	 & Intel 6700K 4.0GHz CPU, GTX 1080 GPU & N/A & N/A & 68.5 & CM \protect\\ 
SRT \cite{SRT} & VGG-16	 & ImageNet, ALOV, Deform-SOT & 	DAF	 & N/A & 	N/A	 & N/A	 & N/A & 	BB \protect\\ 
IMLCF \cite{IMLCF} & 	VGG-16 & 	ImageNet, ILSVRC-VID & 	DAF	 & Intel 1.40GHz CPU, GTX 1080Ti GPU & 	Matlab	 & MatConvNet & 	N/A	 & CM \protect\\ 
SINT++ \cite{SINT++}	 & VGG-16 & 	ImageNet, OTB2013, OTB2015, VOT2014	 & DAF & 	Intel I7-6700K CPU, 32GB RAM, GTX 1080 GPU & Python & Caffe, Keras & 	N/A & 	AC \protect\\ 
MAM \cite{MAM} & VGG-16, Faster-RCNN & 	ImageNet, PASCAL VOC 2007, OTB100, TC128	& DAF	& 3.40GHz CPU, Titan GPU	 & Matlab & Caffe & 3 & SM \protect\\ 
PTAV \cite{PTAV-ICCV,PTAV} & 	VGGNet & 	ALOV & 	HOG, DAF & 	N/A, GTX Titan Z GPU & 	C/C++ & Caffe & 27 & CM \protect\\ 
UDT \cite{UDT} & VGGNet & 	ILSVRC	 & DAF & 	Intel I7-4790K 4.00GHz, GTX 1080Ti GPU & 	Matlab	 & MatConvNet & 	55	 & CM \protect\\ 
DRRL \cite{DRRL} & 	VGGNet	 & ImageNet, VOT2016 & 	DAF & 	N/A, GTX 1060 GPU & Python & TensorFlow & 	6.3	 & OS \protect\\ 
FCSFN \cite{FCSFN} & 	VGG-19	 & ImageNet, ALOV & 	DAF	 & N/A	 & N/A & 	N/A	 & N/A	 & CM \protect\\ 

Siam-MCF \cite{Siam-MCF} & 	ResNet-50	 & ImageNet, ILSVRC-VID	 & DAF & 	Intel Xeon E5 CPU, GTX 1080Ti GPU & 	Python &  TensorFlow & 	20 & 	CM \protect\\ 
SiamMask \cite{SiamMask} & 	ResNet-50 & 	ImageNet, MSCOCO, ILSVRC-VID, YouTube-VOS & 	DAF & 	N/A, RTX 2080 GPU & 	Python & 	PyTorch & 	55~60	 & SGM, RBB \protect\\
SiamRPN++ \cite{SiamRPN++} & 	ResNet-50	 & ImageNet, MSCOCO, ILSVRC-DET, ILSVRC-VID, YTBB & 	DAF	 & N/A, Titan Xp Pascal GPU & Python & 	PyTorch	 & 35 & 	OS, BB \protect\\ 
CGACD \cite{CGACD} & ResNet-50 & ILSVRC-VID, YTBB, GOT-10k, COCO, ILSVRC-DET & 	DAF	 & 3.5GHz CPU, RTX 2080Ti GPU	 & Python  & PyTorch & 70	 & BB \protect\\ 
CSA \cite{CSA} & ResNet-50 & GOT-10k & 	DAF	 & Intel I9 CPU, 64GB RAM, RTX 2080Ti GPU & Python & PyTorch & 100 & CM \protect\\ 
SiamAttn \cite{SiamAttn} & 	ResNet-50 & COCO, YouTube-VOS, LaSOT, TrackingNet & 	DAF	 & N/A, RTX 2080Ti GPU & Python  & PyTorch & 33 & BB, SGM \protect\\ 
SiamBAN \cite{SiamBAN} & ResNet-50 & ILSVRC-VID, YTBB, COCO, ILSVRC-DET, GOT-10k, LaSOT & 	DAF	 & Intel Xeon 4108 1.8GHz CPU, 64GB RAM, GTX 1080Ti GPU & Python & PyTorch & 40	 & BB \protect\\ 
SiamCAR \cite{SiamCAR} & ResNet-50 & ILSVRC-DET, COCO, ILSVRC-VID, YTBB & DAF & N/A, 4 RTX 2080Ti GPU	 & Python & PyTorch & 52.2 & BB \protect\\ 
SiamDW \cite{SiamDW}	 & ResNet, ResNeXt, Inception	 & ImageNet, ILSVRC-VID, YTBB & 	DAF & 	Intel Xeon 2.40GHz CPU, GTX 1080 GPU	 & Python & PyTorch & 	13~93 & 	CM, FM \protect\\ 
FlowTrack \cite{FlowTrack} & 	FlowNet	 & Flying chairs, Middlebur, KITTI, Sintel, ILSVRC-VID & 	DAF, DMF & 	Intel I7-6700 CPU, 48GB RAM, GTX Titan X GPU	 & Matlab &  MatConvNet & 	12 & 	CM \protect\\ 
RASNet \cite{RASNet} & Attention networks	 & ILSVRC-DET & 	DAF	 & Intel Xeon 2.20GHz CPU, Titan Xp Pascal GPU	 & Matlab	 & MatConvNet & 83 & 	CM \protect\\ 
ACFN \cite{ACFN} & 	Attentional correlation filter network & OTB2013, OTB2015, VOT2014, VOT2015	 & HOG, Color, DAF & 	Intel I7-6900K 3.20GHz CPU, 32GB RAM, GTX 1070 GPU	 & Matlab, Python & MatConvNet, TensorFlow & 	15 & 	OS \protect\\ 
RFL \cite{RFL} & Convolutional LSTM	 & ILSVRC-VID & 	DAF	 & Intel I7-6700 3.40GHz CPU, GTX 1080 GPU & 	Python &  TensorFlow & 	~15  &	CM \protect\\ 
DRLT \cite{DRLT} & 	YOLO & 	ImageNet, PASCAL VOC & 	DAF	 & N/A, GTX 1080 GPU & 	Python & TensorFlow & ~45 & 	BB \protect\\ 
TGGAN \cite{TGGAN}	 & - &  	ALOV, VOT2015 & 	DAF	 & N/A, GTX Titan X GPU	 & Python &  Keras & 3.1 & CM \protect\\ 
DCTN \cite{DCTN}	 & - & 	TC128, NUS-PRO, MOT2015	 & DAF, DMF & 	N/A & 	N/A & 	N/A	 & 27 & CM \protect\\ 
YCNN \cite{YCNN}	 & - & 	ImageNet, ALOV300++	 & DAF & 	N/A, Tesla K40c GPU	 & Python &  TensorFlow & 	45 & 	CM \protect\\ 
RDT \cite{RDT}	 & - & 	VOT2015 & 	DAF & 	Intel I7-4790K 4.00GHz, 24GB RAM, GTX Titan X GPU & Python &  TensorFlow & 43 & CM, OS \protect\\ 
%----------------------------------------------
SiamRCNN \cite{SiamRCNN} & 	Faster R-CNN, ResNet-101-FPN & ILSVRC-VID, COCO, YouTube-VOS, GOT-10k, LaSOT & DAF & N/A, Tesla V100 GPU & Python  & TensorFlow & 4.7 & BB, SGM \protect\\ 
FGTrack \cite{FGTrack} & ResNet-18, FlowNet2	 & ILSVRC-VID, TrackingNet, YouTube-VOS & DAF, DOF & N/A, GTX 1080Ti GPU & Python & PyTorch & 19.6 & BB \protect\\ 
CRVFL \cite{CRVFL} & - & N/A & 	DAF	 & Intel I7 CPU & N/A & N/A & 1.5-2 & OS \protect\\ 
VTCNN \cite{VTCNN} & - & N/A & 	DAF	 & Intel I7 3.4GHz	 & Matlab & N/A & 5.5 & OS \protect\\ 
GlobalTrack \cite{GlobalTrack} & Faster R-CNN, ResNet-50 & COCO, GOT-10k, LaSOT & DAF & N/A, Titan X GPU & Python  & PyTorch & 6 & BB, SGM \protect\\ 
SPLT \cite{SPLT} & MobileNet-v1, ResNet-50 & ILSVRC-VID, ILSVRC-DET & DAF & Inter I7 CPU, 32GB RAM, GTX 1080Ti GPU & Python  & TensorFlow, Keras & 25.7 & BB, OS \protect\\ 
  \hline %inserts single line
  \end{tabular}
  \label{tab2}
 }
\vspace{-4mm}
\end{table*}
%----------------------------------------------------------------%
\begin{table*}
\caption{Only online training for visual tracking. The abbreviations are denoted as: confidence map (CM), bounding box (BB), object score (OS), deep appearance features (DAF), action (AC).} % title of Table
\vspace{-2mm}
\centering % used for centering table
\resizebox{\textwidth}{!}{
\begin{tabular}{c c c c c c c c c} % centered columns (4 columns)
\hline\hline %inserts double horizontal lines
Method & Backbone network & Exploited features & Strategy to alleviate the over-fitting problem & PC (CPU, RAM, Nvidia GPU) & Language & Framework & Speed (fps) & Tracking output \protect\\ 
% Method & Value & Parameter & Value \\ [0.5ex] % inserts table
%heading \protect\\ 
\hline \hline
SMART \cite{SMART} &	ZFNet &	DAF &	Set the learning rates in conv1-conv3 to zero &	Intel 3.10GHz CPU, 256 GB RAM, GTX Titan X GPU &	Matlab &	Caffe &	~27 &	CM \protect\\ 
TCNN \cite{TCNN} &	VGG-M &	DAF &	Only update fully-connected layers &	Intel I7-5820K 3.30GHz CPU, GTX Titan X GPU &	Matlab &	MatConvNet &	~1.5 &	OS  \protect\\ 
C2FT \cite{C2FT} &	VGG-M &	DAF & Coarse-to-fine localization &	Intel Xeon E5-2670 2.60GHz, GTX 1080Ti GPU & N/A &	N/A & N/A &	AC \protect\\ 
TSN \cite{TSN} &	VGG-16 &	DAF &	Coarse-to-fine framework &	N/A, GTX 980Ti GPU &	Matlab &	MatConvNet &	~1 &	CM \protect\\ 
DNT \cite{DNT} &	VGG-16 &	DAF &	Set uniform weight decay in the objective functions	 & 3.40GHz CPU, GTX Titan GPU &	Matlab &	Caffe &	5 &	CM \protect\\ 
DSLT \cite{DSLT} &	VGG-16 &	DAF &	Use seven last frames for model update &	Intel I7 4.00GHz CPU, GTX Titan X GPU &	Matlab &	Caffe &	5.7 &	CM \protect\\ 
LSART \cite{LSART} & VGG-16 &	DAF	 & Two-stream training network to learn network parameters &	Intel 4.00GHz CPU, 32GB RAM, GTX Titan X GPU &	Matlab &	Caffe &	~1 &	CM \protect\\ 
adaDDCF \cite{adaDDCF} &	VGG-16 &	DAF &	Regularization item for training of each layer &	3.40GHz CPU, Tesla K40 GPU	 & Matlab &	MatConvNet &	9 &	CM \protect\\ 
HSTC \cite{HSTC} &	VGG-16 &	DAF &	Dropout layer and convolutional with the mask layer	 & Intel Xeon 2.10GHz CPU, GTX 1080 GPU &	Matlab &	Caffe &	2.1 &	CM \protect\\ 
P-Track \cite{P-Track} &	VGG-16 &	DAF &	Learning policy for update and re-initialization  &	N/A, Tesla K40 GPU &	N/A &	N/A &	~10 &	CM \protect\\ 
OSAA \cite{OSAA} &	ResNet-50 or MobileNet-v2 &	DAF & - &	N/A, Tesla V100 GPU & Python &	PyTorch &	N/A &	BB \protect\\ 
STCT &	Custom &	DAF &	Sequential training method & 3.40GHz CPU, GTX Titan GPU &	Matlab &	Caffe &	2.5 &	CM \protect\\ 
DeepTrack \cite{DeepTrack} &	Custom &	DAF &	Temporal sampling mechanism for the batch generation in SGD algorithm &	Quad-core CPU, GTX 980 GPU &	Matlab &	N/A &	2.5 &	OS \protect\\ 
CNT \cite{CNT} &	Custom &	DAF &	Incremental update scheme &	Intel I7-3770 3.40GHz CPU, GPU &	Matlab &	N/A &	5 &	BB \protect\\ 
RDLT \cite{RDLT} &	Custom &	DAF	 & Build relationship between the stable factor and iteration number &	Intel I7 2.20GHz CPU &	Matlab &	N/A &	~5 &	CM \protect\\ 
P2T \cite{P2T} &	Custom &	DAF &	Generate large scale of part pairs in each mini-batch &	Intel I7-4790 3.60GHz CPU, 32GB RAM, GTX 980 GPU &	Matlab &	Caffe &	~2 &	BB \protect\\ 
AEPCF \cite{AEPCF} &	Custom &	DAF &	Select a proper learning rate &	Intel I7 3.40GHz, 32GB RAM, GPU &	N/A &	N/A &	4.15 &	CM \protect\\ 
FRPN2T-Siam \cite{FRPN2T-siam} &	Custom & DAF &	Only update fully-connected layers &	N/A &	Matlab &	Caffe &	N/A &	CM \protect\\ 

RLS \cite{RLS} & Custom &	DAF &	Recursive LSE-aided online learning method &	N/A & Python &	N/A &	N/A &	OS \protect\\

  \hline %inserts single line
  \end{tabular}
  \label{tab3}
 }
\vspace{-4mm}
\end{table*}
%----------------------------------------------------------------%
\begin{table*}
\caption{Both offline and online training for visual tracking. The abbreviations are denoted as: confidence map (CM), bounding box (BB), rotated bounding box (RBB), object score (OS), voting map (VM), action (AC), segmentation mask (SGM), deep appearance features (DAF), deep motion features (DMF), compressed deep appearance features (CDAF).} % title of Table
\vspace{-2mm}
\centering % used for centering table
\resizebox{\textwidth}{!}{
\begin{tabular}{c c c c c c c c c c} % centered columns (4 columns)
\hline\hline %inserts double horizontal lines
Method & Backbone network & Offline training(s) & Online network training & Exploited features  & PC (CPU, RAM, Nvidia GPU) & Language & Framework & Speed (fps) & Tracking output \protect\\ 
% Method & Value & Parameter & Value \\ [0.5ex] % inserts table
%heading \protect\\ 
\hline\hline 
DRN \cite{DRN} &	AlexNet &	ImageNet &	Yes &	DAF &	N/A, K20 GPU &	Matlab &	Caffe &	1.3 &	CM \protect\\ 
DSiam/DSiamM \cite{DSiam} &	AlexNet, VGG-19 &	ImageNet, ILSVRC-VID &	Yes &	DAF &	N/A, GTX Titan X GPU &	Matlab &	MatConvNet &	45 &	CM \protect\\ 
TripletLoss \cite{Tripletloss} &	AlexNet &	ImageNet, ILSVRC-VID, ILSVRC &	Dependent &	DAF &	Intel I7-6700 3.40GHz CPU, GTX 1080Ti GPU &	Matlab &	MatConvNet &	55~86 &	CM \protect\\ 
MM \cite{MM} &	AlexNet &	ImageNet, OTB2015, ILSVRC &	Yes &	DAF &	Intel I7-6700 4.00GHz CPU, 16GB RAM, GTX 1060 GPU &	Matlab &	MatConvNet &	1.2 &	OS \protect\\ 
TAAT \cite{TAAT} &	AlexNet, VGGNet, ResNet &	ImageNet, ALOV, ILSVRC-VID &	Yes &	DAF &	Intel Xeon 1.60GHz CPU, 16GB RAM, GTX Titan X GPU &	Matlab &	Caffe &	15 &	BB \protect\\ 
DPST \cite{DPST} &	VGG-M &	ImageNet, ILSVRC-VID, ALOV &	Only on the first frame &	DAF &	Intel I7 3.60GHz CPU, GTX 1080Ti GPU &	Matlab &	MatConvNet &	~1 &	OS \protect\\ 
MDNet \cite{MDNet} &	VGG-M &	ImageNet, OTB2015, ILSVRC-VID &	Yes &	DAF &	Intel Xeon 2.20GHz CPU, Tesla K20m GPU &	Matlab &	MatConvNet &	~1 &	OS \protect\\ 
GNet \cite{GNet} &	VGG-M &	ImageNet, VOT &	Yes &	DAF &	Intel Xeon 2.66GHz CPU, Tesla K40 GPU &	Matlab &	MatConvNet &	1 &	OS \protect\\ 
BranchOut \cite{BranchOut} &	VGG-M &	ImageNet, OTB2015, ILSVRC &	Yes &	DAF &	N/A &	Matlab &	MatConvNet &	N/A & OS \protect\\ 
SANet \cite{SANet} &	VGG-M &	ImageNet, OTB2015, ILSVRC &	Yes &	DAF	 & Intel I7 3.70GHz CPU, GTX Titan Z GPU &	Matlab &	MatConvNet &	~1 &	OS \protect\\ 
RT-MDNet \cite{RT-MDNet} &	VGG-M &	ImageNet, ILSVRC-VID &	Yes &	DAF &	Intel I7-6850K 3.60GHz, Titan Xp Pascal GPU &	Python &	PyTorch &	46 &	OS \protect\\ 
TRACA \cite{TRACA} &	VGG-M &	ImageNet, PASCAL VOC &	Yes &	CDAF &	Intel I7-2700K 3.50GHz CPU, 16GB RAM, GTX 1080 GPU &	Matlab &	MatConvNet &	101.3 &	CM \protect\\ 
VITAL \cite{VITAL} &	VGG-M &	ImageNet, OTB2015, ILSVRC &	Yes &	DAF &	Intel I7 3.60GHz CPU, Tesla K40c GPU &	Matlab &	MatConvNet &	1.5 &	OS \protect\\ 
DAT \cite{DAT} &	VGG-M &	ImageNet, OTB2015, ILSVRC &	Yes &	DAF & Intel I7-3.40GHz CPU, GTX 1080 GPU &	Python &	PyTorch &	1 &	CM \protect\\ 
ACT \cite{ACT} &	VGG-M &	ImageNet-Video, ILSVRC &	Yes &	DAF &	3.40GHz CPU, 32GB RAM, GTX Titan GPU &	Python &	PyTorch &	30 &	OS \protect\\ 
MGNet \cite{MGNet} &	VGG-M &	ImageNet, OTB2015, ILSVRC &	Yes	 & DAF, DMF &	Intel I7-5930K 3.50GHz CPU, GTX Titan X GPU &	Matlab &	MatConvNet &	~2 &	OS \protect\\ 
DRL-IS \cite{DRL-IS} &	VGG-M &	ImageNet, VOT2013~2015 &	Yes &	DAF &	Intel I7 3.40GHz CPU, 24GB RAM, GTX 1080Ti GPU &	Python &	PyTorch &	10.2 &	AC \protect\\ 
ADNet \cite{ADNet-CVPR,ADNet-TNNLS} &	VGG-M &	ImageNet, VOT2013~2015, ALOV &	Yes &	DAF &	Intel I7-4790K, 32GB RAM, GTX Titan X GPU &	Matlab &	MatConvNet &	15 &	AC, OS \protect\\ 
FMFT \cite{FMFT} &	VGG-16 &	ImageNet &	Yes &	DAF &	Intel Xeon 3.50GHz CPU, GTX Titan X GPU &	Matlab &	MatConvNet &	N/A	 & CM \protect\\ 
DET \cite{DET} &	VGG-16 &	ImageNet, ALOV, VOT2014, VOT2015 &	Yes &	DAF &	Intel I7-4790 3.60GHz CPU, GTX Titan X GPU &	Python &	Keras &	3.4 &	OS \protect\\ 
DCFNet/DCFNet2 \cite{DCFNet} &	VGGNet &	ImageNet, TC128, UAV123, NUS-PRO &	Yes &	DAF &	Intel Xeon 2.40GHz CPU, GTX 1080 GPU &	Matlab &	MatConvNet &	65 &	CM \protect\\ 
STP \cite{STP} &	VGGNet &	ImageNet &	Yes &	Votes &	N/A, GTX Titan X GPU &	Python &	PyTorch &	4 &	VM \protect\\ 
MRCNN \cite{MRCNN} &	VGGNet &	ImageNet, VOT2015 &	Yes &	DAF &	Intel I7 3.50GHz CPU, GTX 1080 GPU &	Matlab &	MatConvNet &	~1.2 &	CM \protect\\ 
CODA \cite{CODA} &	VGG-19, SSD &	ImageNet, VOT2013, VOT2014, VOT2015 &	Yes &	DAF &	Intel I7-4770K CPU, 32GB RAM, GTX 1070 GPU &	Matlab &	Caffe &	34.8 &	CM \protect\\ 
ATOM \cite{ATOM} &	ResNet-18, IoU-Nets &	ImageNet, COCO, LaSOT, TrackingNet &	Yes &	DAF &	N/A, GTX 1080 GPU &	Python &	PyTorch &	30 &	CM \protect\\ 

D3S \cite{D3S} &	ResNet-50 &	Youtube-VOS &	Yes & DAF &	N/A, GTX 1080 GPU &	Python & PyTorch & 25 &	RBB, SGM \protect\\ 
MetaUpdater \cite{MetaUpdater} &	ResNet-50 &	ImageNet, LaSOT  &	Yes &	DAF &	Intel I9 CPU, 64GB RAM, GTX 2080Ti GPU &	Python &	TensorFlow &	13 &	CM \protect\\ 
CRAC \cite{CRAC} &	ResNet-50 &	ImageNet, KITTI, VisDrone-2018 & Yes &	DAF &	N/A &	Python &	PyTorch, MatConvNet &	56 &	OS \protect\\ 
COMET \cite{COMET} &	ResNet-50 &	ImageNet, LaSOT, GOT-10K, NfS, VisDrone-2019 &	Yes &	DAF &	N/A, Tesla V100 GPU &	Python & PyTorch &	24 &	CM \protect\\
FGLT \cite{FGLT} &	VGG-M, ResNet-50, PWC-Net &	ImageNet, ILSVRC-VID, COCO, ILSVRC-DET, YTBB, FlyingChairs, FlyingThings3D &	Yes &	DAF &	Intel Xeon 3.50GHz CPU, GTX 1080Ti GPU & Python & PyTorch &	N/A & CM, BB \protect\\ 
LRVN \cite{LRVN} &	VGG-M, MobileNet &	ImageNet, ILSVRC-VID, ILSVRC-DET &	Yes &	DAF &	Intel I7 CPU, 32GB RAM, GTX Titan X GPU &	Python &	TensorFlow &	2.7 &	BB, OS \protect\\ 
UCT/UCT-Lite \cite{UCT} &	ResNet101 &	ImageNet, TC128, UAV123 &	Only on the first frame &	DAF &	Intel I7-6700 CPU, 48GB RAM, GTX Titan X GPU &	Matlab &	Caffe &	41 &	CM \protect\\ 
FPRNet \cite{FPRNet} &	ResNet-101, FlowNet &	ImageNet, ILSVRC, SceneFlow &	Yes &	DAF, DMF &	N/A	 &  Matlab  & Caffe &	N/A &	BB \protect\\ 
ADT \cite{ADT} &	- &	ImageNet, ALOV300++, UAV123, NUS-PRO &	Only on the first frame &	DAF &	Intel 2.40GHz CPU, GTX TITAN X GPU &	Python & TensorFlow & 7 &	CM \protect\\ 
%--------------------------------------------------
DiMP \cite{DiMP} & 	ResNet-18, ResNet-50 & 	ImageNet, TrackingNet, LaSOT, GOT10k, COCO &  Meta-learning & DAF	 & N/A, GTX 1080 GPU & Python & PyTorch & 43~57 & OS \protect\\ 

PrDiMP \cite{PrDiMP} &	ResNet-18 or ResNet-50 & ImageNet, LaSOT, GOT-10k, TrackingNet, COCO &	Meta-learning &	DAF &	N/A &	Python & PyTorch &	30~40 &	CM \protect\\ 

BGBDT \cite{BGBDT} & SSD or FasterRCNN &	ImageNet, COCO, GOT-10k &	Meta-learning &	DAF &	N/A, GTX 1080 GPU &	Python & PyTorch &	3~10 & BB \protect\\ 

MLT \cite{MLT} &	AlexNet &	ImageNet, ILSVRC-2015, ILSVRC-2017 &	Meta-learning &	DAF &	Intel I7-4790K 4.0GHz CPU, 32GB RAM, GTX Titan X GPU &	Python &	TensorFlow &	48.1 &	CM \protect\\ 

ROAM \cite{ROAM} &	VGG-16 &	ImageNet, ILSVRC-VID, ILSVRC-DET, TrackingNet, LaSOT, GOT-10k, COCO &	Meta-learning &	DAF &	Intel I9 3.6GHz CPU, 4 RTX 2080 GPU &	Python & PyTorch &	13 &	CM \protect\\ 

TMAML \cite{TMAML} &	ResNet-18 with RetinaNet or FCOS &	ImageNet, COCO, GOT-10k, TrackingNet, LaSOT &	Meta-learning &	DAF &	N/A, P100 GPU &	Python & N/A &	40 & BB \protect\\ 

Meta-Tracker \cite{Meta-Tracker} &	dependent &	ImageNet, ILSVRC-DET, VOT-2013, VOT-2014, VOT-2015 & Meta-learning &	DAF &	N/A, GTX Titan X GPU &	Python & PyTorch &	dependent &	dependent \protect\\ 
  \hline %inserts single line
  \end{tabular}
  \label{tab4}
 }
\vspace{-4mm}
\end{table*}
%----------------------------------------------------------------%
\subsubsection{\textbf{Only Online Training}} \label{Sec2.3.2}
To discriminate unseen targets that may consider as the target in evaluation videos, some DL-based visual tracking methods use online training of whole or a part of DNNs to adapt network parameters according to the large variety of target appearance. Because of the time-consuming process of offline training on large-scale training data and insufficient discrimination of pre-trained models for representing tracking particular targets, the methods shown in Table~\ref{tab3} use directly training of DNNs and inference process alternatively online. However, these methods usually exploit some strategies to prevent over-fitting problem and divergence. 
%----------------------------------------------------------------%
\subsubsection{\textbf{Both Offline and Online Training}} \label{Sec2.3.3}
To exploit the maximum capacity of DNNs for visual tracking, the methods shown in Table~\ref{tab4} use both offline and online training. The offline and online learned features are known as shared and domain-specific representations, which majorly can discriminate the target from foreground information or intra-class distractors, respectively. Because visual tracking is a hard and challenging problem, the DL-based visual trackers attempt to simultaneously employ feature transferability and online domain adaptation.
%----------------------------------------------------------------%
\subsubsection{\textbf{Data Augmentation}} \label{Sec2.3.4}
Data augmentation comprises a set of techniques for increasing training samples' size to improve data quality and avoid the over-fitting problem. Visual trackers broadly employ these techniques based on the few-data regime of this task (see Table~\ref{tab9_DataAug}). The geometric transformations \& color space augmentations are vastly exploited for visual tracking. However, other algorithms, such as employing GANs \cite{VITAL}, can effectively impact tracking performance by capturing various appearance changes.
%----------------------------------------------------------------%
\begin{table}
\caption{Data Augmentations for Visual Tracking Methods.} % title of Table
\vspace{-2mm}
\centering % used for centering table
\resizebox{9cm}{!}{
\begin{tabular}{c c} % centered columns (4 columns)
\hline\hline %inserts double horizontal lines
Method & Augmentations \protect\\ 
% Method & Value & Parameter & Value \\ [0.5ex] % inserts table
%heading \protect\\ 
\hline \hline
GOTURN \cite{GOTURN} & Motion model, random crops  \protect\\ 
MDNet \cite{MDNet} & Multiple positive examples around the target  \protect\\ 
DeepTrack \cite{DeepTrack_BMVC,DeepTrack} & Horizontal flip  \protect\\ 
RFL \cite{RFL} & Random color distortion, translation, stretching  \protect\\ 
VRCPF \cite{VRCPF} & Annotated face ROIs using PASCAL VOC 2007   \protect\\ 
DNT \cite{DNT} & Center-shifted random patches, translation schemes   \protect\\ 
UPDT \cite{UPDT} & Flip, rotation, shift, blur, dropout  \protect\\ 
DaSiamRPN \cite{DaSiamRPN} & Translation, scale variations and illumination changes, motion blur \protect\\
STP \cite{STP} & Randomly shifted pairs \protect\\
Siam-MCF \cite{Siam-MCF} & Random cropping, color distortion, horizontal flipping, small resizing perturbations \protect\\
TRACA \cite{TRACA} & Blur, flip  \protect\\
VITAL \cite{VITAL} & Randomly masks \protect\\
SiamRPN \cite{SiamRPN} & Affine transformation \protect\\
YCNN \cite{YCNN} & Rotation, translation, illumination variation, mosaic, salt \& pepper noise \protect\\
DeepFWDCF \cite{DeepFWDCF} & Gray-scale rotation invariant LBP histograms  \protect\\
ORHF \cite{ORHF} & Augmentation of negative samples  \protect\\
ATOM \cite{ATOM} & Translation, rotation, blur, dropout, flip, color jittering \protect\\
MAM \cite{MAM} & All detected windows from target category \protect\\
DiMP50 \cite{DiMP} & Translation, rotation, blur, dropout, flip, color jittering  \protect\\
PrDiMP50 \cite{PrDiMP} & Translation, rotation, blur, dropout, flip, color jittering  \protect\\
TAAT \cite{TAAT} & Spatial \& temporal pairs  \protect\\
CGACD \cite{CGACD} & RoI augmentation  \protect\\ 
MLT \cite{MLT} & Horizontal flip, noise, Gaussian blur, translation \protect\\ 
ROAM \cite{ROAM} & Stretching and scaling the images \protect\\ 
SiamRCNN \cite{SiamRCNN} & Motion blur, gray-scale, gamma, flip, and scale augmentations \protect\\ 
TMAML \cite{TMAML} & Random scaling, shifting, zoom in/out \protect\\ 

\hline %inserts single line
\end{tabular}
\label{tab9_DataAug}
}
\vspace{-4mm}
\end{table}
%----------------------------------------------------------------%
\subsubsection{\textbf{Meta-Learning}} \label{Sec2.3.5}
As a well-known paradigm in machine learning, meta-learning \cite{Meta-Survey} (an alternative for data augmentation) has provided promising results for visual tracking task. Generally, it aims to provide experience on several learning tasks and use them to improve the performance of a new task. Inspired by \textit{model-agnostic meta-learning} (MAML) \cite{MAML}, visual trackers mainly seek to exploit meta-learners for constructing more flexible target models regarding unseen targets/scenarios (see Table~\ref{tab4}). For instance, it can be leveraged into the initialization \cite{Meta-Tracker,TMAML}, fast model adaptation \cite{BGBDT,MLT}, or model update \cite{ROAM,MetaUpdater} procedures of visual trackers. However, some visual trackers \cite{DiMP,PrDiMP} exploit meta-learning ideas to adjust their model weights during tracking, which is different from the classic meta-learning definition.
\vspace{-.2cm}
%----------------------------------------------------------------%
\subsection{Network Objective} \label{Sec2.4}
For the training and inference stages, DL-based visual trackers localize the given target based on network objective function. Hence, these methods are categorized into classification-based, regression-based, or both classification and regression-based methods as follows. This sub-section does not include the methods that exploit deep off-the-shelf features because these methods do not design and train the networks and usually employ pre-trained DNNs for feature extraction. 
%----------------------------------------------------------------%
\subsubsection{\textbf{Classification-based Objective Function}} \label{Sec2.4.1}
Motivated by other computer vision tasks such as image detection, classification-based visual tracking methods employ object proposal methods to produce hundreds of candidate/proposal BBs extracted from the search region. These methods aim to select the high score proposal by classifying the proposals to the target and background classes. This two-class (or binary) classification involves visual targets from various class and moving patterns and individual sequences, including challenging scenarios. Due to the main attention of these methods on inter-class classification, tracking a visual target in the presence of the same labeled targets is intensely prone to drift-problem. Also, tracking the arbitrary appearance of targets may lead to recognizing different targets with varying appearances. Therefore, the performance of the classification-based visual tracking methods is also related to their object proposal method, which usually produces a considerable number of candidate BBs. On the other side, some recent DL-based trackers utilize this objective function to take optimal actions \cite{DRRL,ADNet-CVPR,ADNet-TNNLS,C2FT,DRL-IS,EAST,SINT++,C2FT}.
%----------------------------------------------------------------%
\subsubsection{\textbf{Regression-based Objective Function}} \label{Sec2.4.2}
Due to the continuous instinct of estimation space of visual tracking, regression-based methods usually aim to directly localize the target in the subsequent frames by minimizing a regularized least-squares function. Generally, extensive training data are needed to train these methods effectively. The primary goal of regression-based methods is to refine the formulation of L2 or L1 loss functions, such as utilizing shrinkage loss in the learning procedure \cite{DSLT}, modeling both regression coefficients and patch reliability to optimize a neural network efficiently \cite{LSART}, or applying a cost-sensitive loss to enhance unsupervised learning performance \cite{UDT}. Meanwhile, recent visual trackers define a loss function for BB regression (e.g., \cite{CGACD,ROAM,FGTrack}) to provide accurate localization. 
%----------------------------------------------------------------%
\subsubsection{\textbf{Classification- and Regression-based Objective Function}} \label{Sec2.4.3}
To take advantages of both foreground-background/category classification and ridge regression (i.e., regularized least-squares objective function), a broad range of trackers employ both classification- and regression-based objective functions for visual tracking (see Fig.~\ref{fig:Taxonomy}), which their goal is to bridge the gap between the recent tracking-by-detection and continuous localization process of visual tracking. These methods commonly utilize classification-based methods to find the most similar object proposal to target, and then the estimated region will be refined by a BB regression method \cite{MDNet,TCNN,SRT,SANet,ACT,DaSiamRPN,RT-MDNet,SiamRPN,DAT,MGNet,SiamRPN++,MRCNN,TAAT}. The target regions are estimated by classification scores and optimized regression/matching functions \cite{FMFT,CFSRL,P2T,ATOM,CRPN,SPM-Tracker,SiamDW,SiamMask,DiMP,ADT,SMART,DRL-IS,COMET} to enhance efficiency and accuracy. The classification outputs are mainly inferred for candidate proposals' confidence scores, foreground detection, candidate window response, actions, and so forth.
\vspace{-.25cm}
%----------------------------------------------------------------%
\subsection{Network Output} \label{Sec2.5}
According to the network objective, the DL-based methods generate different network outputs to estimate or refine the estimated target location. Based on their network outputs, the DL-based methods are classified into six main categories (see Fig.~\ref{fig:Taxonomy} and Table~\ref{tab2} to Table~\ref{tab4}), namely confidence map (also includes score map, response map, and voting map), BB (also includes rotated BB), object score (also includes the probability of object proposal, verification score, similarity score, and layer-wise score), action, feature maps, and segmentation mask. Besides template-based methods, segmentation-based trackers have been explored to boost tracking performance. Despite initial works \cite{SegmentTracker_JOTS,SegmentTracker_DecisionTree,SegmentTracker_SuperPixel}, employing independent deep networks for tracking \& VOS may lead to irretrievable tracking failures and high computational complexity. Thus, segmentation-based trackers aim to jointly track and segment visual targets by adding a segmentation branch to the network \cite{FMFT,SiamMask,D3S,SiamAttn} or off-the-shelf \textit{BB-to-segmentation} (Box2Seg) networks \cite{Box2Seg1,Box2Seg2} to the base tracking network.
\vspace{-.5cm}
%----------------------------------------------------------------%
\subsection{Exploitation of Correlation Filters Advantages} \label{Sec2.6}
The DCF-based methods aim to learn a set of discriminative filters that an element-wise multiplication of them with a set of training samples in the frequency domain determines spatial target location. Since DCF has provided competitive tracking performance and computational efficiency compared to sophisticated techniques, DL-based visual trackers use correlation filter advantages. These methods are categorized based on how they exploit DCF advantages by using either a whole DCF framework or some benefits, such as its objective function or correlation filters/layers. Considerable visual tracking methods are based on integrating deep features in the DCF framework (see Fig.~\ref{fig:Taxonomy}). These methods aim to improve the robustness of target representation against challenging attributes, while other methods attempt to benefit the computational efficiency of correlation filter(s) \cite{CFNet}, correlation layer(s) \cite{FlowTrack,adaDDCF,FICFNet,TADT,MTHCF,DSNet}, and the objective function of correlation filters \cite{UCT,DSiam,DCFNet,RASNet,ATOM,UDT}.
\vspace{-.2cm}
%----------------------------------------------------------------%
\subsection{Aerial-view Tracking} \label{Sec2.7}
By pervasive applications of flying robots, tracking from aerial views introduces extra attractive challenges, such as tiny objects, weather conditions, dense environments, long occlusions, significant viewpoint change, etc. Aerial-view trackers can be classified into class-specific \& class-agnostic methods. 
Class-specific trackers mostly focus on human or vehicle classes, such as the \textit{unified contextual relation actor-critic} (CRAC) \cite{CRAC}, a GAN-based vehicle tracker that aims to model contextual relation and transfer the ground-view features to the aerial-ones. In contrast, class-agnostic trackers can track arbitrary classes of targets. Some DCF-based trackers \cite{BEVT,KAOT_ICRA,KAOT_TMM,MKCT,SASR} were the first generation of flying robot trackers, which address the inherent limitations of correlation filters (e.g., boundary effect and filter corruption) given aerial view conditions. However, the \textit{coarse-to-fine tracker} (C2FT) \cite{C2FT} employs deep RL coarse- \& fine-trackers (for estimating entire BB and its refinement) to address significant aspect-ratio change of targets from aerial-views. Finally, the \textit{context-aware IoU-guided network for small object tracking} (COMET) aims to narrow the performance gap between the aerial-view trackers \& state-of-the-art ones. It employs an offline proposal generation strategy \& a multitask two-stream network to exploit context information and handle out-of-view \& occlusion effectively.
\vspace{-.2cm}
%----------------------------------------------------------------%
\subsection{Long-term Tracking} \label{Sec2.8}
Long-term tracking performs on more realistic scenarios, including (relatively) long videos in which targets may disappear \& reappear. Despite the close relationship to practical applications, limited trackers have been proposed for this task. One way is the extension of a short-term tracker by various strategies. For instance, DaSiamRPN \cite{DaSiamRPN} uses a local-to-global search region strategy to handle out-of-view \& full occlusion, while LCTdeep \cite{LCTdeep} utilizes a detection module with incremental updates. The FGLT \cite{FGLT} employs both MDNet \cite{MDNet} \& SiamRPN++ \cite{SiamRPN++} trackers for the tracking result judgment, and a detection module modifies tracking failures. The \textit{memory model via the Siamese network for long-term tracking} (MMLT) \cite{MMLT} modifies the SiamFC tracker \cite{SiamFC} by re-detection and memory management parts. Moreover, the \textit{improved Siamese tracker} (i-Siam) \cite{i-Siam} revisits the SiamFC tracker via a negative signal suppression approach \& a diverse multi-template one. The multi-level CF-based tracker \cite{MLCFT} employs an oriented re-detection technique, while MetaUpdater \cite{MetaUpdater} exploits a SiamRPN-based re-detector and an online verifier with a meta-updater. Finally, the SPLT tracker \cite{SPLT} is based on SiamRPN \cite{SiamRPN} and comprises perusal and skimming modules for local tracking \& search window selection. \\
\indent On the other hand, various long-term trackers have been inspired by successful detection methods. For instance, GlobalTrack \cite{GlobalTrack} is a two-stage tracker, including a query-guided region proposal network \& query-guided region CNN to generate object candidates and produce the final predictions, respectively. Also, the LRVN tracker \cite{LRVN} consists of the combination of an offline-learned regression network with an online-updated verification network to generate target candidates \& evaluate/update them on reliable observations. Lastly, the SiamRCNN tracker \cite{SiamRCNN} introduces a hard example mining procedure and tracklet dynamic programming algorithm to detect potential distractors \& select the best target at each time-step.
\vspace{-.2cm}
%----------------------------------------------------------------%
\subsection{Online Tracking} \label{Sec2.9}
While initial DL-based methods had focused on their performance, recent trackers aim to be accurate, robust, and efficient simultaneously. According to Fig.~\ref{fig:Taxonomy}, a wide variety of algorithms are classified as online trackers regarding different hardware implementations (Table~\ref{tab1}-Table~\ref{tab4}). Most of these trackers (e.g., \cite{SINT,GOTURN,SiamFC,SiamRPN,SiamRPN++}) are based on offline-trained SNNs that do not update the target model (i.e., initial frame) during tracking. Some deep DCF-based trackers (e.g., \cite{ASRCF,STRCF,SMART,DSNet}) exploit efficient optimizations \& computations in the frequency domain, although employing pre-trained networks limits their speeds. Lately, custom-based trackers (e.g., \cite{ATOM,DiMP,PrDiMP,COMET,D3S,TMAML}) employ shallow networks with robust optimization strategies to attain high-speed tracking. Finally, numerous techniques have been used to speed up DL-based trackers (e.g., learning-based search strategy \cite{ACT}, domain adaption \cite{CRAC,TMAML}, embedding space learning \cite{RT-MDNet}, collaborative framework \cite{DCTN}, offline-trained CNN \cite{YCNN}, and efficient updating \& scale estimation strategies \cite{UCT}).
%----------------------------------------------------------------%
\section{Visual Tracking Benchmark Datasets}\label{sec:3}
Visual tracking benchmark datasets have been introduced to provide fair and standardized evaluations of single-object tracking algorithms. These benchmarks are mainly categorized based on generic or aerial-view tracking applications while providing short- or long-term scenarios. These datasets contain various sequences, frames, attributes, and classes (or clusters). The attributes include \textit{illumination variation} (IV), \textit{scale variation} (SV), \textit{occlusion} (OCC), \textit{deformation} (DEF), \textit{motion blur} (MB), \textit{fast motion} (FM), \textit{in-plane rotation} (IPR), \textit{out-of-plane rotation} (OPR), \textit{out-of-view} (OV), \textit{background clutter} (BC), \textit{low resolution} (LR), \textit{aspect ratio change} (ARC), \textit{camera motion} (CM), \textit{full occlusion} (FOC), \textit{partial occlusion} (POC), \textit{similar object} (SIB), \textit{viewpoint change} (VC), \textit{light} (LI), \textit{surface cover} (SC), \textit{specularity} (SP), \textit{transparency} (TR), \textit{shape} (SH), \textit{motion smoothness} (MS), \textit{motion coherence} (MCO), \textit{confusion} (CON), \textit{low contrast} (LC), \textit{zooming camera} (ZC), \textit{long duration} (LD), \textit{shadow change} (SHC), \textit{flash} (FL), \textit{dim light} (DL), \textit{camera shaking} (CS), \textit{rotation} (ROT), \textit{fast background change} (FBC), \textit{motion change} (MOC), \textit{object color change} (OCO), \textit{scene complexity} (SCO), \textit{absolute motion} (AM), \textit{size} (SZ), \textit{relative speed} (RS), \textit{distractors} (DI), \textit{length} (LE), \textit{fast camera motion} (FCM), \textit{object motion} (OM), \textit{object blur} (OB), \textit{large occlusion} (LOC), \textit{small objects} (SOB), \textit{occlusion with background clutter} (O-B), \textit{occlusion with rotation} (O-R) and \textit{long-term tracking} (LT). Table~\ref{tab5} compares the applications, scenarios, characteristics, missing labeled data for unsupervised training, and the overlap of single object tracking datasets. By different evaluation protocols, existing visual tracking benchmarks assess the accuracy \& robustness of trackers in realistic scenarios. The homogenized evaluation protocols facilitate straightforward comparison and development of visual trackers. Below the most popular visual tracking benchmark datasets and evaluation metrics are briefly described.
%----------------------------------------------------------------%
\begin{table*}
\caption{Comparison of visual tracking datasets. The abbreviations are denoted as RN: row number, NoV: number of videos, NoF: number of frames, NoA: number of attributes, OD: overlapped datasets, AL: absent labels, NoC: number of classes or clusters, AD: average duration (s: seconds).} % title of Table
\vspace{-1mm}
\centering % used for centering table
\resizebox{\textwidth}{!}{
\begin{tabular}{c c c c c c c c c c c c} % centered columns (4 columns)
\hline\hline %inserts double horizontal lines
Year & Application & Dataset & Scenario & NoV & NoF & NoA & OD & AL & NoC & AD & Attributes \protect\\ 
% Method & Value & Parameter & Value \\ [0.5ex] % inserts table
%heading \protect\\ 
\hline\hline 
2013 & Generic	&	OTB2013 &	ST &	51 &	29K &	11 &	VOT, OTB2015, TC128 &	No &	10 &	19.4s & IV, SV, OCC, DEF, MB, FM, IPR, OPR, OV, BC, LR \protect\\ 

2013-2019 &	Generic &	VOT2013-2019 & ST	&	16-60 &	6K-21K &	12 &	OTB, ALOV$++$, TC128, UAV123, NUS-PRO &	No &	12-24 &	12s & IV, SV, OCC, DEF, MB, BC, ARC, CM, MOC, OCO, SCO, AM \protect\\ 

2014 & Generic &	ALOV$++$ & ST	&	314 &	89K	 & 14 &	VOT, YouTube &	No &	64 &	16.2s & OCC, BC, CM, LI, SC, SP, TR, SH, MS, MCO, CON, LC, ZC, LD \protect\\ 

2015 & Generic &	OTB2015 & ST &	100 &	59K &	11 &	OTB2013, VOT, TC128 &	No &	16 &	19.8s & IV, SV, OCC, DEF, MB, FM, IPR, OPR, OV, BC, LR \protect\\ 

2015 & Generic &	TC128 &	ST &	129 &	55K	 & 11 &	OTB, VOT &	No &	27 &	15.6s & IV, SV, OCC, DEF, MB, FM, IPR, OPR, OV, BC, LR \protect\\ 

2016 & UAV & UAV123 & ST	&	123 &	113K &	12 &	VOT &	No &	9 &	30.6s & IV, SV, FM, OV, BC, LR, ARC, CM, FCM, FOC, POC, SIB, VC \protect\\ 
%------------------------------------------------------------------------------------------------------%
2016 & UAV & UAV20L & LT &	20 & 59K & 12 & VOT & No & 5 & 75s & IV, SV, FM, OV, BC, LR, ARC, CM, FCM, FOC, POC, SIB, VC \protect\\ 
%------------------------------------------------------------------------------------------------------%
2016 & Generic &	NUS-PRO & ST	&	365 &	135K &	12 &	VOT, YouTube &	No &	17 &	12.6s & SV, DEF, BC, FOC, POC, SIB, SHC, FL, DL, CS, ROT, FBC \protect\\ 

2017 & Generic &	NfS & ST	&	100 &	383K &	9 &	YouTube &	No &	17 &	15.6s & IV, SV, OCC, DEF, FM, OV, BC, LR, VC \protect\\ 

2017 & UAV &	DTB & ST	&	70 &	15K &	11 &	YouTube &	No &	15 &	7.2s & SV, OCC, DEF, MB, IPR, OPR, OV, BC, ARC, FCM, SIB \protect\\ 

2018 & Generic	&	TrackingNet & ST	&	30643 &	14.43M &	15 &	YTBB &	No &	27 &	16.6s & IV, SV, DEF, MB, FM, IPR, OPR, OV, BC, LR, ARC, CM, FOC, POC, SIB \protect\\ 

2018 & Generic	&	TinyTLP / TLPattr & ST	& 50 &	30K &	6 &	YouTube &	No &	N/A &	20s & FM, IV, SV, POC, OV, BC \protect\\ 

2018 & Generic	&	OxUvA &	LT &	366 &	1.55M &	6 &	YTBB &	Yes &	22 &	144s & SV, OV, SZ, RS, DI, LE \protect\\ 

2018 & Generic	&	TLP &	LT &	50 &	676K &	0 &	YouTube &	No &	N/A &	484.8s & - \protect\\ 

2018 & UAV	&	BUAA-PRO &	ST &	150	 & 8.7K &	12 &	NUS-PRO, YTBB &	No &	12 &	2s & SV, DEF, BC, FOC, POC, SIB, SHC, FL, DL, CS, ROT, FBC \protect\\ 

2018 & Generic	&	GOT10k &	ST &	10000 &	1.5M &	6 &	VOT, WordNet, ImageNet &	Yes &	563 &	16s & IV, SV, OCC, FM, ARC, LO \protect\\ 
%------------------------------------------------------------------------------------------------------%
2018 & UAV & UAVDT & ST & 50 & 80K & 9 & - & No & 3 & N/A & BC, CM, OM, SOB, IV, OB, SV, LOC, LT  \protect\\ 

2018 & Generic	&	LTB35 / VOT2018-LT &	LT &	35 &	147K &	10 &	YouTube, VOT, UAV20L &	No &	6 &	N/A & FOC, POC, OV, CM, FM, SC, ARC, VC, SIB, DEF \protect \\ 
%------------------------------------------------------------------------------------------------------%
2018-2020 & UAV	& VisDrone2018-2020 & ST & 132 & 106.4K & 12 & - & No & 4 & N/A & IV, SV, FM, OV, BC, LR, ARC, CM, FOC, POC, SIB, VC \protect \\ 

2019-2020 & UAV	& VisDrone2019-2020L & LT & 25 & 82.6K & 12 & - & No & 4 & N/A & IV, SV, FM, OV, BC, LR, ARC, CM, FOC, POC, SIB, VC \protect \\ 

2019 & Generic	&	LaSOT &	LT &	1400 &	3.5M &	14 &	YouTube, ImageNet &	Yes &	70 &	84.3s & IV, SV, DEF, MB, FM, OV, BC, LR, ARC, CM, FOC, POC, VC, ROT \protect\\ 
%------------------------------------------------------------------------------------------------------%
2020 & UAV & Small-90 & ST & 90 & N/A &	11 & UAV123, VOT, OTB, TC128 & No & N/A & N/A & IV, SV, OCC, DEF, MB, FM, IPR, OPR, OV, BC, LR \protect\\ 

2020 & UAV & Small-112 & ST & 112 & N/A & 11 & UAV123, VOT, OTB, TC128, VisDrone & No & N/A & N/A & IV, SV, OCC, DEF, MB, FM, IPR, OPR, OV, BC, LR \protect\\ 

2021 & Generic & TracKlinic & ST & 2390 & 280K & 9 & OTB, TC128, LaSOT & No & N/A & N/A & IV, SV, OCC, MB, OV, BC, ROT, O-B, O-R \protect\\ 
  \hline %inserts single line
  \end{tabular}
  \label{tab5}
 }
 \vspace{-4mm}
\end{table*}
%----------------------------------------------------------------%
\vspace{-.2cm}
\subsection{Short-term Tracking Datasets} \label{Sec3.1}
\textbf{Generic Object Tracking.}
As one of the first object tracking benchmarks, OTB2013 \cite{OTB2013} is developed by fully annotated video sequences to address the issues of reported tracking results based on a few video sequences or inconsistent initial conditions or parameters. The OTB2015 \cite{OTB2015} is an extended OTB2013 dataset with the aim of unbiased performance comparisons. To compare the performance of visual trackers on color sequences, the \textit{Temple-Color 128} (TColor128 or TC128) \cite{TC128} collected a set of 129 fully annotated video sequences that 78 ones are different from the OTB datasets. The \textit{Amsterdam library of ordinary videos} (ALOV) dataset \cite{SurveySmeulders} has been gathered to cover diverse video sequences and attributes. By emphasizing challenging visual tracking scenarios, the ALOV dataset comprises 304 assorted short videos and 10 longer ones. The video sequences are chosen from real-life YouTube videos and have 13 difficulty degrees. The videos of ALOV have been categorized according to one of its attributes (Table~\ref{tab5}), although in the OTB datasets, each video has been annotated by several visual attributes.\\
\indent Motivated by the large dataset's inequality with a useful one, the VOT dataset \cite{VOT-2013,VOT-2014,VOT-2015,VOT-2016,VOT-2017,VOT-2018,VOT-2019} aims to provide a diverse and sufficiently small dataset from existing ones, annotated per-frame by rotatable BBs and visual properties. To evaluate different visual tracking methods fast and straightforward, the VOT includes a \textit{visual tracking exchange} (TraX) protocol \cite{TraX} that not only prepares data, runs experiments, and performs analyses but also can detect tracking failures (i.e., losing the target) and re-initialize the tracker after each failure to assess tracking robustness. Despite some small and saturated tracking datasets in the wild, mostly provided for object detection tasks, the large-scale TrackingNet benchmark dataset \cite{TrackingNet} has been proposed to properly feed deep visual trackers. It provides videos for tracking in the wild with 500 original videos, more than 14 million upright BB annotations, densely annotated data in time, rich distribution of object classes, and real-world scenarios by sampled YouTube videos. \\
\indent For tracking pedestrian and rigid objects, the \textit{NUS people and rigid objects} (NUS-PRO) dataset \cite{NUS-PRO} has been provided 365 video sequences from YouTube that are majorly captured by moving cameras and annotated the level of occluded objects of each frame with no occlusion, partial occlusion, and full occlusion labels. By higher frame rate (240 FPS) cameras, the \textit{need for speed} (NfS) dataset \cite{NfS} provides video sequences from real-world scenarios to systematically investigate trade-off bandwidth constraints related to real-time analysis of visual trackers. These videos are either recorded by hand-held iPhone/iPad cameras or from YouTube videos. Two short sequence datasets, namely TinyTLP \& TLPattr \cite{TLP}, are derived from the long-term TLP dataset. For each sequence, one visual attribute has been specified for investigating various challenges. The large high-diversity dataset, called GOT-10k \cite{GOT-10k}, includes more than ten thousand videos classified into 563 classes of moving objects and 87 classes of motion to cover as many challenging patterns in real-world scenarios as possible. The GOT-10k has informative continuous attributes, including absent labels, which show the target does not exist in the frame. Finally, the TracKlinic \cite{TracKlinic} introduces a toolkit (collected from OTB2015, TC128, \& LaSOT) that consists of just one challenging factor per sequence to evaluate visual trackers. It also provides two challenging O-B \& O-R attributes, including occlusion with background clutter \& rotation. \\
%----------------------------------------------------------------%
\indent \textbf{Aerial View Object Tracking.}
Tracking from aerial views has been developed in recent years, considering the broad range of applications. The \textit{unmanned aerial vehicle 123} (UAV123) \cite{UAV123} provides a sparse and low altitude aerial-view tracking dataset that contains the realistic and synthetic HD video sequences captured by professional-grade flying robots, a board-cam mounted on small low-cost flying robots \& simulator ones. \textit{Drone tracking benchmark} (DTB) \cite{DTB} is a dataset captured by flying robots or drones that consists of RGB videos with massive displacement of target location due to abrupt camera motion. The BUAA-PRO dataset \cite{BUAA-PRO} is a segmentation-based benchmark dataset to address the problem of inevitable non-target elements in BBs. It exploits the segmentation mask-based version of a level-based occlusion attribute. The UAVDT dataset \cite{UAVDT2018} provides an aerial-view dataset with high object density scenarios (e.g., different weather conditions, camera views, flying altitudes) focusing on pedestrians and vehicles. Furthermore, VisDrone dataset \cite{VisDrone2018,VisDrone2019} includes videos captured by different drone platforms in real-world scenarios. For small object tracking, the Small-90 dataset \cite{Small90Dataset} presents aerial videos mostly collected from other visual tracking datasets. By adding 22 more challenging sequences, the Small-112 dataset \cite{Small90Dataset} has been formed based on the Small-90 dataset. 
%----------------------------------------------------------------%
\vspace{-.2cm}
\subsection{Long-term Tracking Datasets} \label{Sec3.1.1}
\textbf{Generic Object Tracking.}
With the aim of long-term tracking of frequently disappearance targets, the OxUvA dataset \cite{OxUvA} includes 14 hours of videos from \textit{YouTube-BoundingBoxes} (or YTBB) \cite{YouTube-BB} to provide development and test sets with continuous attributes. Also, it provides absent labels, which show that the target does not exist in some frames. The TLP dataset \cite{TLP} also has been collected high-resolution videos with a longer duration per sequence from YouTube, which provides the possibility of studying tracking consistency. However, target disappearances do not frequently occur in the TLP dataset. Hence, the LTB-35 \cite{Long-term_NYSM} presents an enriched long-term dataset with consistent target disappearances (twelve disappearances on average for each video). The \textit{large-scale single object tracking} (LaSOT) \cite{LaSOT} has been developed to address the problems of existing datasets, such as small scale, lack of high-quality, dense annotations, short video sequences, and category bias. The object categories are from the ImageNet and a few visual tracking applications (such as drones) with an equal number of videos per category. The training and testing subsets include 1120 (2.3M frames) and 280 (690K frames) video sequences, respectively. \\ %----------------------------------------------------------------%
\indent\textbf{Aerial View Object Tracking.}
As the parent set of the UAV-123 dataset, the UAV20L is an aerial surveillance dataset, including one continuous shot video sequences. It consists of tolerable occlusions and provides difficult scenarios for small object tracking. Moreover, the VisDrone-2019/2020L dataset \cite{VisDrone2019} includes 25 challenging sequences (i.e., 12/13 videos in the day/night) with tiny targets.
%----------------------------------------------------------------%
\vspace{-.2cm}
\subsection{Evaluation Metrics} \label{Sec3.2}
Visual trackers are evaluated by two fundamental evaluation categories of performance measures and performance plots to perform experimental comparisons on large-scale datasets. These metrics are briefly described as follows.
%----------------------------------------------------------------%
\subsubsection{\textbf{Performance Measures}} \label{Sec3.2.1}
Performance measures attempt to intuitively interpret performance comparisons in terms of complementary metrics of accuracy, robustness and tracking speed. For long-term trackers, the measures close relate to their detection counterparts to reflect re-detection \& target absence prediction capabilities. In the following, these measures are concisely investigated. 
%----------------------------------------------------------------%
\begin{enumerate}[wide, label=(\Alph*), labelwidth=!,labelindent=0pt]
\item \textbf{Short-term Tracking Measures:}
\begin{enumerate}[wide, label=(\roman*), labelwidth=! ]%,labelindent=0pt
\item \textbf{Center location error (CLE)/ (Normalized) precision:} The CLE or precision metric is defined as the average Euclidean distance between the precise ground-truth locations of the target and estimated locations by a visual tracker. The CLE is the oldest metric that is sensitive to dataset annotation and does not consider tracking failures, and ignores the target’s BB, resulting in significant errors. The normalized precision \cite{TrackingNet} aims to relieve the sensitivity of CLE to the size of BBs and frame resolutions. Given the size of ground-truth BB ($b_g$), this metric normalizes the CLE over $b_g$ to keep its consistency for various target scales.
%----------------------------------------------------------------%
\item \textbf{Accuracy:} For this metric, first, the overlap score is calculated as $S=\frac{\left|b_t\cap b_g\right|}{\left|b_t\cup b_g\right|}$ which b${}_{g}$, b${}_{t}$, $\cap $, $\cup $ and $\left|.\right|$ represent the ground-truth BB, an estimated BB by a visual tracking method, intersection operator, union operator, and the number of pixels in the resulted region, respectively. By considering a certain threshold, the overlap score indicates a visual tracker's success in one frame. The accuracy is then calculated by the \textit{average overlap scores} (AOS) during the tracking when a visual tracker's estimations have overlap with the ground-truth ones. This metric jointly considers both location and region to measure the estimated target's drift rate up to its failure.
%----------------------------------------------------------------%
\item \textbf{Robustness/ failure score:} The robustness or failure score is defined as the number of required re-initializations when a tracker loses (or drifts) the target during the tracking task. The failure is detected when the overlap score drops to zero.
%----------------------------------------------------------------%
\item \textbf{Expected average overlap (EAO):} This score is interpreted as the combination of accuracy and robustness scores. Given N${}_{s}$ frames long sequences, the EAO score is calculated as ${\widehat{\mathit{\Phi}}}_{N_s}=\left\langle \frac{1}{N_s}\sum^{N_s}_{i=1}{{\mathit{\Phi}}_i}\right\rangle $, where ${\mathit{\Phi}}_i$ is defined as the average of per-frame overlaps until the end of sequences, even if failure leads to zero overlaps.
%----------------------------------------------------------------%
\item \textbf{Area under curve (AUC):} The AUC score has defined the average success rates (normalized between 0 and 1) according to the pre-defined thresholds. To rank the visual tracking methods based on their overall performance, the AUC score summarizes the AOS of visual tracking methods across a sequence.
\end{enumerate}
%----------------------------------------------------------------%
\item \textbf{Long-term Tracking Measures:}
\begin{enumerate}[wide, label=(\roman*), labelwidth=!]
\item \textbf{Precision ($Pr$):} Tracking measures for long-term trackers depend on being a target in the scene and prediction confidence to be higher than a classification threshold for each frame. The precision \cite{Long-term_NYSM} is calculated by the \textit{intersection over union} (IoU) between the ground-truth ($b_g$) and predicted target ($b_t$), normalized by the number of frames with existing predictions. The integration of these scores over all precision thresholds provides the overall tracking precision.
\item \textbf{Recall ($Re$):} Similar to the precision, it calculates the IoU between the $b_g$ and $b_t$, which is normalized by the number of frames with no absent targets. The overall tracking recall \cite{Long-term_NYSM} is achieved by integrating the scores over all recall thresholds.
\item \textbf{F-score:} It compromises the precision \& recall scores by calculating $F=\frac{2Pr.Re}{Pr+Re}$ to rank the trackers according to their maximum values over all thresholds.
\item \textbf{Maximum Geometric Mean (MaxGM):} Inspired by binary classification, the MaxGM employs the \textit{true positive rate} (TPR) and \textit{true negative rate} (TNR) for evaluation of trackers. While the TPR reports the fraction of correctly located targets, the TNR presents the fraction of correctly reported absent targets. As a single metric, the geometric mean is defined as $GM=\sqrt{TPR.TNR}$ but it will be zero for the trackers that cannot predict absent targets. Hence, $MaxGM=\max_{0\leq p\leq 1}\sqrt{\{(1-p).TPR)\}\{(1-p).TNR+p\}}$ provides a more informative comparison in terms of various probabilistic thresholds $p$.
\end{enumerate}
\end{enumerate}
%----------------------------------------------------------------%
\subsubsection{\textbf{Performance Plots}} \label{Sec3.2.2}
Generally, visual trackers are analyzed in terms of various thresholds to provide more intuitive quantitative comparisons. These metrics are summarized as follows. 
%----------------------------------------------------------------%
\begin{enumerate}[wide, label=(\Alph*), labelwidth=!, labelindent=0pt]
\item \textbf{Short-term Tracking Plots:}
\begin{enumerate}[wide, label=(\roman*), labelwidth=!]%, labelindent=0pt
\item \textbf{Precision plot:} Given the CLEs per different thresholds, the precision plot shows the percentage of video frames in which the estimated locations have at most the specific threshold with the ground-truth locations.

\item \textbf{Success plot:} Given the calculated various accuracy per thresholds, the success plot measures the percentage of frames in which the estimated overlaps and the ground-truth ones have larger overlap than a certain threshold. 

\item \textbf{Expected average overlap curve:} For an individual length of video sequences, the expected average overlap curve has resulted from the range of values in a specific interval $\left[N_{lo},N_{hi}\right]$ as $\widehat{\mathit{\Phi}}=\frac{1}{N_{hi}-N_{lo}}\sum^{N_{hi}}_{N_s=N_{lo}}{{\widehat{\mathit{\Phi}}}_{N_s}}$.
\end{enumerate}
%----------------------------------------------------------------%
\item \textbf{Long-term Tracking Plots:}
\begin{enumerate}[wide, label=(\roman*), labelwidth=!]%, labelindent=0pt
\item \textbf{Precision/ Recall plot:} It is used to compare long-term tracking performances and analyze their detection capabilities in terms of various thresholds.
\item \textbf{F-score plot:} This is the main curve to rank the tracking methods based on the highest score on the plot.
\end{enumerate}
\end{enumerate}
%----------------------------------------------------------------%
\vspace{-.3cm}
\section{Experimental Analyses}\label{sec:4}
To analyze the performance of state-of-the-art visual tracking methods, 48 different methods are quantitatively compared on seven well-known tracking datasets, namely OTB2013 \cite{OTB2013}, OTB2015 \cite{OTB2015}, VOT2018 \cite{VOT-2018}, LaSOT \cite{LaSOT}, UAV-123 \cite{UAV123}, UAVDT \cite{UAVDT2018}, and VisDrone2019-test-dev \cite{VisDrone2019}. Due to the page limitation, all experimental results are publicly available on \url{https://github.com/MMarvasti/Deep-Learning-for-Visual-Tracking-Survey}. The included 48 DL-based trackers in the experiments are shown in Table~\ref{tab6}. All evaluations are performed on an Intel I7-9700K 3.60GHz CPU with 32GB RAM with the aid of MatConvNet toolbox \cite{MatConvNet} that uses an NVIDIA GeForce RTX 2080Ti GPU for its computations. The OTB, LaSOT, UAV123, UAVDT, and VisDrone2019 toolkits evaluate the visual trackers in terms of the well-known precision \& success plots and then rank the methods based on the AUC score. For performance comparison on the VOT2018 dataset, the visual trackers have been assessed based on the TraX evaluation protocol using three primary measures of accuracy, robustness, and EAO to provide the \textit{Accuracy-Robustness} (AR) plots, expected average overlap curve, and ordering plots according to its five challenging visual attributes \cite{VOT-2018}.
%----------------------------------------------------------------%
\begin{table}
\caption{State-of-the-art visual tracking methods for experimental comparisons on visual tracking datasets.}
\vspace{-2mm}
% title of Table
\centering % used for centering table
\resizebox{9cm}{!}{
\begin{tabular}{c c c c} % centered columns (4 columns)
\hline\hline %inserts double horizontal lines
Published in & Visual Tracking Method & Exploited Features & Test Datasets \protect\\ 
% Method & Value & Parameter & Value \\ [0.5ex] % inserts table
%heading \protect\\ 
\hline\hline 
ICCV 2015 &	HCFT \cite{HCFT} & DAF & OTB, LaSOT \protect\\ 
ICCV 2015 &	DeepSRDCF \cite{DeepSRDCF} & DAF, HOG & OTB, VOT2018 \protect\\ 
ECCV 2016 &	CCOT \cite{CCOT} & DAF & OTB, VOT2018, UAVDT \protect\\ 
ECCVW 2016 &	SiamFC \cite{SiamFC} & DAF & OTB, LaSOT, UAVDT \protect\\ 
CVPR 2016 &	SINT \cite{SINT} & DAF & OTB, LaSOT, UAVDT \protect\\ 
CVPR 2016 &	MDNet \cite{MDNet} & DAF & OTB, LaSOT, UAVDT \protect\\ 
CVPR 2016 &	HDT \cite{HDT} & DAF & OTB, UAVDT \protect\\ 
ICCV 2017, TIP 2019 &	PTAV \cite{PTAV-ICCV,PTAV} & DAF, HOG & OTB, LaSOT, UAVDT \protect\\ 
ICCV 2017 &	CREST \cite{CREST} & DAF & OTB, UAVDT \protect\\ 
ICCV 2017 &	Meta-CREST \cite{CREST} & DAF & OTB \protect\\ 
ICCV 2017 &	UCT \cite{UCT} & DAF & OTB, VOT2018 \protect\\ 
ICCV 2017 &	DSiam \cite{DSiam} & DAF & VOT2018, LaSOT \protect\\ 
CVPR 2017 &	CFNet \cite{CFNet} & DAF & OTB, VOT2018, LaSOT, UAVDT \protect\\ 
CVPR 2017 &	ECO \cite{ECO} & DAF, HOG, CN & OTB, VOT2018, LaSOT, UAV123, UAVDT, VisDrone2019 \protect\\ 
CVPR 2017 &	DeepCSRDCF \cite{DeepCSRDCF} & DAF, HOG, CN & VOT2018, LaSOT \protect\\ 
CVPR 2017 &	MCPF \cite{MCPF} & DAF & OTB, VOT2018 \protect\\ 
CVPR 2017 &	ACFN \cite{ACFN} & DAF, HOG, Color & OTB \protect\\ 
arXiv 2017 &	DCFNet \cite{DCFNet} & DAF & OTB \protect\\ 
arXiv 2017 &	DCFNet2 \cite{DCFNet} & DAF & OTB, VOT2018 \protect\\ 
ECCV 2018 &	TripletLoss-CFNet \cite{Tripletloss} & DAF & OTB \protect\\ 
ECCV 2018 &	TripletLoss-SiamFC \cite{Tripletloss} & DAF & OTB \protect\\ 
ECCV 2018 &	TripletLoss-CFNet2 \cite{Tripletloss} & DAF & OTB \protect\\ 
ECCV 2018 &	UPDT \cite{UPDT} & DAF, HOG, CN & VOT2018 \protect\\ 
ECCV 2018 &	DaSiamRPN \cite{DaSiamRPN} & DAF & VOT2018, UAV123 \protect\\ 
ECCV 2018 &	StructSiam \cite{StructSiam} & DAF & LaSOT \protect\\ 
ECCVW 2018 &	Siam-MCF \cite{Siam-MCF} & DAF & OTB \protect\\ 
CVPR 2018 &	TRACA \cite{TRACA} & CDAF & OTB, VOT2018, LaSOT \protect\\ 
CVPR 2018 &	VITAL \cite{VITAL} & DAF & OTB, LaSOT \protect\\ 
CVPR 2018 &	DeepSTRCF \cite{STRCF} & DAF, HOG, CN & OTB, VOT2018, LaSOT, UAV123 \protect\\ 
CVPR 2018 &	SiamRPN \cite{SiamRPN} & DAF & OTB, VOT2018, UAV123 \protect\\ 
CVPR 2018 &	SA-Siam \cite{SA-Siam} & DAF & OTB, VOT2018 \protect\\ 
CVPR 2018 &	LSART \cite{LSART} & DAF & VOT2018 \protect\\ 
CVPR 2018 &	DRT \cite{DRT} & DAF, HOG, CN & VOT2018 \protect\\ 
NIPS 2018 &	DAT \cite{DAT} & DAF & OTB, VOT2018 \protect\\ 
PAMI 2018 &	HCFTs \cite{HCFTs} & DAF & OTB \protect\\ 
IJCV 2018 &	LCTdeep \cite{LCTdeep} & DAF & OTB \protect\\ 
TIP 2018 &	CFCF \cite{CFCF} & DAF, HOG & VOT2018 \protect\\ 
CVPR 2019 &	C-RPN \cite{CRPN} & DAF & OTB, VOT2018, LaSOT \protect\\ 
CVPR 2019 &	GCT \cite{GCT} & DAF & OTB, VOT2018, UAV123 \protect\\ 
CVPR 2019 &	SiamMask \cite{SiamMask} & DAF & VOT2018, UAVDT, VisDrone2019 \protect\\ 
CVPR 2019 &	SiamRPN$++$ \cite{SiamRPN++} & DAF & OTB, VOT2018, UAV123, UAVDT, VisDrone2019 \protect\\ 
CVPR 2019 &	TADT \cite{TADT} & DAF & OTB \protect\\ 
CVPR 2019 &	ASRCF \cite{ASRCF} & DAF, HOG & OTB, LaSOT \protect\\ 
CVPR 2019 &	SiamDW-SiamRPN \cite{SiamDW} & DAF & OTB, VOT2018, UAVDT, VisDrone2019 \protect\\ 
CVPR 2019 &	SiamDW-SiamFC \cite{SiamDW} & DAF & OTB, VOT2018 \protect\\ 
%------------------------------------------------------------------------%
CVPR 2019 &	ATOM \cite{ATOM} & DAF & VOT2018, LaSOT, UAV123, UAVDT, VisDrone2019 \protect\\ 
ICCV 2019 &	DiMP50 \cite{DiMP} & DAF & VOT2018, LaSOT, UAV123, UAVDT, VisDrone2019 \protect\\ 
CVPR 2020 &	PrDiMP50 \cite{PrDiMP} & DAF & LaSOT, UAVDT, VisDrone2019 \protect\\ 
  \hline %inserts single line
  \end{tabular}
  \label{tab6}
 }
 \vspace{-4mm}
\end{table}
%----------------------------------------------------------------%
\subsection{Quantitative Comparisons} \label{Sec4.1}
\begin{figure*}[hbt!]
\justify
\subfigure{\includegraphics[width=4.5cm, height=3cm]{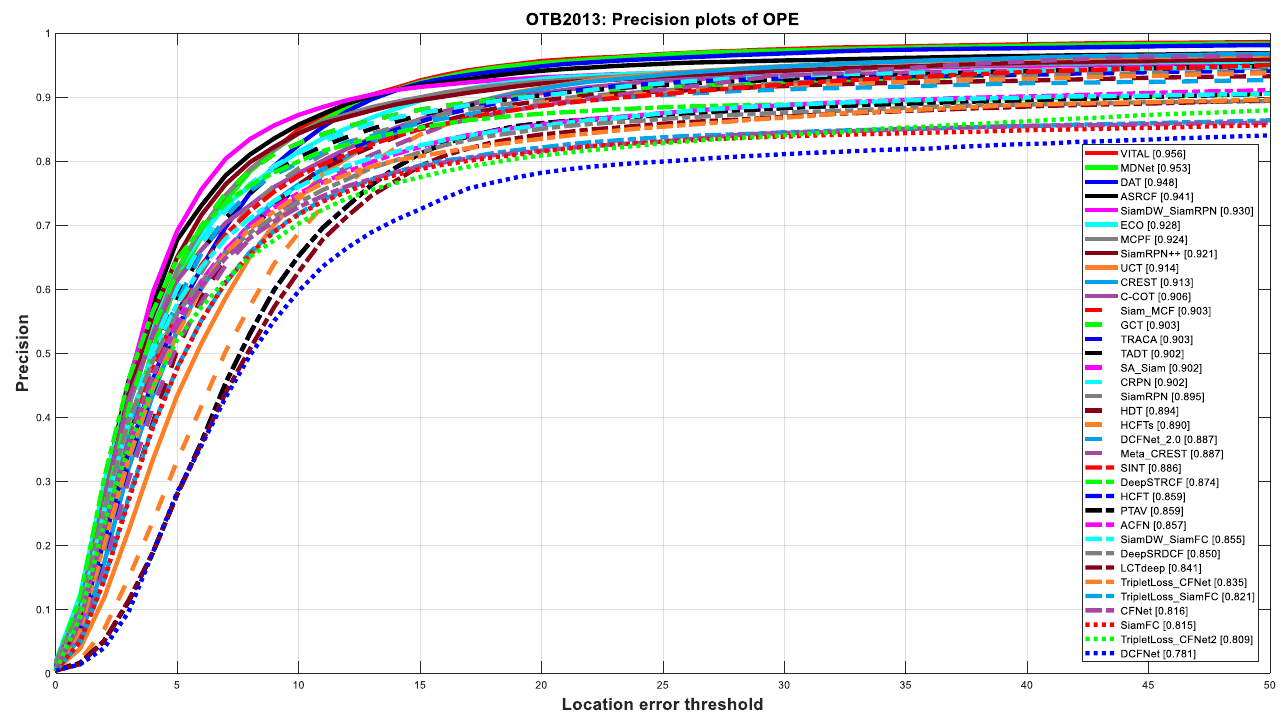}} 
\hspace{-.25cm}
\subfigure{\includegraphics[width=4.5cm, height=3cm]{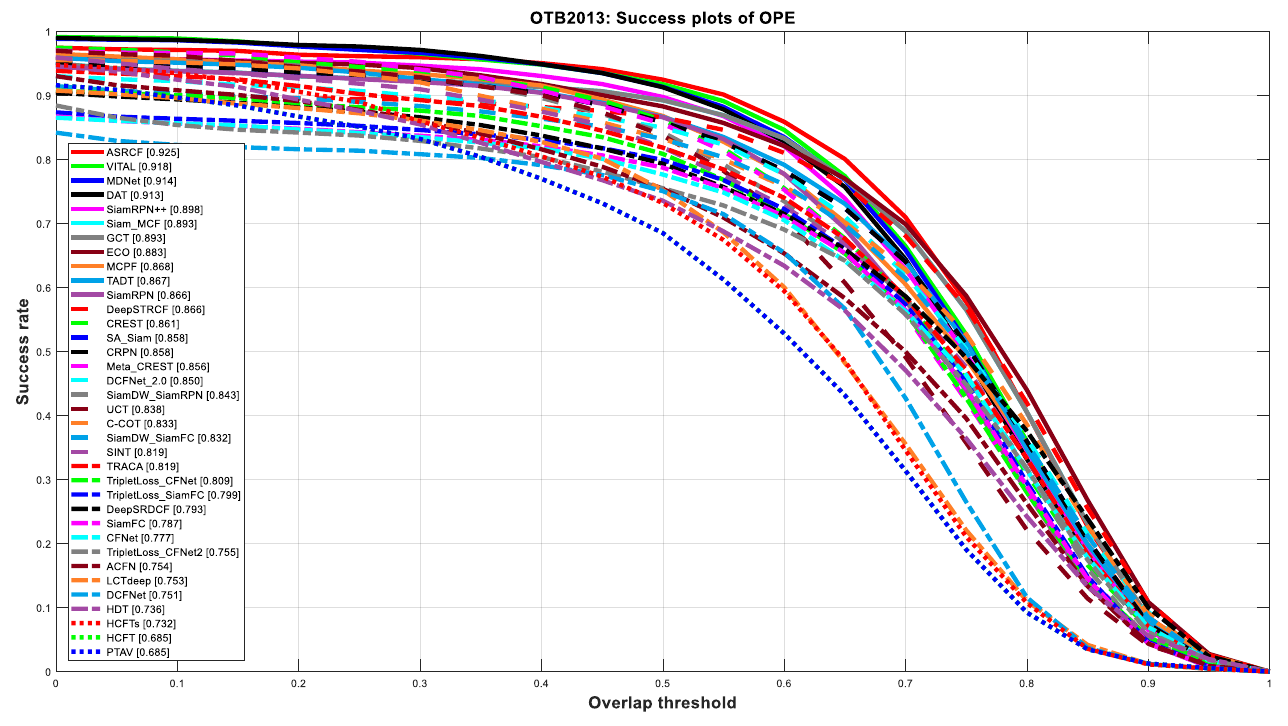}}
\hspace{-.25cm}
\subfigure{\includegraphics[width=4.5cm, height=3cm]{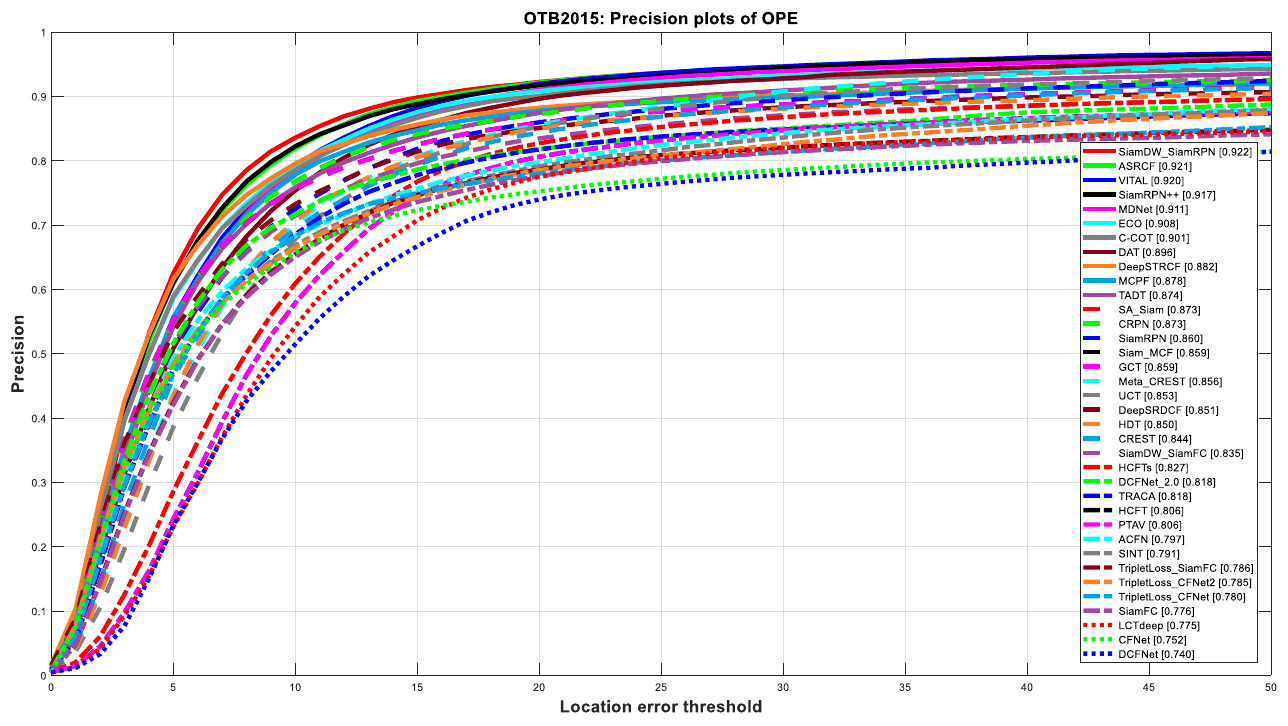}}
\hspace{-.25cm}
\subfigure{\includegraphics[width=4.5cm, height=3cm]{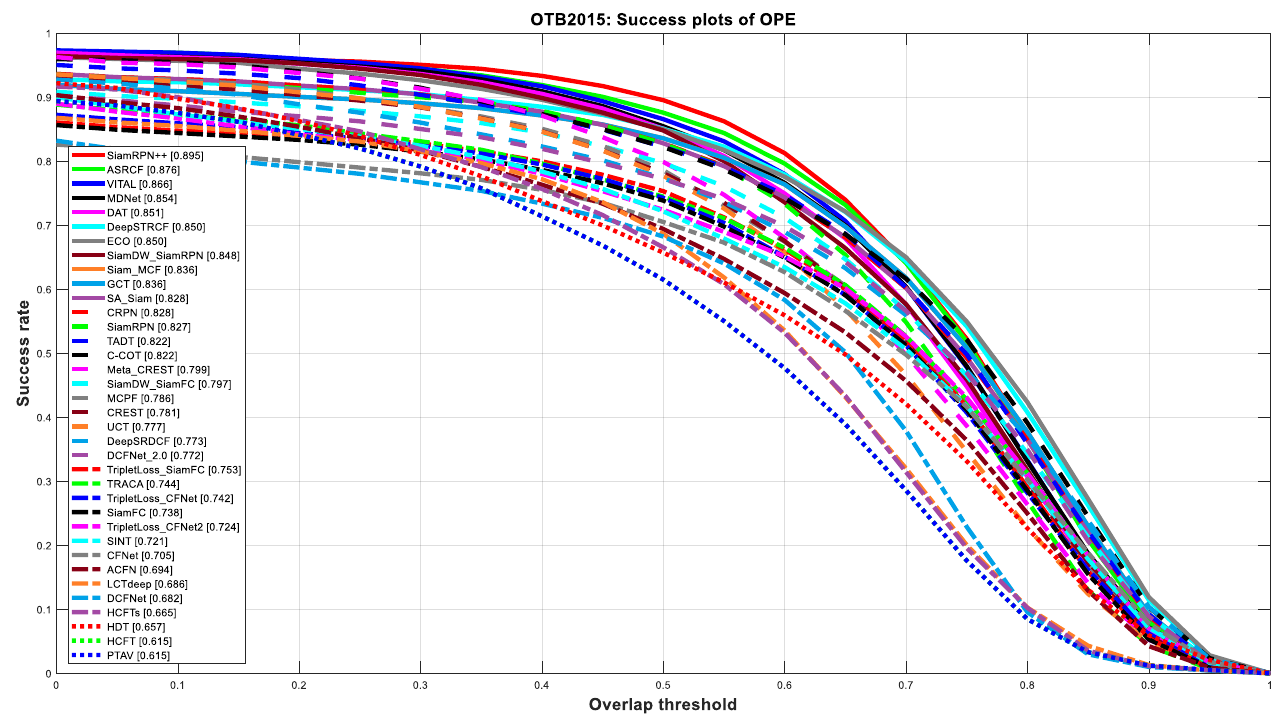}}
\vspace{-.5cm}
\justify
\subfigure{\includegraphics[width=4.5cm, height=3cm]{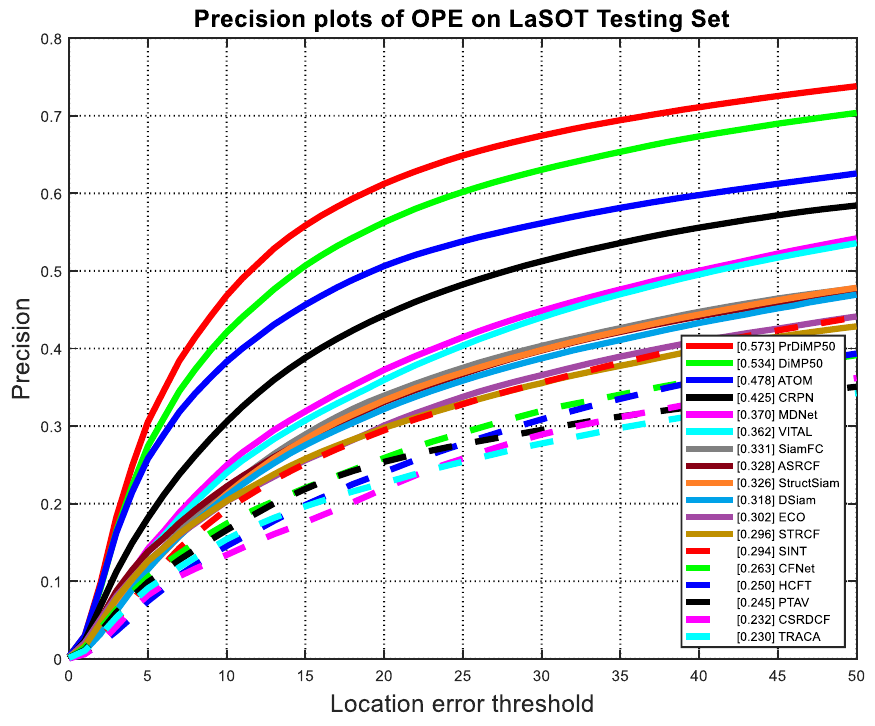}} 
\hspace{-.25cm}
\subfigure{\includegraphics[width=4.5cm, height=3cm]{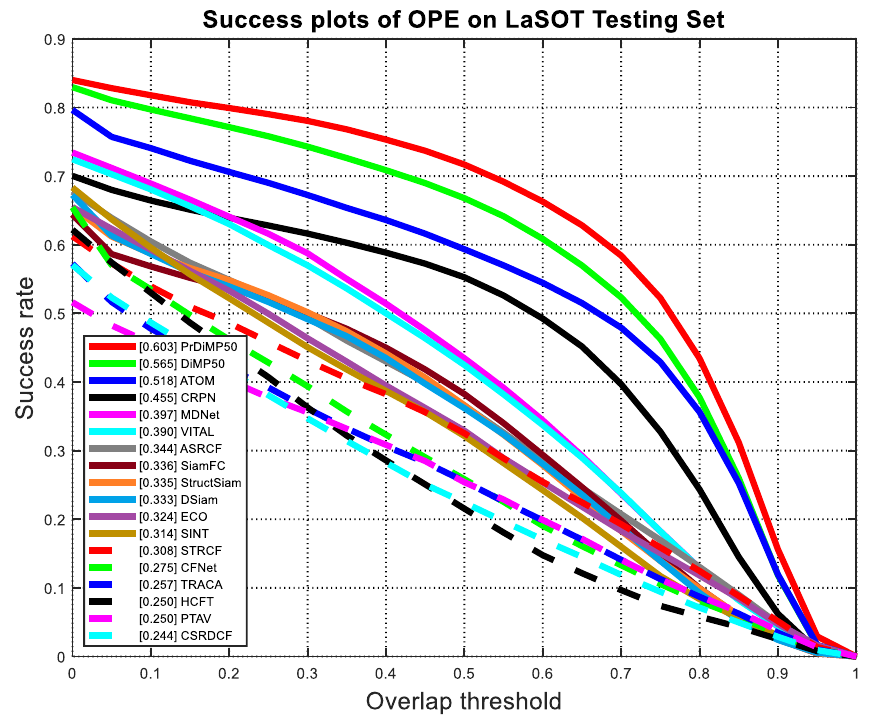}}
\hspace{-.25cm}
\subfigure{\includegraphics[width=4.5cm, height=3cm]{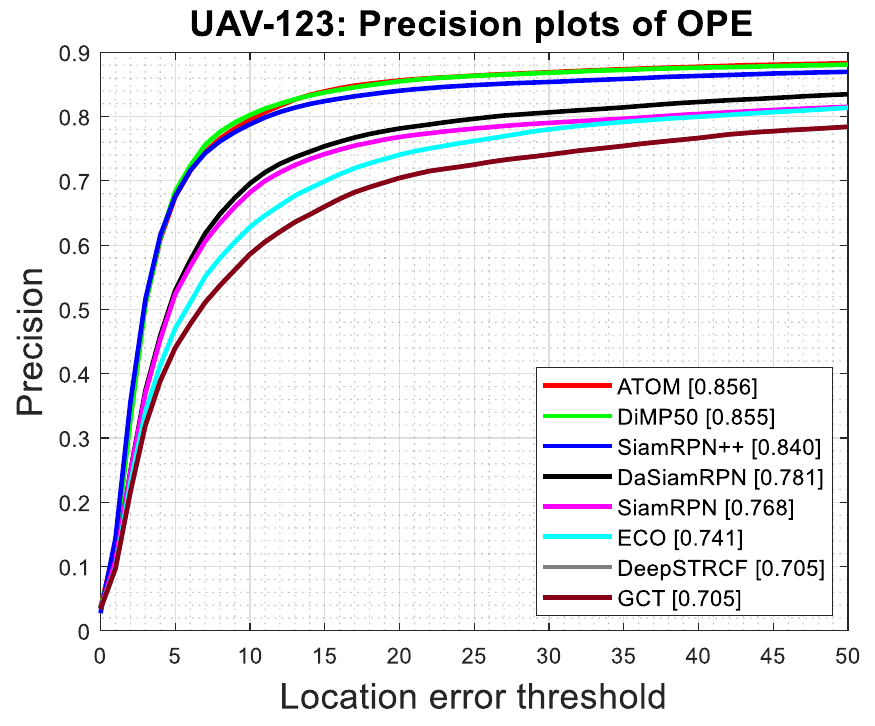}}
\hspace{-.25cm}
\subfigure{\includegraphics[width=4.5cm, height=3cm]{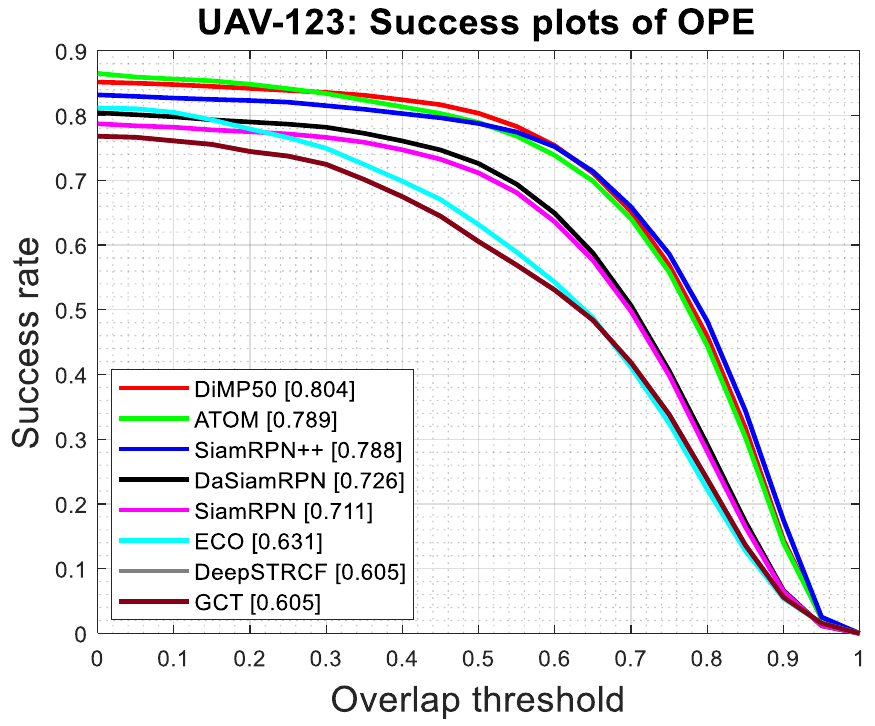}}
\vspace{-.5cm}
\justify
\subfigure{\includegraphics[width=4.5cm, height=3cm]{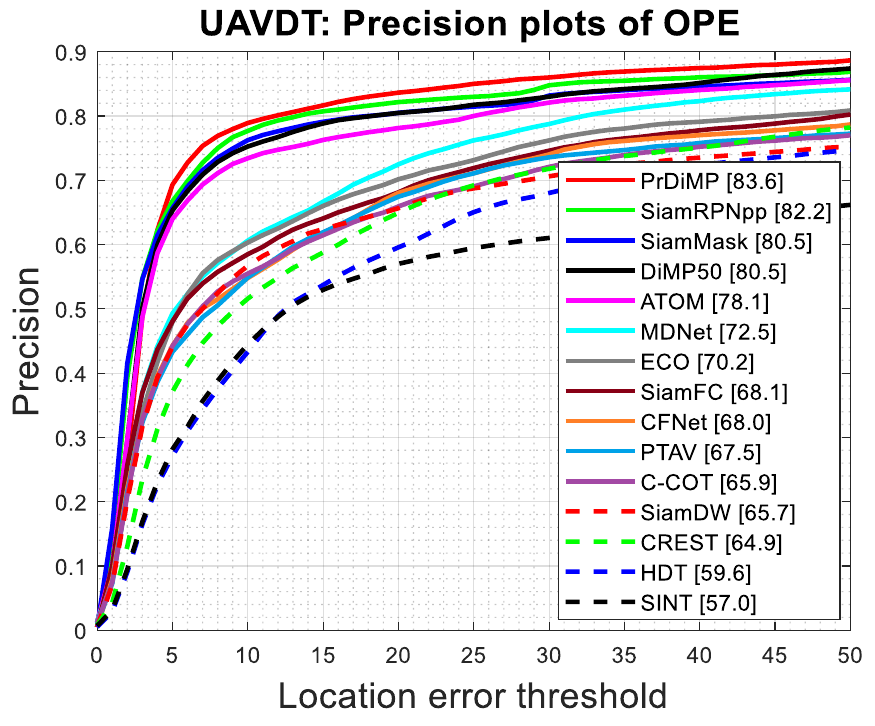}} 
\hspace{-.25cm}
\subfigure{\includegraphics[width=4.5cm, height=3cm]{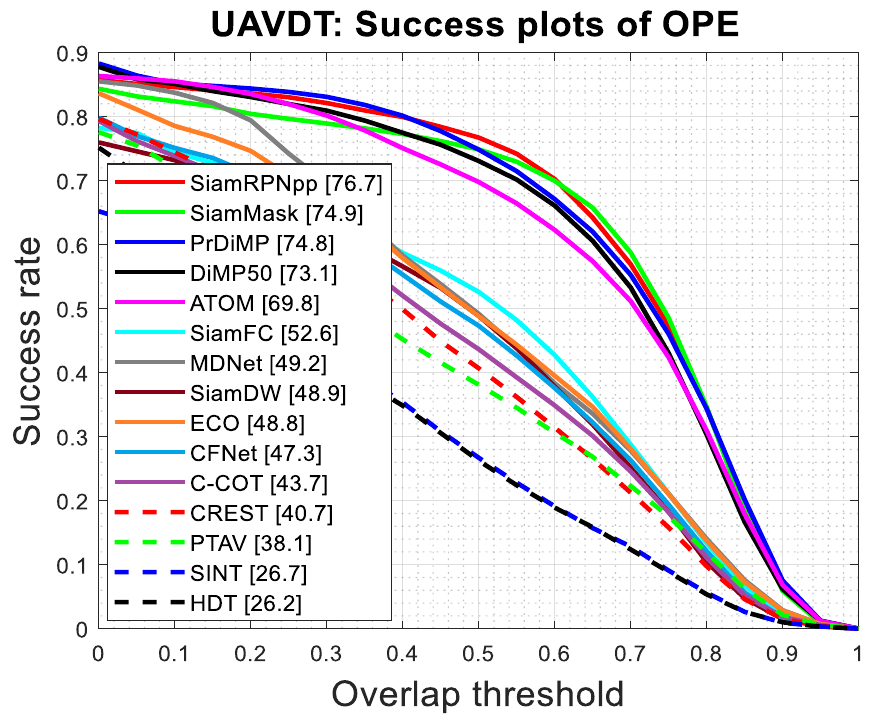}}
\hspace{-.25cm}
\subfigure{\includegraphics[width=4.5cm, height=3cm]{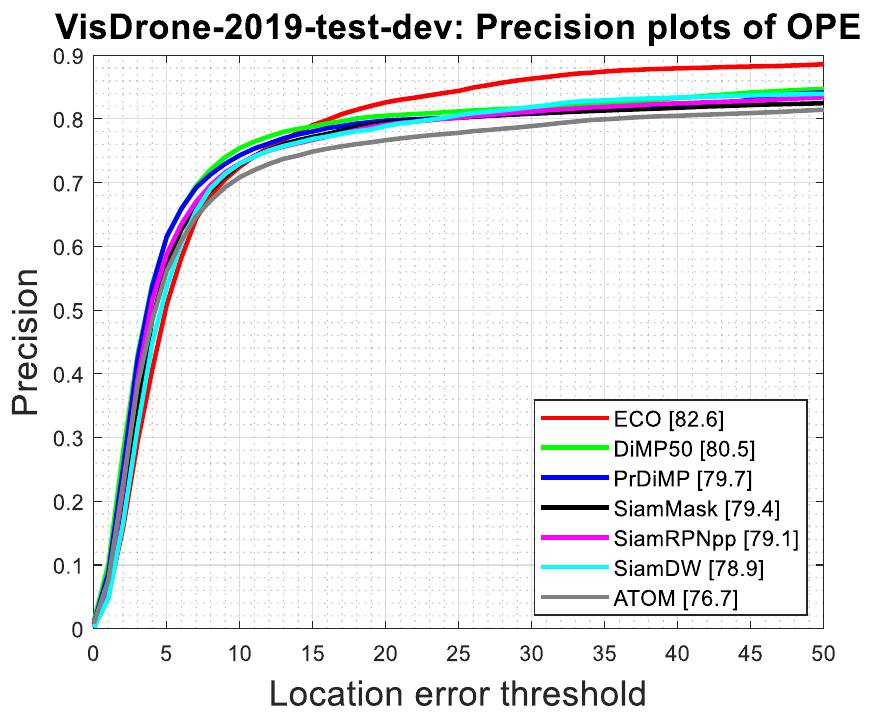}}
\hspace{-.25cm}
\subfigure{\includegraphics[width=4.5cm, height=3cm]{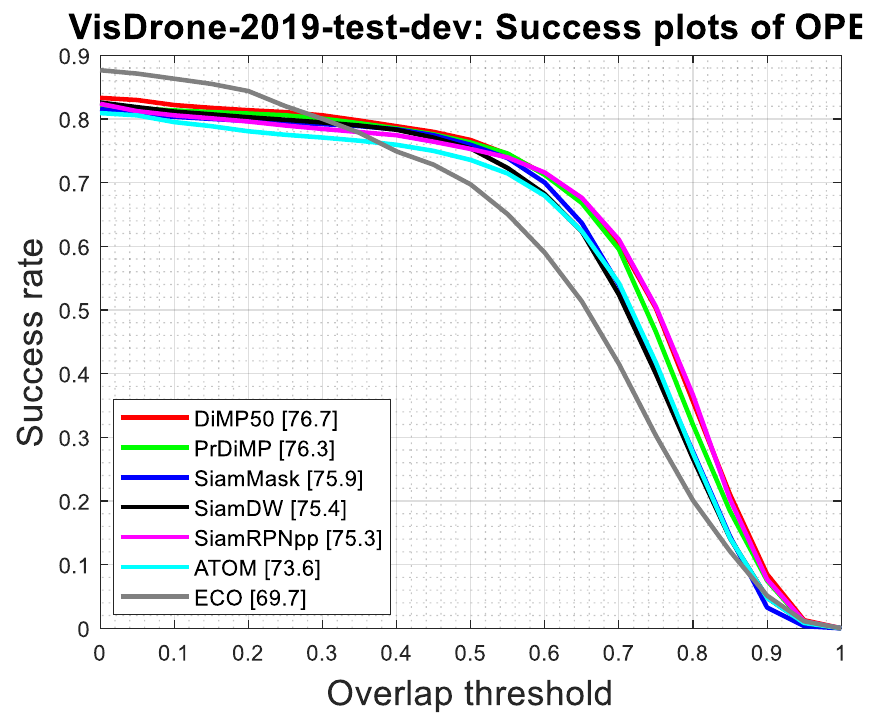}}
\vspace{-.3cm}
\caption{Overall experimental comparison of state-of-the-art visual tracking methods on the OTB2013, OTB2015, LaSOT, UAVDT, and VisDrone2019 visual tracking datasets.}
\label{fig:Results_OTB_LaSOT} 
\vspace{-2mm}
\end{figure*}

\begin{figure*}[hbt!]
\justify
\subfigure{\includegraphics[width=6cm, height=3cm]{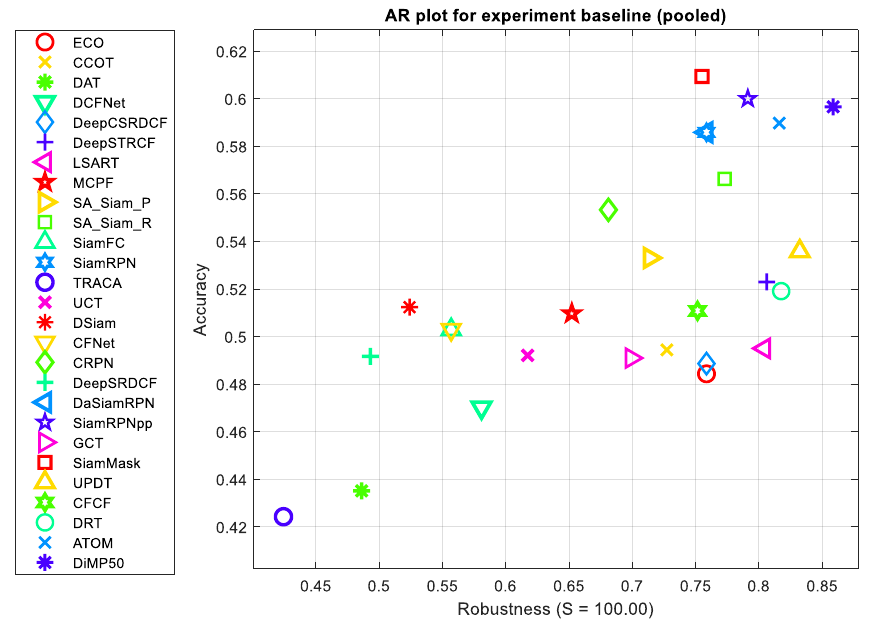}} 
\hspace{0mm}
\subfigure{\includegraphics[width=6cm, height=3cm]{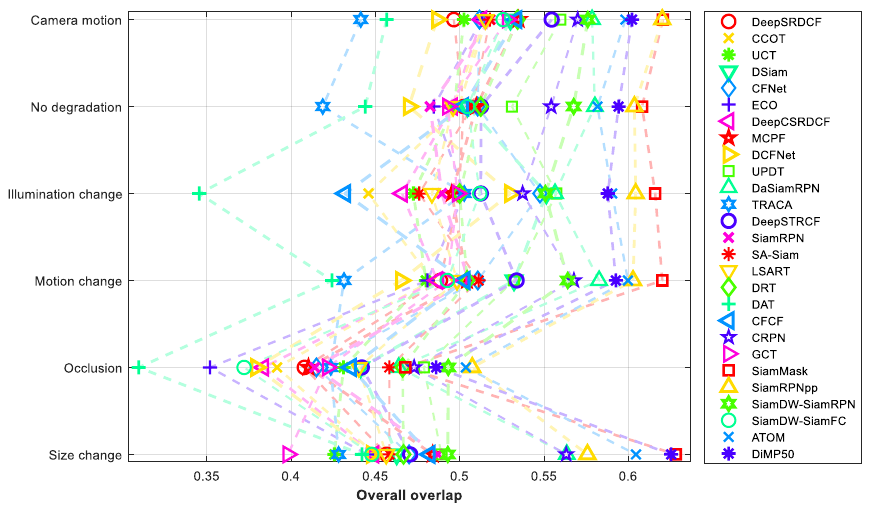}}
\hspace{0mm}
\subfigure{\includegraphics[width=6cm, height=3cm]{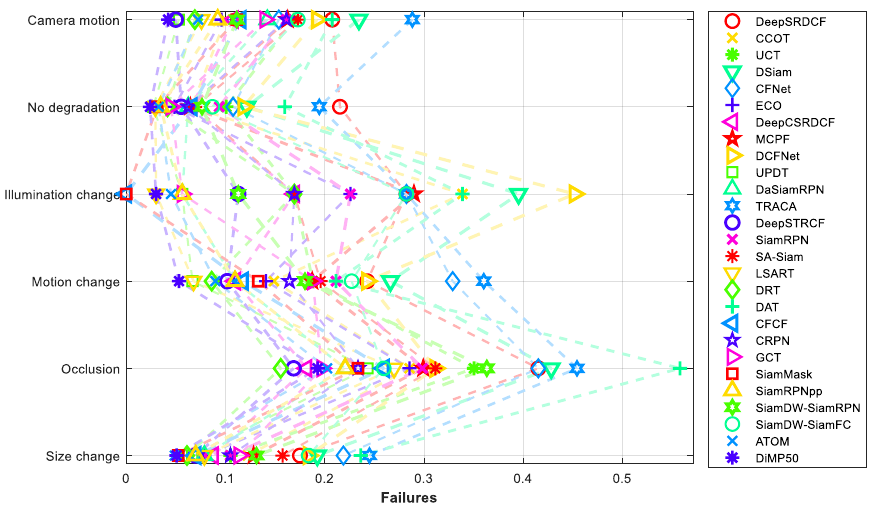}}
\vspace{-.7cm}
\caption{Performance comparison of visual tracking methods on VOT2018 dataset.}
 \label{fig:Results_VOT} 
\vspace{-2ex}
\end{figure*}

According to the results shown in Fig.~\ref{fig:Results_OTB_LaSOT}, the top-3 visual tracking methods in terms of the precision metric are the VITAL, MDNet, and DAT on the OTB2013 dataset, the SiamDW-SiamRPN, ASRCF, and VITAL on the OTB2015 dataset, and the PrDiMP50, DiMP50, and ATOM on the LaSOT dataset, respectively. In terms of success metric, the ASRCF, VITAL, and MDNet on the OTB2013 dataset, the SiamRPN++, SANet, and ASRCF on the OTB2015 dataset, and the PrDiMP50, DiMP50, and ATOM on the LaSOT dataset have achieved the best performance, respectively. On the VOT2018 dataset (see Fig.~\ref{fig:Results_VOT}), the top-3 visual trackers are the SiamMask, SiamRPN++, and DiMP50 in terms of accuracy measure while the PrDiMP50, DiMP50, and ATOM trackers have the best robustness, respectively. For the aerial-view tracking, the PrDiMP50, DiMP50, SiamRPN++, and SiamMask have provided the best results for average precision \& success metrics.
%----------------------------------------------------------------%
% \begin{figure*}[hbt!]
% \justify
% \begin{subfigure}[b]{0.38\textwidth}
% \includegraphics[width=6.3cm, height=3cm]{Fig4-1.pdf}
% \end{subfigure}%
% \hspace{-4mm}
% \begin{subfigure}[b]{0.38\textwidth}
% \includegraphics[width=6.3cm, height=3cm]{Fig4-2.pdf}
% \end{subfigure}
% \hspace{-6mm}
% \begin{subfigure}[b]{0.2\textwidth}
% \includegraphics[width=5cm, height=3cm]{Fig4-3.pdf}
% \end{subfigure}%
% %\vspace{-4mm}
% \justify
% \begin{subfigure}[b]{0.38\textwidth}
% \includegraphics[width=6.3cm, height=3cm]{Fig4-4.pdf}
% \end{subfigure}%
% \hspace{-4mm}
% \begin{subfigure}[b]{0.38\textwidth}
% \includegraphics[width=6.3cm, height=3cm]{Fig4-5.pdf}
% \end{subfigure}
% \hspace{-6mm}
% \begin{subfigure}[b]{0.2\textwidth}
% \includegraphics[width=5cm, height=3cm]{Fig4-6.pdf}
% \end{subfigure}
% \caption{Overall experimental comparison of state-of-the-art visual tracking methods on the OTB2013, OTB2015, and LaSOT visual tracking datasets.}
% \label{fig:Results_OTB_LaSOT} 
% \vspace{-2mm}
% \end{figure*}

%----------------------------------------------------------------%
% \begin{figure*}[hbt!]
% \justify
% \vspace{-2mm}
% \includegraphics[width=18.1cm, height=3.5cm]{Fig5.pdf}
% \vspace{-6mm}
%  \caption{Performance comparison of visual tracking methods on VOT2018 dataset.}
%  \label{fig:Results_VOT} 
% \vspace{-2ex}
% \end{figure*}

%----------------------------------------------------------------%
%\input{Writing/Table7.tex}
%----------------------------------------------------------------%
\indent On the other hand, the best trackers based on both precision-success measures (see Fig.~\ref{fig:Results_OTB_LaSOT}) are the VITAL, MDNet, and ASRCF on the OTB2013 dataset, the SiamRPN++, ASRCF, and VITAL on the OTB2015 dataset, the PrDiMP50, DiMP50, and ATOM on the LaSOT dataset, and the PrDiMP50, DiMP50, and SiamRPN++ on the aerial-view datasets (i.e., the UAV123, UAVDT, and VisDrone2019). On the VOT2018 dataset, the DiMP50, SiamRPN++, and ATOM are the best performing trackers based on the EAO score. Moreover, the PrDiMP50, DiMP50, SiamRPN++, and ATOM have achieved the best AUC scores while the SiamRPN, SiamRPN++, and CFNet are the fastest visual trackers, respectively. 
According to the results (i.e., Fig.~\ref{fig:Results_OTB_LaSOT}, and Fig.~\ref{fig:Results_VOT}), the best visual tracking methods that repeated the best results on different tracking datasets are the PrDiMP50 \cite{PrDiMP}, DiMP50 \cite{DiMP}, ATOM \cite{ATOM}, VITAL \cite{VITAL}, MDNet \cite{MDNet}, DAT \cite{DAT}, ASRCF \cite{ASRCF}, SiamDW-SiamRPN \cite{SiamDW}, SiamRPN++ \cite{SiamRPN++}, C-RPN \cite{CRPN}, StructSiam \cite{CRPN}, SiamMask \cite{SiamMask}, DaSiamRPN \cite{DaSiamRPN}, UPDT \cite{UPDT}, LSART \cite{LSART}, DeepSTRCF \cite{STRCF}, and DRT \cite{DRT}. These methods will be investigated in Sec.~\ref{Discuss}.
%----------------------------------------------------------------%
\vspace{-.2cm}
\subsection{Most Challenging Attributes per Benchmark Dataset} \label{Sec4.2}
\begin{figure*}[hbt!]
\justify
\subfigure{\includegraphics[width=3.6cm, height=2.5cm]{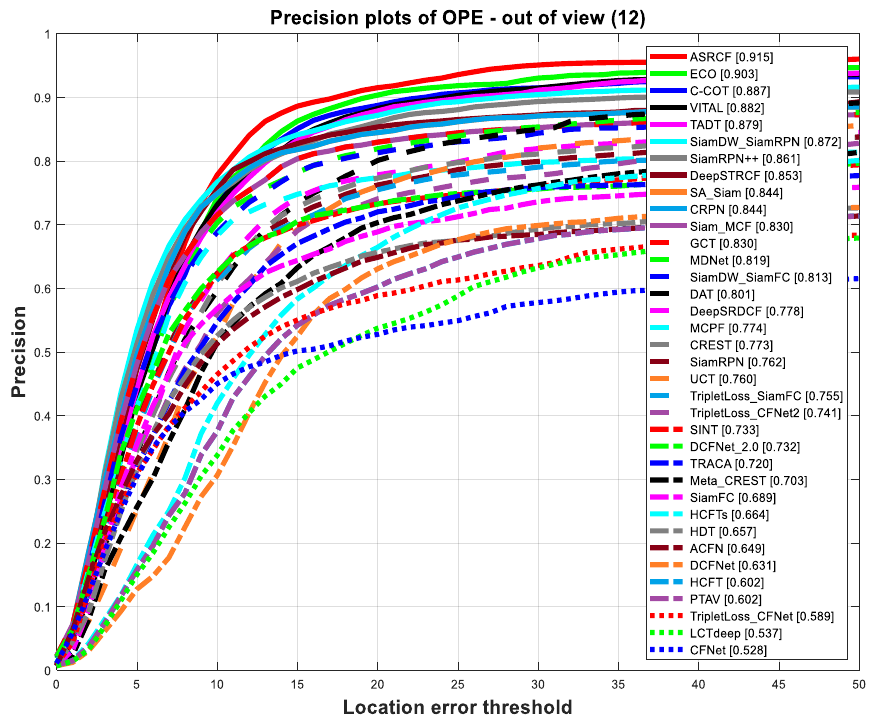}} 
\hspace{-.25cm}
\subfigure{\includegraphics[width=3.6cm, height=2.5cm]{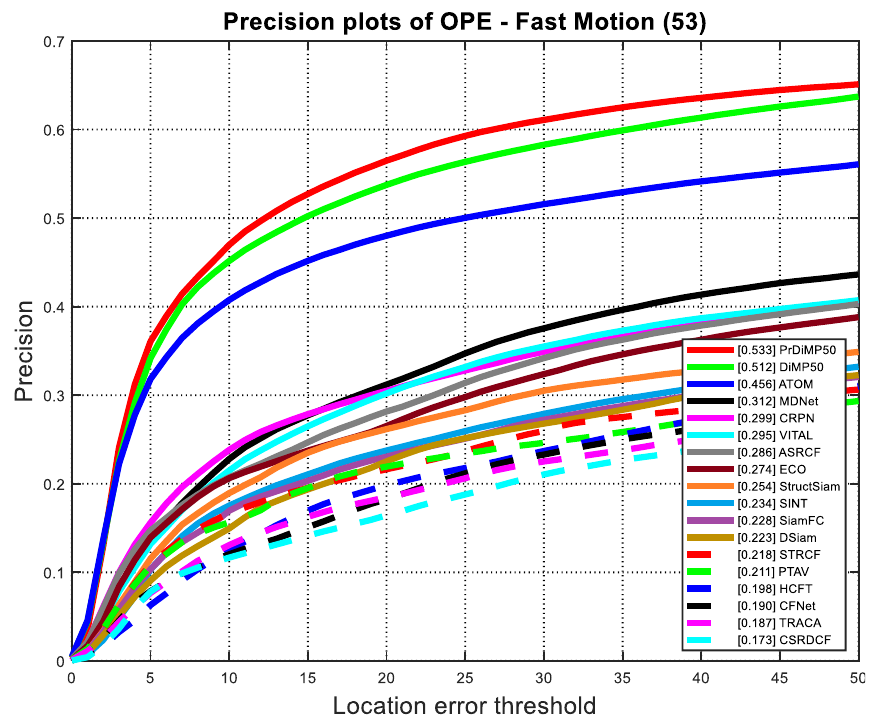}}
\hspace{-.25cm}
\subfigure{\includegraphics[width=3.6cm, height=2.5cm]{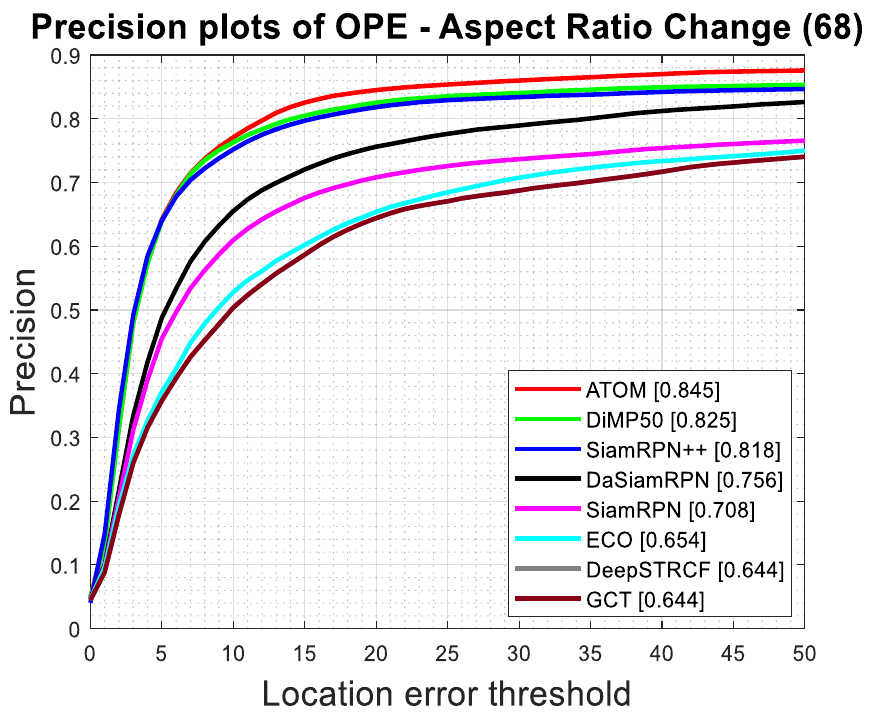}}
\hspace{-.25cm}
\subfigure{\includegraphics[width=3.6cm, height=2.5cm]{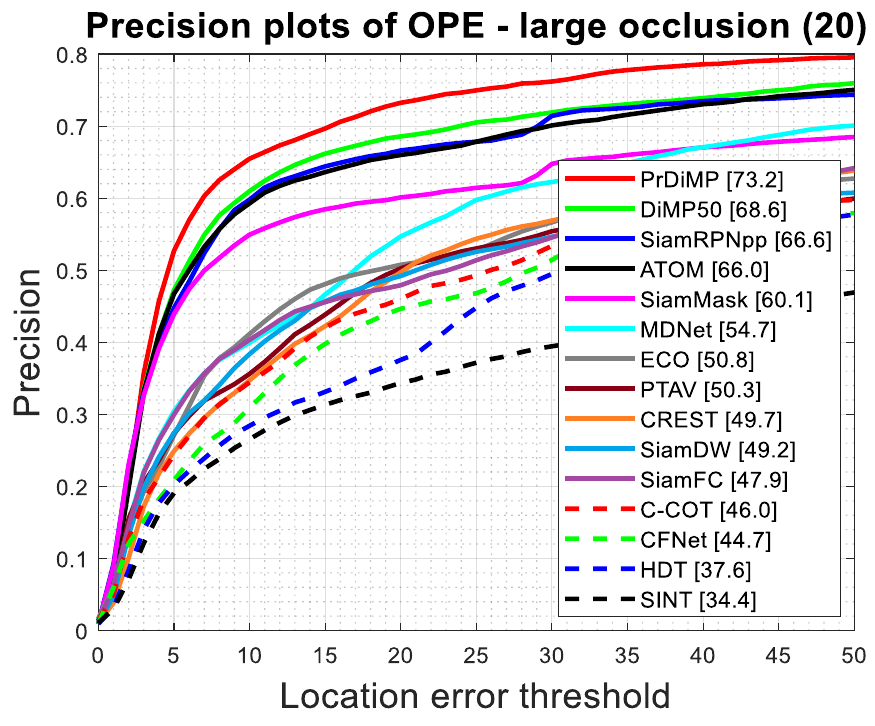}}
\hspace{-.25cm}
\subfigure{\includegraphics[width=3.6cm, height=2.5cm]{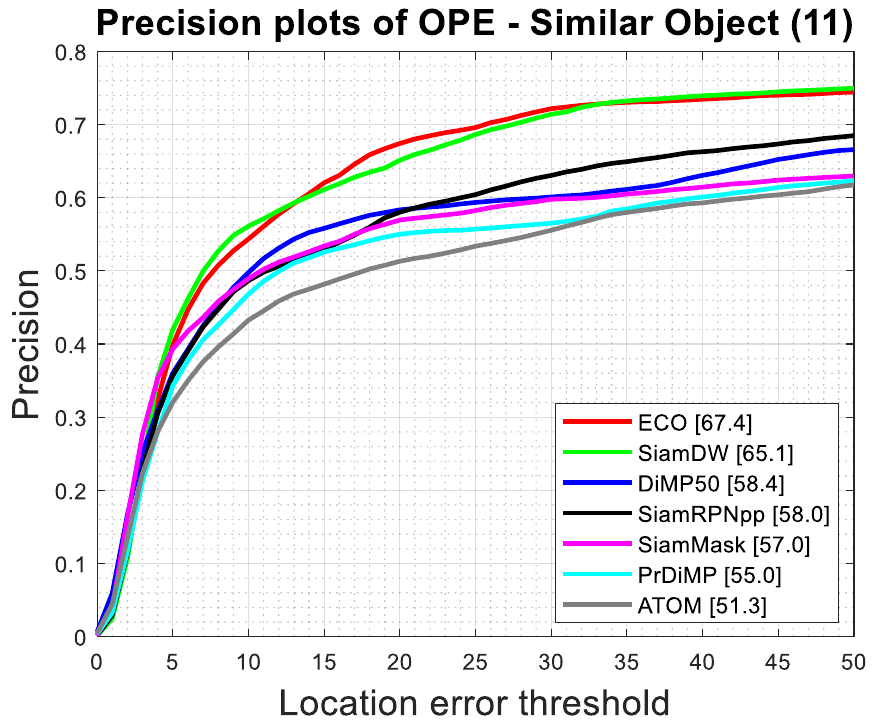}}
\vspace{-.5cm}
\justify
\subfigure{\includegraphics[width=3.6cm, height=2.5cm]{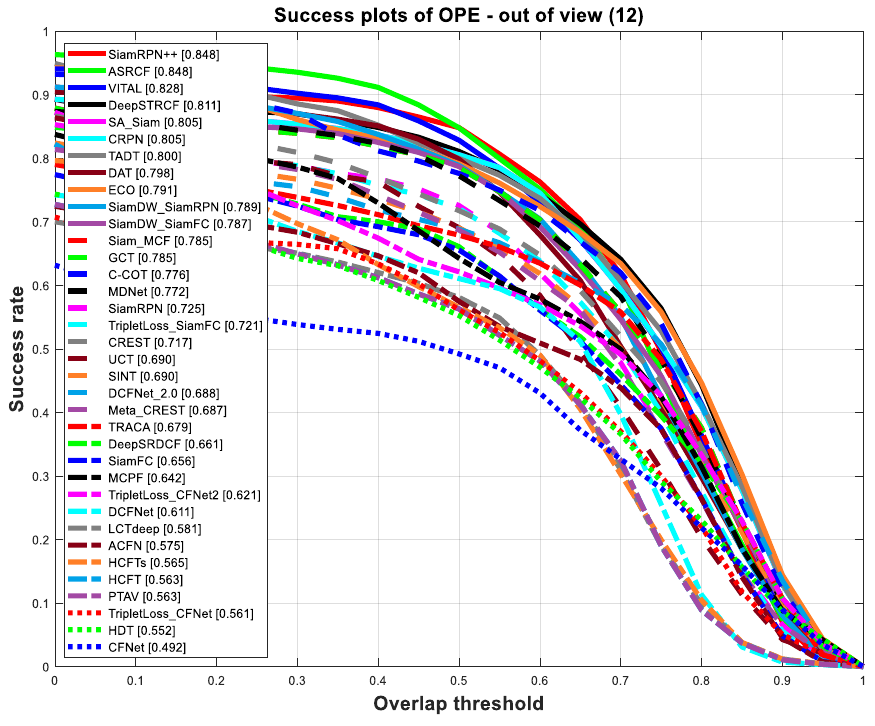}} 
\hspace{-.25cm}
\subfigure{\includegraphics[width=3.6cm, height=2.5cm]{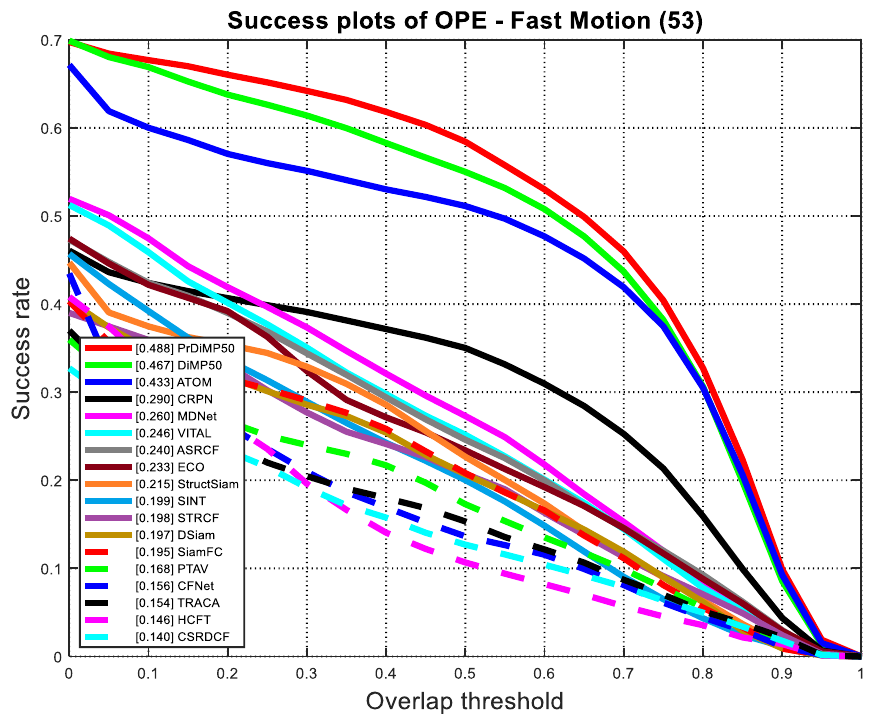}}
\hspace{-.25cm}
\subfigure{\includegraphics[width=3.6cm, height=2.5cm]{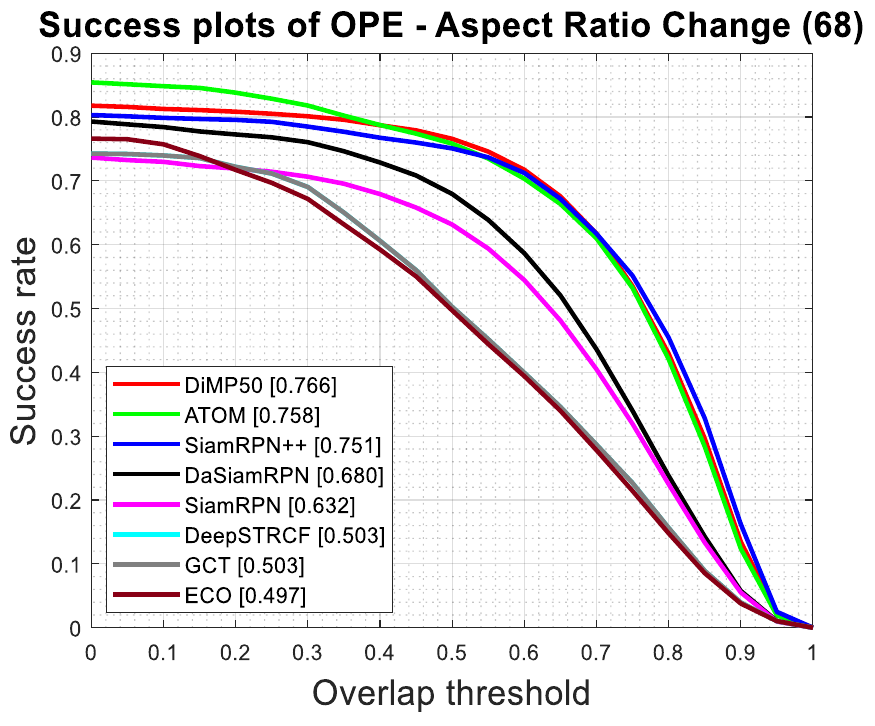}}
\hspace{-.25cm}
\subfigure{\includegraphics[width=3.6cm, height=2.5cm]{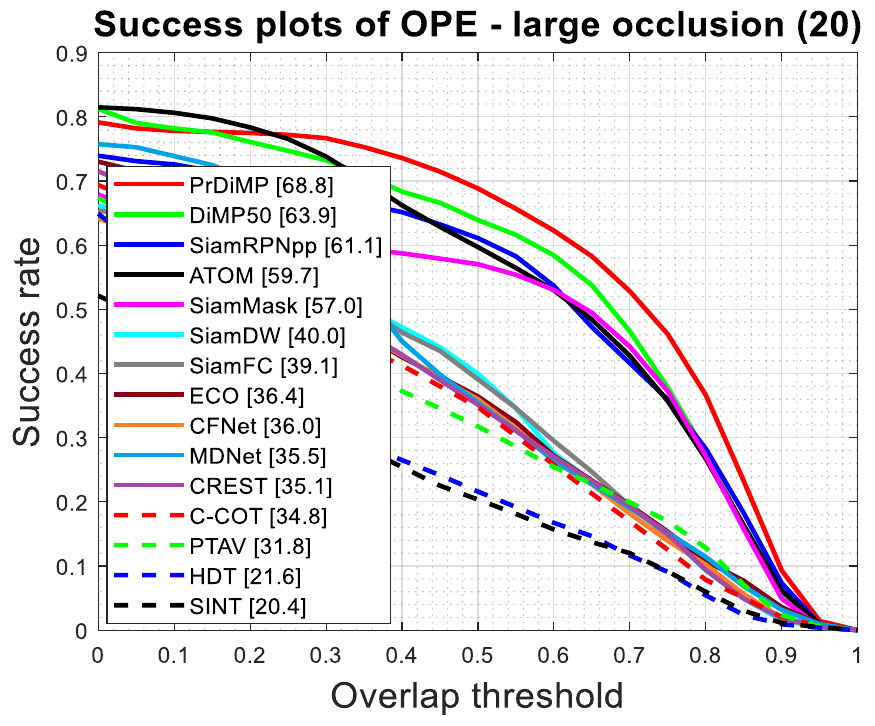}}
\hspace{-.25cm}
\subfigure{\includegraphics[width=3.6cm, height=2.5cm]{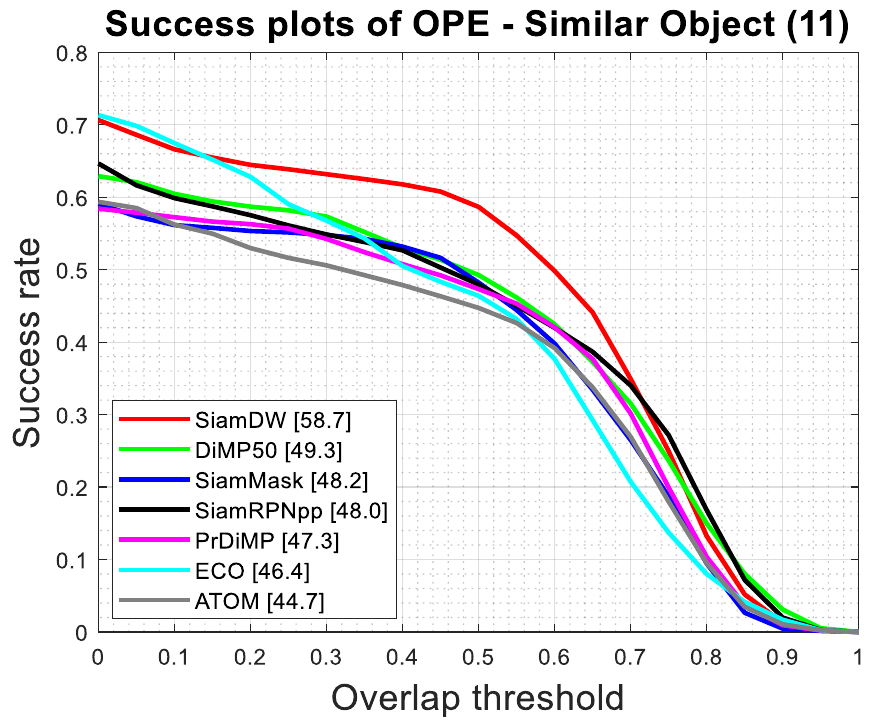}}
\vspace{-.3cm}
\caption{Comparison of state-of-the-art trackers in terms of the most challenging attributes on the OTB2015, LaSOT, UAV123, UAVDT, and VisDrone2019 datasets, left to right column, respectively.}
\label{fig:Att_Comp} 
\vspace{-2mm}
\end{figure*}

Following on the VOT challenges \cite{VOT-2016,VOT-2017,VOT-2018}, which have specified the most challenging visual tracking attributes, this work also introduces the most challenging attributes on the OTB, LaSOT, UAV123, UAVDT, and VisDrone datasets. These attributes are determined by the median accuracy \& robustness per attribute on the VOT or the median precision \& success per attribute on other datasets. Table~\ref{tab8} shows the most challenging attributes for each benchmark dataset. The OCC/LOC, OV, FM, DEF, LR, ARC, and SIB are selected as the most challenging attributes that can effectively impact the performance of DL-based visual trackers. Fig.~\ref{fig:Att_Comp} compares the performances of these methods on the most challenging attributes on the OTB2015, LaSOT, UAV123, UAVDT, and VisDrone2019 datasets.\\
%----------------------------------------------------------------%
\begin{table*}
\caption{Five most challenging attributes of benchmark datasets. [first to third challenging attributes are shown by \colorbox{red}{red}, \colorbox{yellow}{yellow}, and \colorbox{green}{green} colors.]} % title of Table
\centering % used for centering table
 \vspace{-2mm}
\resizebox{\textwidth}{!}{
\begin{tabular}{c c c c c c c c c c c c c c c c c c c c c c c c c c} % centered columns (4 columns)
\hline \hline 
Dataset & Metric & IV & DEF & MB & CM & OCC & POC & FOC & ROT & IPR & OPR & BC & VC & SV & FM & OV & LR & ARC & MC & SIB & OM & SOB & OB & LT & LOC  \\ \hline \hline

\multirow{2}{*}{OTB2015} & Precision & 0.7807 & \cellcolor{green!25}0.7382 & 0.7642 & - & \cellcolor{yellow!25}0.7347 & - & - & - & 0.7575 & 0.7611 & 0.7576 & - & 0.7471 & 0.7506 & \cellcolor{red!25}0.6911 & 0.7532 & - & -  & -  & -  & -  & -  & -  & - \\ \cline{2-26}
 & Success & 0.6330 & \cellcolor{red!25}0.5682 & 0.6466 & - & 0.6027 & - & - & - & 0.6154 & 0.6172 & 0.6144 & - & 0.6022 & 0.6268 & \cellcolor{yellow!25}0.5683 & \cellcolor{green!25}0.5906 & - & -  & -  & -  & -  & -  & -  & - \\ \hline \hline

\multirow{2}{*}{VOT2018} & Accuracy & \cellcolor{green!25}0.5026 & - & - & 0.5258 & \cellcolor{red!25}0.4312 & - & - & - & - & - & - & - & \cellcolor{yellow!25}0.4627 & - & - & - & - & 0.5044  & -  & -  & -  & -  & -  & - \\ \cline{2-26}
 & Robustness & \cellcolor{green!25}0.1695 & - & - & 0.1423 & \cellcolor{red!25}0.2856 & - & - & - & - & - & - & - & 0.1051 & - & - & - & - & \cellcolor{yellow!25}0.1802  & -  & -  & -  & -  & -  & - \\ \hline  \hline

\multirow{2}{*}{LaSOT} & Precision & 0.2839 & \cellcolor{green!25}0.1778 & 0.2149 & 0.2306 & - & 0.1937 & 0.1904 & 0.2016 & - & - & 0.2218 & 0.2034 & 0.2266 & \cellcolor{yellow!25}0.1733 & \cellcolor{red!25}0.1608 & 0.2248 & 0.2026 & -  & -  & -  & -  & -  & -  & - \\ \cline{2-26}
 & Success & 0.2580 & 0.2081 & 0.2216 & 0.2506 & - & 0.2112 & \cellcolor{yellow!25}0.1666 & 0.2186 & - & - & 0.2329 & 0.1773 & 0.2394 & \cellcolor{red!25}0.1398 & \cellcolor{green!25}0.1726 & 0.1772 & 0.2119 & -  & -  & -  & -  & -  & -  & - \\ \hline  \hline
 
\multirow{2}{*}{UAV123} & Precision & 0.6573 & - & - & 0.6317 & - & 0.6991 & 0.6805 & - & - & - & 0.6361 & .6462 & 0.6656 & \cellcolor{yellow!25}0.6019 & 0.6429 & \cellcolor{green!25}0.6100 & \cellcolor{red!25}0.5690 & -  & 0.6991  & -  & -  & -  & -  & - \\ \cline{2-26}
 & Success & 0.4926 & - & - & \cellcolor{yellow!25}0.4034 & - & 0.5447 & 0.5201 & - & - & - & 0.4673 & 0.4794 & 0.5176 & \cellcolor{green!25}0.4158 & 0.4743 & 0.4724 & \cellcolor{red!25}0.3433 & -  & 0.5447 & -  & -  & -  & -  & - \\ \hline  \hline
 
\multirow{2}{*}{UAVDT} & Precision & 0.7723 & - & - & \cellcolor{green!25}0.6877 & - & - & - & - & - & - & \cellcolor{yellow!25}0.6634 & - & 0.7003 & - & - & - & - & -  & 0.7638 & 0.7039 & 0.7638 & 0.7280 & 0.8235 & \cellcolor{red!25}0.5779 \\ \cline{2-26}
 & Success & 0.5844 & - & - & 0.5515 & - & - & - & - & - & - & \cellcolor{green!25}0.5051 & - & 0.5603 & - & - & - & - & -  & 0.5536 & 0.5513 & 0.5536 & \cellcolor{yellow!25}0.5463 & 0.6213 & \cellcolor{red!25}0.4600 \\ \hline  \hline
 
\multirow{2}{*}{VisDrone2019} & Precision & 0.7790 & - & - & 0.7407 & - & 0.7174 & 0.6919 & - & - & - & \cellcolor{yellow!25}0.5775 & 0.8058 & 0.7392 & 0.7473 & 0.8088 & \cellcolor{green!25}0.6031 & 0.7522 & -  & \cellcolor{red!25}0.5445 & -  & -  & -  & -  & - \\ \cline{2-26}
 & Success & 0.6272 & - & - & 0.5816 & - & 0.5511 & 0.5407 & - & - & - & \cellcolor{green!25}0.4112 & 0.6564 & 0.5891 & 0.5862 & 0.6676 & \cellcolor{red!25}0.3681 & 0.5861 & -  & \cellcolor{yellow!25}0.3914 & -  & -  & -  & -  & - \\ \hline  \hline
 
  \hline %inserts single line
  \end{tabular}
  \label{tab8}
 }
 \vspace{-2mm}
\end{table*}
%----------------------------------------------------------------%
%----------------------------------------------------------------%
%\input{Writing/Fig7.tex}
%----------------------------------------------------------------%
According to the OCC attribute, the most accurate \& robust visual trackers on the VOT2018 dataset are the SiamRPN++ \cite{SiamRPN++} and DRT \cite{DRT}, respectively. In terms of success metric, the SiamRPN++ \cite{SiamRPN++} is the best visual tracker to tackle the DEF and OV attributes, while the Siam-MCF \cite{Siam-MCF} is the best one to deal with the visual tracking in LR videos on the OTB2015 dataset. 
The ASRCF \cite{ASRCF}, ECO \cite{ECO}, and SiamDW-SiamRPN \cite{SiamDW} are the best trackers in precision metric to face with OV, OCC, and DEF attributes on the OTB-2015 dataset. 
The PrDiMP50 \cite{PrDiMP}, DiMP50 \cite{DiMP}, and ATOM \cite{ATOM} trackers are the absolute best methods on the LaSOT dataset on all visual attributes.
On the UAV123 dataset, the ATOM and DiMP50 have achieved the best results in terms of precision and success metrics, respectively. Also, the PrDiMP50 is the best tracker for handling large occlusions on the UAVDT dataset. Finally, the SiamDW is the best tracker to tackle similar objects on the VisDrone2019-test-dev dataset.\\
\indent As shown in Fig.~\ref{fig:Results_VOT}, the DCF-based methods have achieved fewer failures among the other methods, while the SNN- \& custom-network based trackers have gained more overlap between the estimated BBs and ground-truth ones. The SiamRPN-based methods (i.e., \cite{SiamMask,SiamRPN++,DaSiamRPN,SiamDW}) accurately handle scenarios under each of CM, IV, MC, OCC, or SC attributes by adopting deeper and wider backbone networks, including classification and regression branches. Moreover, the ATOM, DiMP50, and PrDiMP50 exploit powerful classification \& regression networks and optimization processes for online training and fast adaptation. Thus, these trackers have provided significant advances on various tracking benchmarks.
By considering the fusion of hand-crafted and deep features \cite{STRCF,UPDT,DRT}, temporal regularization term \cite{STRCF}, reliability term \cite{DRT}, data augmentation \cite{UPDT}, and exploitation of ResNet-50 model \cite{UPDT}, the DCF-based methods have attained desirable robustness against CM attribute. Furthermore, the computational efficiency and the robustness of DCF-based trackers are attractive for aerial-view trackers.\\
\indent To effectively deal with the IV attribute, focusing on the discrimination power between the target and its background is the main problem. The strategies such as training a fully convolutional network for correlation filter cost function, spatial-aware KRR and spatial-aware CNN, and employing semi-supervised video object segmentation improve the robustness of DL-based trackers when significant IV occurs. To robustly deal with MC and OCC attributes, the DCF- and CNN-based trackers have performed the best. However, the SNN-based methods with the aid of region proposal subnetwork and proposal refinement can robustly estimate the tightest BB under severe scale changes. However, recently, the IoU-based refinement network (based on IoU-Net \cite{IoUNet}) employed in ATOM, DiMP, and PrDiMP trackers can effectively handle aspect-ratio change of target during tracking. 
%----------------------------------------------------------------%
\vspace{-.3cm}
\subsection{Discussion} \label{Discuss} \label{Sec4.3}
The overall best methods (i.e., PrDiMP50 \cite{PrDiMP}, DiMP50 \cite{DiMP}, ATOM \cite{ATOM}, VITAL \cite{VITAL}, MDNet \cite{MDNet}, DAT \cite{DAT}, ASRCF \cite{ASRCF}, SiamDW-SiamRPN \cite{SiamDW}, SiamRPN++ \cite{SiamRPN++}, C-RPN \cite{CRPN}, StructSiam \cite{StructSiam}, SiamMask \cite{SiamMask}, DaSiamRPN \cite{DaSiamRPN}, UPDT \cite{UPDT}, LSART \cite{LSART}, DeepSTRCF \cite{STRCF}, and DRT \cite{DRT}) belong to a wide range of network architectures. For instance, the MDNet, LSART, and DAT (uses the MDNet architecture) utilize CNNs to localize a visual target while the ASRCF, UPDT, DRT, and DeepSTRCF exploit deep off-the-shelf features. All the ATOM, DiMP, and PrDiMP trackers employ custom classification \& refinement networks. Besides the VITAL that is a GAN-based tracker, the C-RPN, StructSiam, SiamMask, DaSiamRPN, SiamDW, and SiamRPN++ have the SNN architecture. 
Although the most recent attractive deep architectures for visual tracking are based on Siamese or custom networks, GAN- and RL-based trackers have been recently developed for some specific purposes, such as addressing the imbalance distribution of training samples \cite{VITAL} or selecting an appropriate real-time search strategy \cite{ACT,DRRL}. GAN-based trackers can successfully augment positive samples to enrich the target appearance model. These trackers also enjoy cost-sensitive losses to focus on hard negative samples. RL-based trackers learn continuous actions to provide more reliable search \& verification strategies for visual trackers. Besides, the combinations of RL-trackers with other architectures may add more advantages; for instance, recurrent RL-based tracking considers time dependencies to the key components (i.e., actions \& states). By doing so, these trackers boost their performance by verifying confidence through an RNN motion model. \\
\indent In addition to providing a desirable balance between the performance and speed of Siamese or custom network-based trackers, the architectures are modified to integrate with diverse deep backbone networks, searching strategies, and learning schemes but also exploit fully convolutional networks, correlation layers, region proposal networks, video object detection/segmentation modules. The interesting point is that five SNN-based methods including the SiamDW-SiamRPN, SiamRPN++, C-RPN, SiamMask, and DaSiamRPN are based on the fast SiamRPN method \cite{SiamRPN}, which is consisted of Siamese subnetwork and region proposal subnetwork; these subnetworks are leveraged for feature extraction and proposal extraction on correlation feature maps to solve the visual tracking problem by one-shot detection task. The main advantages of SiamRPN are the time efficiency and precise estimations with integrating proposal selection and refinement strategies into a Siamese network.\\
\indent Interestingly, the ASRCF, UPDT, DRT, and DeepSTRCF, which exploit deep off-the-shelf features, are among the top-performing visual tracking methods. Moreover, five methods of UPDT, DeepSTRCF, DRT, LSART, and ASRCF take the advantages of the DCF framework. On the other side, the best performing visual trackers, namely PrDiMP50, DiMP50, ATOM, VITAL, MDNet, DAT, SiamDW, SiamRPN++, C-RPN, StructSiam, SiamMask, DaSiamRPN, and LSART exploit specialized deep features for visual tracking purpose. Although diversified backbone networks are employed for these methods, state-of-the-art methods have been leveraging deeper networks such as the ResNet-50 to strengthen the discriminative power of target modeling. From the network training perspective, the SiamDW-SiamRPN, SiamRPN++, C-RPN, StructSiam, SiamMask, and DaSiamRPN use offline training, the LSART utilizes online training, and the PrDiMP50, DiMP50, and ATOM take offline \& online training procedures. In particular, the PrDiMP50 and DiMP50 exploit meta-learning based networks to improve network adaptation for the tracking task. The offline trained trackers aim to provide dominant representations to achieve real-time tracking speed. Handling significant appearance variations needs to adapt to network parameters during tracking, but online training has an over-fitting risk because of limited training samples. Hence, the VITAL, MDNet, and DAT by employing adversarial learning, domain-independent information, and attention maps as regularization terms benefit both offline and online training of DNNs. However, these methods provide a tracking speed of about one \textit{frame per second} (FPS) that is not suitable for real-time applications. In contrast, recent proposed PrDiMP50, DiMP50, and ATOM trackers exploit custom-designed networks, efficient optimization strategies to achieve acceptable tracking speed.
From the perspective of the objective function of DNNs, the VITAL and StructSiam are classification-based, the LSART is regression-based, and the other best-performing trackers \cite{MDNet,DaSiamRPN,DAT,CRPN,SiamDW,SiamMask,SiamRPN++,PrDiMP,DiMP,ATOM} employ both classification and regression objectives. For instance, five modified versions of the SiamRPN \cite{SiamRPN} (i.e., SiamDW-SiamRPN \cite{SiamDW}, SiamRPN++ \cite{SiamRPN++}, C-RPN \cite{CRPN}, SiamMask \cite{SiamMask}, and DaSiamRPN \cite{DaSiamRPN}) have two branches for classification and regression. Besides, the ATOM, DiMP50, and PrDiMP50 use a classification network for distinguishing target from the background and an IoU-Net for BB regression.\\
\indent Based on the motivation categorization of the best trackers, the recent advanced methods rely on 1) alleviating the imbalanced distribution of visual training data by the data augmentation \cite{UPDT,DaSiamRPN} and generative network from adversarial learning \cite{VITAL}, 2) efficient training and learning procedures by reformulating classification/regression problems \cite{UPDT,DaSiamRPN,VITAL,STRCF,DRT,LSART,ASRCF,ATOM,DiMP,PrDiMP} and providing specified features for visual tracking \cite{MDNet,DaSiamRPN,StructSiam,VITAL,DAT,CRPN,SiamDW,SiamMask,SiamRPN++,ATOM,DiMP,PrDiMP}, 3) exploiting state-of-the-art architectures to provide more discriminative representations by leveraging ResNet models as the backbone networks \cite{UPDT,SiamDW,SiamMask,SiamRPN++,ATOM,DiMP,PrDiMP}, and 4) extracting complementary features by employing additional information such as contextual \cite{UPDT,DaSiamRPN,StructSiam,SiamDW} or temporal information \cite{DaSiamRPN,VITAL,STRCF,DAT}. The VITAL, DaSiamRPN, and UPDT attempt to alleviate the imbalanced distribution of positive and negative training data samples and extract more discriminative features. The VITAL uses adversarial learning to augment positive samples and decrease simple negative ones and preserve the most discriminative and robust features during tracking. Furthermore, the DaSiamRPN utilizes both data augmentation and negative semantic samples to consider visual distractors and improve visual tracking robustness. The UPDT uses standard data augmentation and a quality measure for estimated states to fuse shallow and deep features effectively. Finally, the ATOM employs standard data augmentation to improve its online adaptation, while the DiMP50 \& PrDiMP50 trackers enjoy meta-learning strategies to form their training set. \\
\indent To improve the learning process of the best DL-based methods, the ATOM, UPDT, DeepSTRCF, DRT, LSART, and ASRCF have revised the conventional ridge regression of DCF formulation. Moreover, the DaSiamRPN and VITAL utilize the distractor-aware objective function and reformulated objective function of GANs using a cost-sensitive loss to improve the training process of these visual trackers, respectively. Finally, the PrDiMP tracker computes the similarity of predictive and ground-truth distributions by \textit{Kullback-Leibler} (KL) divergence.
Training of DL-based methods on large-scale datasets adapts their network function for visual tracking. The SiamDW, SiamRPN++, and SiamMask methods have aimed to leverage state-of-the-art deep networks as a backbone network of Siamese trackers. The ATOM, DiMP50, and PrDiMP tracker employ ResNet blocks as the backbone network, while the DiMP \& PrDiMP train these blocks on tracking datasets. While these methods exploit ResNet models, the SiamDW proposes new residual modules and architectures to prevent significant receptive field increase and simultaneously improve feature discriminability and localization accuracy. Also, the ResNet-driven SNN-based tracker proposed by the SiamRPN++ includes different layer-wise and depth-wise aggregations to fill the performance gap between SNN-based and CNN-based methods. In addition to the spatial information, the DAT (using reciprocative learning) and DeepSTRCF (using online \textit{passive-aggressive} (PA) learning) also consider temporal information in different ways to provide more robust features. Generally, six learning schemes of the similarity learning (i.e., SiamDW, SiamRPN++, C-RPN, StructSiam, SiamMask, DaSiamRPN, ATOM), blue{meta-learning (i.e., PrDiMP50, DiMP50)}, multi-domain learning (i.e., MDNet, DAT), adversarial learning (i.e., VITAL), spatial-aware regressions learning (i.e., LSART), and DCF learning are utilized. \\
\indent In the following, the best visual tracking methods are studied based on their advantages and disadvantage. The ATOM, DiMP, and PrDiMP trackers consider visual tracking as two-step classification and target estimation procedures. These trackers are robust to handle CM, MC, SV, and ARC attributes by employing custom networks and elaborated optimization strategies. However, the SOB, LR, and OB attributes can dramatically impact their performances.
Three SNN-based methods of the C-RPN, StructSiam, and DaSiamRPN exploit the shallow AlexNet as their backbone network (see Table~\ref{tab2}), which is the main weakness of these trackers according to their discriminative power. To improve tracking robustness in the presence of significant SV and visual DI, the C-RPN cascades multiple RPNs in a Siamese network to exploit from hard negative sampling (to provide more balanced training samples), multi-level features, and multiple steps of regressions. To decrease the sensitivity of SNN-based methods specifically for non-rigid appearance change and POC attributes, the StructSiam detects contextual information of local patterns and their relationships and matches them by a Siamese network in real-time speed. By adopting the local-to-global search strategy and the \textit{non-maximum suppression} (NMS) to re-detect target and reduce potential distractors, the DaSiamRPN correctly handles the FOC, OV, POC, and BC challenges. In contrast, the SiamMask, SiamDW-SiamRPN, and SiamRPN++ exploit the ResNet models. To rely on rich target representation, the SiamMask uses three-branch architecture to estimate the target location by a rotated BB, including the target's binary mask. The most failure reasons for SiamMask are the MB \& OV attributes that produce erroneous target masks. To reduce the performance margin of the SNN-based methods with state-of-the-art visual tracking methods, the SiamDW-SiamRPN and SiamRPN++ study the exploitation of deep backbone networks to reduce the sensitivity of these methods to the most challenging attributes.\\
\indent The MDNet and the other methods based on it (e.g., DAT) are still among the best visual tracking methods. Because of specialized offline and online training of these networks on large-scale visual tracking datasets, these methods can handle various challenging situations, hardly miss the visual targets, and have a satisfactory performance to track LR targets. However, these methods suffer from high computational complexity, intra-class discrimination of targets with similar semantics, and performing discrete space for scale estimation. The VITAL can tolerate massive DEF, IPR, and OPR because it focuses on hard negative samples through high-order cost-sensitive loss. However, it does not have a robust performance in the case of significant SV due to the producing a fixed size of weight mask via a generative network. The LSART utilizes the modified \textit{Kernelized ridge regression} (KRR) by the weighted combination of patch-wise similarities to concentrate on the target's reliable regions. Due to the consideration of rotation information and online adaptation of CNN models, this method provides promising responses to tackle the DEF and IPR challenges. \\
%----------------------------------------------------------------%
\begin{figure*}[hbt!]
\centering
\vspace{-2mm}
\includegraphics[width=0.98\linewidth]{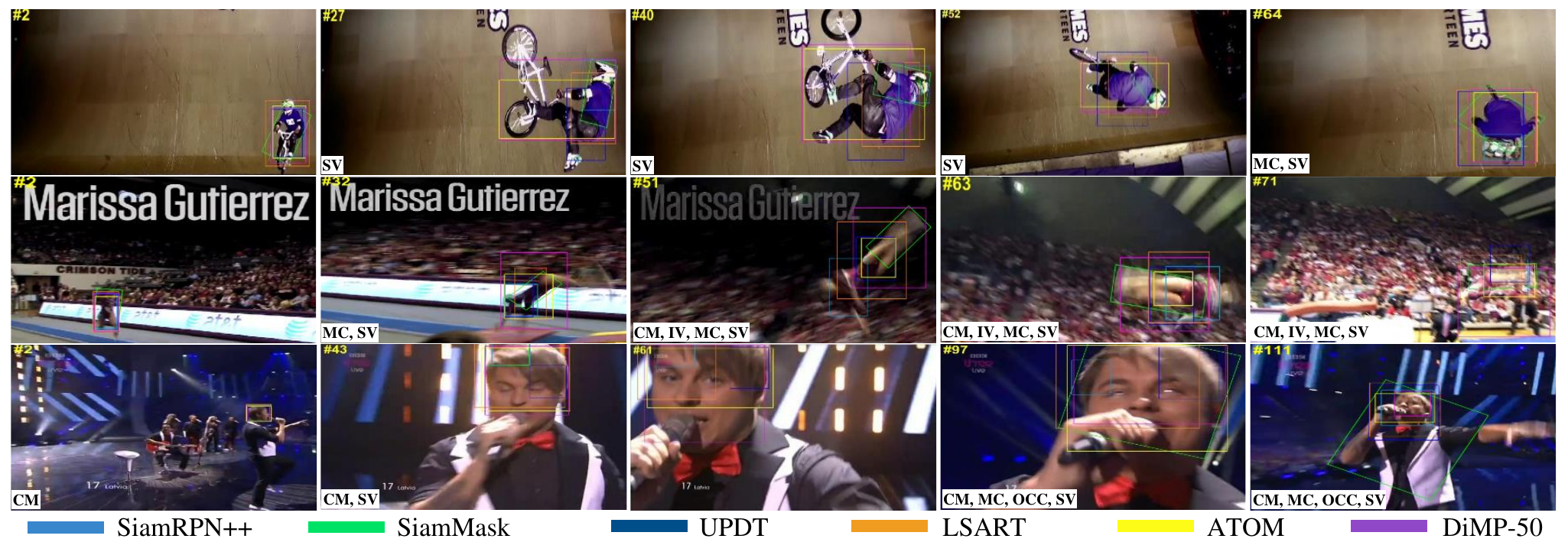}
\vspace{-.5cm}
\caption{Qualitative comparison of state-of-the-art visual trackers on the BMX, Gymnastics3, and Singer3 video sequences from VOT2018 dataset. The \#frame number and annotated attributes are shown on each frame.}
\label{fig:QualitativeComparison}
\vspace{-2ex}
\end{figure*}
%----------------------------------------------------------------%
The DeepSTRCF, ASRCF, DRT, and UPDT are the DCF-based methods that exploit deep off-the-shelf features and fuse them with shallow ones (e.g., HOG and CN) to improve the robustness of visual tracking (see Table~\ref{tab1}). To reduce the adverse impact of the OCC and OV attributes, the DeepSTRCF adds a temporal regularization term to the spatially regularized DCF formulation. The revisited formulation helps the DeepSTRCF enduring some appearance variations such as the IV, IPR, OPR, and POC. Using object-aware spatial regularization and reliability terms, the ASRCF and DRT methods attempt to optimize models to effectively learn adaptive correlation filters. Both these methods have studied major imperfections of DCF-based methods such as circular shifted sampling process, same feature space for localization and scale estimation processes, the strict focus on discrimination, and sparse and non-uniform distribution of correlation responses. Hence, these methods handle the DEF, BC, and SV, suitably. Finally, the UPDT focuses on enhancing the visual tracking robustness through independently training a shallow feature-based DCF and a deep off-the-shelf feature-based DCF and considering augmented training samples with an adaptive fusion model. Although these methods demonstrate the competitive performance of well-designed DCF-based trackers compared to more sophisticated trackers, they suffer from the limitations of pre-trained models, aspect ratio variation, model degradation, and considerable appearance variation. \\
\indent Finally, we have modified the VOT toolkit to be able to compare state-of-the-art visual trackers qualitatively. Fig.~\ref{fig:QualitativeComparison} shows the tracking results of the SiamRPN++ \cite{SiamRPN++}, SiamMask \cite{SiamMask}, LSART \cite{LSART}, UPDT \cite{UPDT}, ATOM \cite{ATOM}, and DiMP50 \cite{DiMP} on some video sequences of the VOT2018 dataset (modified toolkit \& all videos are publicly available on the aforementioned page). According to the achieved results, the DiMP50, ATOM, and SiamRPN++ have provided the best results. However, failures usually happen when multiple critical attributes simultaneously occur in a scene. For instance, the SiamMask misuses the semi-supervised video object segmentation when the OCC and SV co-occur, or the significant SV dramatically reduces the performance of the SiamRPN++. Despite considerable advances that are emerged in visual tracking, the state-of-the-art visual trackers are still unable to handle serious real-world challenges; severe variations of target appearance, MOC, OCC, SV, CM, DEF, and even IV can have drastic effects on the performance, which may lead to tracking failures. These results demonstrate that the visual trackers are still not completely reliable for real-world applications because they lack the intelligence for scene understanding. Current trackers improve object-scene distinction, but they cannot infer scene information, immediately recognize the global/configural structure of a scene, or organize purposeful decisions based on space and acts within.
%----------------------------------------------------------------%
\vspace{-.3cm}
\section{Conclusion and Future Directions}\label{sec:5}
The state-of-the-art DL-based visual trackers were categorized into a comprehensive taxonomy based on network architecture, network exploitation, training, network objective, network output, the exploitation of correlation filter advantages, aerial-view tracking, long-term tracking, and online tracking. Moreover, the motivations and contributions of these methods were categorized according to the main problems and proposed solutions of DL-based trackers. Furthermore, almost all visual tracking benchmark datasets and evaluation metrics were briefly investigated, and the various state-of-the-art trackers were compared on seven visual tracking datasets.\\
\indent Recently, the DL-based visual tracking methods have investigated different exploitation of deep off-she-shelf features, fusion of deep features \& hand-crafted features, various architectures \& backbone networks, offline \& online training of DNNs on large-scale datasets, update schemes, search strategies, contextual information, temporal information, and how to deal with lacking training data. However, many problems are not precisely solved, and also other problems need to be explored in the future. In the following, some of these future directions are presented for more investigation.\\
\indent First, the main concentration is to design custom neural networks to provide robustness, accuracy, and efficiency simultaneously. These trackers are primarily developed by integrating efficient network architectures with either classification \& regression branches or two-step classification \& BB refinement networks. Most recent works do not re-train/fine-tune their backbone networks to exploit generic features and avoid catastrophic forgetting of general patterns. However, diverse machine learning-based techniques to address this issue have been proposed, such as incremental learning \cite{Forget_HardAttn}, transfer learning penalties \cite{Forget_TransferPenalty}, batch spectral shrinkage \cite{Forget_BSS}, or lifelong learning \cite{Forget_lifelong}. Thus, effective training of backbone networks can boost tracking performance. \\
% Second, most state-of-the-art visual trackers have problems tracking targets in the presence of OB, LR, SIB, and SOB attributes.
\indent Second, generic visual trackers are required to adapt to unseen targets quickly. Hence, efficient online training of neural networks is crucial. Recently, meta-/few-shot learning approaches are mainly used to find an optimal initialization of the base learner to a new target. But, the meta-networks need to be shallow to avoid over-fitting problems. Therefore, exploring effective few-shot learning approaches provides fast convergence of deeper networks. \\
\indent Third, tracking from aerial-views introduces additional challenges for visual tracking. For instance, small/tiny object tracking in videos captured from medium/high-altitudes, severe viewpoint changes, and tracking many targets in dense environments should be considered. Furthermore, these scenarios are consistently involved with out-of-view and large occlusions; thus, developing long-term approaches will help more reliable aerial-view trackers. \\
\indent Fourth, long-term trackers are overlooked despite many advances in short-term trackers. In fact, long-term trackers are closer to practical, real-world scenarios when the target may disappear frequently or occlude for a long time. These trackers should have the ability to re-detect the target once a failure occurs and then continue tracking the correct target during video sequences. Thus, compelling detection \& verification networks are needed to be designed. \\
\indent Finally, existing visual trackers have a deficiency in scene understanding. The state-of-the-art methods cannot interpret dynamic scenes in a meaningful way, immediately recognize global structures, infer existing objects, and perceive basic level categories of different objects or events. Although recent trackers desirably reduce the computational complexity, these trackers can be modified to employ complementary features (e.g., temporal information) and incorporate proposed adversarial learning contributions in this few-data regime task.
%----------------------------------------------------------------%
% use section* for acknowledgment
 %
 \ifCLASSOPTIONcompsoc
   % The Computer Society usually uses the plural form
% \begin{comment}
\section*{Acknowledgments}
%   This work was partly supported by a grant (No. $96013046$) from \textit{Iran National Science Foundation} (INSF). We wish to thank Prof. Kamal Nasrollahi (Visual Analysis of People Lab (VAP), Aalborg University, Denmark) for his beneficial comments.
% \end{comment}
 \else
   % regular IEEE prefers the singular form
   \section*{Acknowledgments}
  We wish to thank Prof.~Kamal~Nasrollahi (Visual Analysis of People Lab (VAP), Aalborg University, Denmark) for his beneficial comments.
 \fi
% 
% The authors would like to thank...
%----------------------------------------------------------------%
\bibliographystyle{IEEEtran}
% \vspace{-4mm}
\bibliography{ref}

% Generated by IEEEtran.bst, version: 1.14 (2015/08/26)
\begin{thebibliography}{100}
\providecommand{\url}[1]{#1}
\csname url@samestyle\endcsname
\providecommand{\newblock}{\relax}
\providecommand{\bibinfo}[2]{#2}
\providecommand{\BIBentrySTDinterwordspacing}{\spaceskip=0pt\relax}
\providecommand{\BIBentryALTinterwordstretchfactor}{4}
\providecommand{\BIBentryALTinterwordspacing}{\spaceskip=\fontdimen2\font plus
\BIBentryALTinterwordstretchfactor\fontdimen3\font minus
  \fontdimen4\font\relax}
\providecommand{\BIBforeignlanguage}[2]{{%
\expandafter\ifx\csname l@#1\endcsname\relax
\typeout{** WARNING: IEEEtran.bst: No hyphenation pattern has been}%
\typeout{** loaded for the language `#1'. Using the pattern for}%
\typeout{** the default language instead.}%
\else
\language=\csname l@#1\endcsname
\fi
#2}}
\providecommand{\BIBdecl}{\relax}
\BIBdecl

\bibitem{ITS-AutVehicle}
M.~{Gao}, L.~{Jin}, Y.~{Jiang}, and B.~{Guo}, ``Manifold {Siamese} network: A
  novel visual tracking {ConvNet} for autonomous vehicles,'' \emph{IEEE Trans.
  Intell. Transp. Sys.}, pp. 1--12, 2019.

\bibitem{SurveyMultiRobot}
C.~Robin and S.~Lacroix, ``{Multi-robot target detection and tracking: Taxonomy
  and survey},'' \emph{Autonomous Robots}, vol.~40, no.~4, pp. 729--760, 2016.

\bibitem{ITS-HumanTracking}
K.~{Lee}, J.~{Hwang}, G.~{Okopal}, and J.~{Pitton},
  ``Ground-moving-platform-based human tracking using visual {SLAM} and
  constrained multiple kernels,'' \emph{IEEE Trans. Intell. Transp. Sys.},
  vol.~17, no.~12, pp. 3602--3612, 2016.

\bibitem{Ababsa2008}
F.~Ababsa, M.~Maidi, J.~Y. Didier, and M.~Mallem, ``{Vision-based tracking for
  mobile augmented reality},'' in \emph{Studies in Computational
  Intelligence}.\hskip 1em plus 0.5em minus 0.4em\relax Springer, 2008, vol.
  120, pp. 297--326.

\bibitem{ReviewUAV}
J.~Hao, Y.~Zhou, G.~Zhang, Q.~Lv, and Q.~Wu, ``{A review of target tracking
  algorithm based on UAV},'' in \emph{Proc. IEEE CBS}, 2019, pp. 328--333.

\bibitem{SurveySoccer}
M.~Manafifard, H.~Ebadi, and H.~{Abrishami Moghaddam}, ``{A survey on player
  tracking in soccer videos},'' \emph{Comput. Vis. Image Und.}, vol. 159, pp.
  19--46, 2017.

\bibitem{ReviewSurgical}
D.~Bouget, M.~Allan, D.~Stoyanov, and P.~Jannin, ``{Vision-based and
  marker-less surgical tool detection and tracking: A review of the
  literature},'' \emph{Medical Image Analysis}, vol.~35, pp. 633--654, 2017.

\bibitem{Ulman2017}
V.~Ulman, M.~Ma{\v{s}}ka, and et~al., ``{An objective comparison of
  cell-tracking algorithms},'' \emph{Nature Methods}, vol.~14, no.~12, pp.
  1141--1152, 2017.

\bibitem{ReviewUnderwater}
J.~Luo, Y.~Han, and L.~Fan, ``{Underwater acoustic target tracking: A
  review},'' \emph{Sensors}, vol.~18, no.~1, p. 112, 2018.

\bibitem{KCF}
J.~F. Henriques, R.~Caseiro, P.~Martins, and J.~Batista, ``{High-speed tracking
  with kernelized correlation filters},'' \emph{IEEE Trans. Pattern Anal. Mach.
  Intell.}, vol.~37, no.~3, pp. 583--596, 2015.

\bibitem{ITS-KCF}
G.~{Ding}, W.~{Chen}, S.~{Zhao}, J.~{Han}, and Q.~{Liu}, ``Real-time scalable
  visual tracking via quadrangle kernelized correlation filters,'' \emph{IEEE
  Trans. Intell. Transp. Sys.}, vol.~19, no.~1, pp. 140--150, 2018.

\bibitem{RADSST}
S.~M. Marvasti-Zadeh, H.~Ghanei-Yakhdan, and S.~Kasaei, ``{Rotation-aware
  discriminative scale space tracking},'' in \emph{Iranian Conf. Electrical
  Engineering (ICEE)}, 2019, pp. 1272--1276.

\bibitem{DCF_Mar1}
S.~M. Marvasti-Zadeh, H.~Ghanei~Yakhdan, and S.~Kasaei, ``Adaptive exploitation
  of pre-trained deep convolutional neural networks for robust visual
  tracking,'' \emph{Multimedia Tools and Applications}, 2021.

\bibitem{DCF_Mar2}
\BIBentryALTinterwordspacing
S.~M. Marvasti~Zadeh, H.~Ghanei-Yakhdan, and S.~Kasaei, ``Beyond
  background-aware correlation filters: Adaptive context modeling by
  hand-crafted and deep rgb features for visual tracking,'' 2020. [Online].
  Available: \url{http://arxiv.org/abs/2004.02932}
\BIBentrySTDinterwordspacing

\bibitem{DCF_Mar3}
S.~M. Marvasti-Zadeh, H.~Ghanei~Yakhdan, and S.~Kasaei, ``Efficient scale
  estimation methods using lightweight deep convolutional neural networks for
  visual tracking,'' \emph{Neural Computing and Applications}, 2021.

\bibitem{DCF_Mar4}
\BIBentryALTinterwordspacing
S.~M. Marvasti-Zadeh, H.~Ghanei-Yakhdan, S.~Kasaei, K.~Nasrollahi, and T.~B.
  Moeslund, ``Effective fusion of deep multitasking representations for robust
  visual tracking,'' 2020. [Online]. Available:
  \url{http://arxiv.org/abs/2004.01382}
\BIBentrySTDinterwordspacing

\bibitem{Xiao2016}
C.~Xiao and A.~Yilmaz, ``{Efficient tracking with distinctive target colors and
  silhouette},'' in \emph{Proc. ICPR}, 2016, pp. 2728--2733.

\bibitem{Bruni2014}
V.~Bruni and D.~Vitulano, ``{An improvement of kernel-based object tracking
  based on human perception},'' \emph{IEEE Trans. Syst., Man, Cybern. Syst.},
  vol.~44, no.~11, pp. 1474--1485, 2014.

\bibitem{Lychkov2018}
I.~I. Lychkov, A.~N. Alfimtsev, and S.~A. Sakulin, ``{Tracking of moving
  objects with regeneration of object feature points},'' in \emph{Proc.
  GloSIC}, 2018, pp. 1--6.

\bibitem{HOG}
N.~Dalal and B.~Triggs, ``{Histograms of oriented gradients for human
  detection},'' in \emph{Proc. IEEE CVPR}, 2005, pp. 886--893.

\bibitem{CN}
J.~{Van De Weijer}, C.~Schmid, and J.~Verbeek, ``{Learning color names from
  real-world images},'' in \emph{Proc. IEEE CVPR}, 2007, pp. 1--8.

\bibitem{SRDCF}
M.~Danelljan, G.~Hager, F.~S. Khan, and M.~Felsberg, ``{Learning spatially
  regularized correlation filters for visual tracking},'' in \emph{Proc. IEEE
  ICCV}, 2015, pp. 4310--4318.

\bibitem{SRDCFdecon}
M.~{Danelljan}, G.~{Häger}, F.~S. {Khan}, and M.~{Felsberg}, ``Adaptive
  decontamination of the training set: A unified formulation for discriminative
  visual tracking,'' in \emph{Proc. IEEE CVPR}, 2016, pp. 1430--1438.

\bibitem{BACF}
H.~K. Galoogahi, A.~Fagg, and S.~Lucey, ``{Learning background-aware
  correlation filters for visual tracking},'' in \emph{Proc. IEEE ICCV}, 2017,
  pp. 1144--1152.

\bibitem{UAV-AutoTrack}
Y.~Li, C.~Fu, F.~Ding, Z.~Huang, and G.~Lu, ``{AutoTrack: Towards
  high-performance visual tracking for UAV with automatic spatio-temporal
  regularization},'' in \emph{Proc. IEEE CVPR}, 2020.

\bibitem{UAV-Aberrance}
Z.~Huang, C.~Fu, Y.~Li, F.~Lin, and P.~Lu, ``{Learning aberrance repressed
  correlation filters for real-time UAV tracking},'' in \emph{Proc. IEEE ICCV},
  2019, pp. 2891--2900.

\bibitem{UAV-Distillation}
F.~Li, C.~Fu, F.~Lin, Y.~Li, and P.~Lu, ``{Training-set distillation for
  real-time UAV object tracking},'' in \emph{Proc. ICRA}, 2020, pp. 1--7.

\bibitem{AlexNet}
A.~Krizhevsky, I.~Sutskever, and G.~E. Hinton, ``{ImageNet classification with
  deep convolutional neural networks},'' in \emph{Proc. NIPS}, vol.~2, 2012,
  pp. 1097--1105.

\bibitem{VGGM}
K.~Chatfield, K.~Simonyan, A.~Vedaldi, and A.~Zisserman, ``{Return of the devil
  in the details: Delving deep into convolutional nets},'' in \emph{Proc.
  BMVC}, 2014, pp. 1--11.

\bibitem{VGGNet}
K.~Simonyan and A.~Zisserman, ``{Very deep convolutional networks for
  large-scale image recognition},'' in \emph{Proc. ICLR}, 2014, pp. 1--14.

\bibitem{GoogLeNet}
C.~Szegedy, W.~Liu, Y.~Jia, P.~Sermanet, S.~Reed, D.~Anguelov, D.~Erhan,
  V.~Vanhoucke, and A.~Rabinovich, ``{Going deeper with convolutions},'' in
  \emph{Proc. IEEE CVPR}, 2015, pp. 1--9.

\bibitem{ResNet}
K.~He, X.~Zhang, S.~Ren, and J.~Sun, ``{Deep residual learning for image
  recognition},'' in \emph{Proc. IEEE CVPR}, 2016, pp. 770--778.

\bibitem{ImageNet}
O.~Russakovsky, J.~Deng, H.~Su, J.~Krause, S.~Satheesh, S.~Ma, Z.~Huang,
  A.~Karpathy, A.~Khosla, M.~Bernstein, A.~C. Berg, and L.~Fei-Fei, ``{ImageNet
  large scale visual recognition challenge},'' \emph{IJCV}, vol. 115, no.~3,
  pp. 211--252, 2015.

\bibitem{VOT-2013}
M.~Kristan, R.~Pflugfelder, A.~Leonardis, J.~Matas, F.~Porikli, and et~al.,
  ``{The visual object tracking {VOT}2013 challenge results},'' in \emph{Proc.
  ICCV}, 2013, pp. 98--111.

\bibitem{VOT-2014}
M.~Kristan, R.~Pflugfelder, A.~Leonardis, J.~Matas, and et~al., ``{The visual
  object tracking {VOT}2014 challenge results},'' in \emph{Proc. ECCV}, 2015,
  pp. 191--217.

\bibitem{VOT-2015}
M.~Kristan, J.~Matas, A.~Leonardis, M.~Felsberg, and et~al., ``{The visual
  object tracking {VOT}2015 challenge results},'' in \emph{Proc. IEEE ICCV},
  2015, pp. 564--586.

\bibitem{VOT-2016}
M.~Kristan, J.~Matas, A.~Leonardis, M.~Felsberg, R.~Pflugfelder, and et~al.,
  ``{The visual object tracking {VOT}2016 challenge results},'' in \emph{Proc.
  ECCVW}, 2016, pp. 777--823.

\bibitem{VOT-2017}
M.~{Kristan}, A.~{Leonardis}, J.~{Matas}, M.~{Felsberg}, R.~{Pflugfelder},
  L.~C. {Zajc}, and et~al., ``The visual object tracking {VOT}2017 challenge
  results,'' in \emph{Proc. IEEE ICCVW}, 2017, pp. 1949--1972.

\bibitem{VOT-2018}
M.~Kristan, A.~Leonardis, J.~Matas, M.~Felsberg, R.~Pflugfelder, and et~al.,
  ``The sixth visual object tracking {VOT}2018 challenge results,'' in
  \emph{Proc. ECCVW}, 2019, pp. 3--53.

\bibitem{VOT-2019}
M.~Kristan and et~al., ``The seventh visual object tracking {VOT2019} challenge
  results,'' in \emph{Proc. ICCVW}, 2019.

\bibitem{SurveyYilmaz}
A.~Yilmaz, O.~Javed, and M.~Shah, ``{Object tracking: A survey},'' \emph{ACM
  Computing Surveys}, vol.~38, no.~4, Dec. 2006.

\bibitem{SurveySmeulders}
A.~W. Smeulders, D.~M. Chu, R.~Cucchiara, S.~Calderara, A.~Dehghan, and
  M.~Shah, ``{Visual tracking: An experimental survey},'' \emph{IEEE Trans.
  Pattern Anal. Mach. Intell.}, vol.~36, no.~7, pp. 1442--1468, 2014.

\bibitem{SurveyYang}
H.~Yang, L.~Shao, F.~Zheng, L.~Wang, and Z.~Song, ``{Recent advances and trends
  in visual tracking: A review},'' \emph{Neurocomputing}, vol.~74, no.~18, pp.
  3823--3831, 2011.

\bibitem{SurveyLi-AppearanceModels}
X.~Li, W.~Hu, C.~Shen, Z.~Zhang, A.~Dick, and A.~{Van Den Hengel}, ``{A survey
  of appearance models in visual object tracking},'' \emph{ACM Trans. Intell.
  Syst. Tec.}, vol.~4, no.~4, pp. 58:1----58:48, 2013.

\bibitem{UAV-Boundary}
C.~Fu, Z.~Huang, Y.~Li, R.~Duan, and P.~Lu, ``{Boundary effect-aware visual
  tracking for UAV with online enhanced background learning and multi-frame
  consensus verification},'' in \emph{Proc. IROS}, 2019, pp. 4415--4422.

\bibitem{UAV-KeyFilter}
Y.~Li, C.~Fu, Z.~Huang, Y.~Zhang, and J.~Pan, ``{Keyfilter-aware real-time uav
  object tracking},'' in \emph{Proc. ICRA}, 2020.

\bibitem{TrackingNoisyTargets}
\BIBentryALTinterwordspacing
M.~Fiaz, A.~Mahmood, and S.~K. Jung, ``Tracking noisy targets: A review of
  recent object tracking approaches,'' 2018. [Online]. Available:
  \url{http://arxiv.org/abs/1802.03098}
\BIBentrySTDinterwordspacing

\bibitem{HandcraftedDeepTrackers}
M.~Fiaz, A.~Mahmood, S.~Javed, and S.~K. Jung, ``{Handcrafted and deep
  trackers: Recent visual object tracking approaches and trends},'' \emph{ACM
  Computing Surveys}, vol.~52, no.~2, pp. 43:1----43:44, 2019.

\bibitem{SurveyDeepTracking}
P.~Li, D.~Wang, L.~Wang, and H.~Lu, ``{Deep visual tracking: Review and
  experimental comparison},'' \emph{Pattern Recognit.}, vol.~76, pp. 323--338,
  2018.

\bibitem{SurveySiamese}
\BIBentryALTinterwordspacing
R.~Pflugfelder, ``An in-depth analysis of visual tracking with {Siamese} neural
  networks,'' 2017. [Online]. Available: \url{http://arxiv.org/abs/1707.00569}
\BIBentrySTDinterwordspacing

\bibitem{HCFT}
C.~Ma, J.~B. Huang, X.~Yang, and M.~H. Yang, ``{Hierarchical convolutional
  features for visual tracking},'' in \emph{Proc. IEEE ICCV}, 2015, pp.
  3074--3082.

\bibitem{DeepSRDCF}
M.~Danelljan, G.~Hager, F.~S. Khan, and M.~Felsberg, ``{Convolutional features
  for correlation filter based visual tracking},'' in \emph{Proc. IEEE ICCVW},
  2016, pp. 621--629.

\bibitem{FCNT}
L.~Wang, W.~Ouyang, X.~Wang, and H.~Lu, ``{Visual tracking with fully
  convolutional networks},'' in \emph{Proc. IEEE ICCV}, 2015, pp. 3119--3127.

\bibitem{CNN-SVM}
S.~Hong, T.~You, S.~Kwak, and B.~Han, ``{Online tracking by learning
  discriminative saliency map with convolutional neural network},'' in
  \emph{Proc. ICML}, 2015, pp. 597--606.

\bibitem{DPST}
Y.~Zha, T.~Ku, Y.~Li, and P.~Zhang, ``{Deep position-sensitive tracking},''
  \emph{IEEE Trans. Multimedia}, no.~8, 2019.

\bibitem{CCOT}
M.~Danelljan, A.~Robinson, F.~S. Khan, and M.~Felsberg, ``{Beyond correlation
  filters: Learning continuous convolution operators for visual tracking},'' in
  \emph{Proc. ECCV}, vol. 9909 LNCS, 2016, pp. 472--488.

\bibitem{GOTURN}
D.~Held, S.~Thrun, and S.~Savarese, ``{Learning to track at 100 FPS with deep
  regression networks},'' in \emph{Proc. ECCV}, 2016, pp. 749--765.

\bibitem{SiamFC}
L.~Bertinetto, J.~Valmadre, J.~F. Henriques, A.~Vedaldi, and P.~H. Torr,
  ``{Fully-convolutional {Siamese} networks for object tracking},'' in
  \emph{Proc. ECCV}, 2016, pp. 850--865.

\bibitem{SINT}
R.~Tao, E.~Gavves, and A.~W. Smeulders, ``{Siamese instance search for
  tracking},'' in \emph{Proc. IEEE CVPR}, 2016, pp. 1420--1429.

\bibitem{MDNet}
H.~Nam and B.~Han, ``{Learning multi-domain convolutional neural networks for
  visual tracking},'' in \emph{Proc. IEEE CVPR}, 2016, pp. 4293--4302.

\bibitem{HDT}
Y.~Qi, S.~Zhang, L.~Qin, H.~Yao, Q.~Huang, J.~Lim, and M.~H. Yang, ``{Hedged
  deep tracking},'' in \emph{Proc. IEEE CVPR}, 2016, pp. 4303--4311.

\bibitem{STCT}
L.~Wang, W.~Ouyang, X.~Wang, and H.~Lu, ``{STCT: Sequentially training
  convolutional networks for visual tracking},'' in \emph{Proc. IEEE CVPR},
  2016, pp. 1373--1381.

\bibitem{RPNT}
G.~Zhu, F.~Porikli, and H.~Li, ``{Robust visual tracking with deep
  convolutional neural network based object proposals on PETS},'' in
  \emph{Proc. IEEE CVPRW}, 2016, pp. 1265--1272.

\bibitem{DeepTrack}
H.~Li and et~al., ``{DeepTrack: Learning discriminative feature representations
  online for robust visual tracking},'' \emph{IEEE Trans. Image Process.},
  vol.~25, no.~4, pp. 1834--1848, 2016.

\bibitem{DeepTrack_BMVC}
H.~Li, Y.~Li, and F.~Porikli, ``{DeepTrack}: Learning discriminative feature
  representations by convolutional neural networks for visual tracking,'' in
  \emph{Proc. BMVC}, 2014.

\bibitem{CNT}
K.~Zhang, Q.~Liu, Y.~Wu, and M.~H. Yang, ``{Robust visual tracking via
  convolutional networks without training},'' \emph{IEEE Trans. Image
  Process.}, vol.~25, no.~4, pp. 1779--1792, 2016.

\bibitem{CF-CNN}
C.~Ma, Y.~Xu, B.~Ni, and X.~Yang, ``{When correlation filters meet
  convolutional neural networks for visual tracking},'' \emph{IEEE Signal
  Process. Lett.}, vol.~23, no.~10, pp. 1454--1458, 2016.

\bibitem{TCNN}
\BIBentryALTinterwordspacing
H.~Nam, M.~Baek, and B.~Han, ``Modeling and propagating {CNNs} in a tree
  structure for visual tracking,'' 2016. [Online]. Available:
  \url{http://arxiv.org/abs/1608.07242}
\BIBentrySTDinterwordspacing

\bibitem{RDLT}
G.~Wu, W.~Lu, G.~Gao, C.~Zhao, and J.~Liu, ``{Regional deep learning model for
  visual tracking},'' \emph{Neurocomputing}, vol. 175, no. PartA, pp. 310--323,
  2015.

\bibitem{PTAV-ICCV}
H.~Fan and H.~Ling, ``{Parallel tracking and verifying: A framework for
  real-time and high accuracy visual tracking},'' in \emph{Proc. IEEE ICCV},
  2017, pp. 5487--5495.

\bibitem{PTAV}
H.~Fan and H.Ling, ``{Parallel tracking and verifying},'' \emph{IEEE Trans.
  Image Process.}, vol.~28, no.~8, pp. 4130--4144, 2019.

\bibitem{CREST}
Y.~Song, C.~Ma, L.~Gong, J.~Zhang, R.~W. Lau, and M.~H. Yang, ``{CREST:
  Convolutional residual learning for visual tracking},'' in \emph{Proc. ICCV},
  2017, pp. 2574--2583.

\bibitem{UCT}
Z.~Zhu, G.~Huang, W.~Zou, D.~Du, and C.~Huang, ``{UCT: Learning unified
  convolutional networks for real-time visual tracking},'' in \emph{Proc.
  ICCVW}, 2018, pp. 1973--1982.

\bibitem{DSiam}
Q.~Guo, W.~Feng, C.~Zhou, R.~Huang, L.~Wan, and S.~Wang, ``{Learning dynamic
  {Siamese} network for visual object tracking},'' in \emph{Proc. IEEE ICCV},
  2017, pp. 1781--1789.

\bibitem{TSN}
Z.~Teng, J.~Xing, Q.~Wang, C.~Lang, S.~Feng, and Y.~Jin, ``{Robust object
  tracking based on temporal and spatial deep networks},'' in \emph{Proc. IEEE
  ICCV}, 2017, pp. 1153--1162.

\bibitem{WECO}
Z.~He, Y.~Fan, J.~Zhuang, Y.~Dong, and H.~Bai, ``{Correlation filters with
  weighted convolution responses},'' in \emph{Proc. ICCVW}, 2018, pp.
  1992--2000.

\bibitem{RFL}
T.~Yang and A.~B. Chan, ``{Recurrent filter learning for visual tracking},'' in
  \emph{Proc. ICCVW}, 2018, pp. 2010--2019.

\bibitem{IBCCF}
F.~Li, Y.~Yao, P.~Li, D.~Zhang, W.~Zuo, and M.~H. Yang, ``{Integrating boundary
  and center correlation filters for visual tracking with aspect ratio
  variation},'' in \emph{Proc. IEEE ICCVW}, 2018, pp. 2001--2009.

\bibitem{DTO}
X.~Wang, H.~Li, Y.~Li, F.~Porikli, and M.~Wang, ``{Deep tracking with
  objectness},'' in \emph{Proc. ICIP}, 2018, pp. 660--664.

\bibitem{SRT}
X.~Xu, B.~Ma, H.~Chang, and X.~Chen, ``{Siamese recurrent architecture for
  visual tracking},'' in \emph{Proc. ICIP}, 2018, pp. 1152--1156.

\bibitem{R-FCSN}
L.~{Yang}, P.~{Jiang}, F.~{Wang}, and X.~{Wang}, ``Region-based fully
  convolutional {Siamese} networks for robust real-time visual tracking,'' in
  \emph{Proc. ICIP}, 2017, pp. 2567--2571.

\bibitem{GNet}
T.~{Kokul}, C.~{Fookes}, S.~{Sridharan}, A.~{Ramanan}, and U.~A.~J.
  {Pinidiyaarachchi}, ``Gate connected convolutional neural network for object
  tracking,'' in \emph{Proc. ICIP}, 2017, pp. 2602--2606.

\bibitem{LST}
K.~{Dai}, Y.~{Wang}, and X.~{Yan}, ``Long-term object tracking based on
  {Siamese} network,'' in \emph{Proc. ICIP}, 2017, pp. 3640--3644.

\bibitem{VRCPF}
B.~{Akok}, F.~{Gurkan}, O.~{Kaplan}, and B.~{Gunsel}, ``Robust object tracking
  by interleaving variable rate color particle filtering and deep learning,''
  in \emph{Proc. ICIP}, 2017, pp. 3665--3669.

\bibitem{DCPF}
R.~J. {Mozhdehi} and H.~{Medeiros}, ``Deep convolutional particle filter for
  visual tracking,'' in \emph{Proc. IEEE ICIP}, 2017, pp. 3650--3654.

\bibitem{CFNet}
J.~Valmadre, L.~Bertinetto, J.~Henriques, A.~Vedaldi, and P.~H. Torr,
  ``{End-to-end representation learning for correlation filter based
  tracking},'' in \emph{Proc. IEEE CVPR}, 2017, pp. 5000--5008.

\bibitem{ECO}
M.~Danelljan, G.~Bhat, F.~{Shahbaz Khan}, and M.~Felsberg, ``{ECO: Efficient
  convolution operators for tracking},'' in \emph{Proc. IEEE CVPR}, 2017, pp.
  6931--6939.

\bibitem{DeepCSRDCF}
A.~Luke{\v{z}}i{\v{c}}, T.~Voj{\'{i}}ř, L.~{{\v{C}}ehovin Zajc}, J.~Matas,
  and M.~Kristan, ``{Discriminative correlation filter tracker with channel and
  spatial reliability},'' \emph{IJCV}, vol. 126, no.~7, pp. 671--688, 2018.

\bibitem{MCPF}
T.~Zhang, C.~Xu, and M.~H. Yang, ``{Multi-task correlation particle filter for
  robust object tracking},'' in \emph{Proc. IEEE CVPR}, 2017, pp. 4819--4827.

\bibitem{BranchOut}
B.~Han, J.~Sim, and H.~Adam, ``{BranchOut: Regularization for online ensemble
  tracking with convolutional neural networks},'' in \emph{Proc. IEEE CVPR},
  2017, pp. 521--530.

\bibitem{DeepLMCF}
M.~Wang, Y.~Liu, and Z.~Huang, ``{Large margin object tracking with circulant
  feature maps},'' in \emph{Proc. IEEE CVPR}, 2017, pp. 4800--4808.

\bibitem{Obli-RaFT}
L.~{Zhang}, J.~{Varadarajan}, P.~N. {Suganthan}, N.~{Ahuja}, and P.~{Moulin},
  ``{Robust visual tracking using oblique random forests},'' in \emph{Proc.
  IEEE CVPR}, 2017, pp. 5825--5834.

\bibitem{ACFN}
J.~Choi, H.~J. Chang, S.~Yun, T.~Fischer, Y.~Demiris, and J.~Y. Choi,
  ``{Attentional correlation filter network for adaptive visual tracking},'' in
  \emph{Proc. IEEE CVPR}, 2017, pp. 4828--4837.

\bibitem{SANet}
H.~Fan and H.~Ling, ``{SANet: Structure-aware network for visual tracking},''
  in \emph{Proc. IEEE CVPRW}, 2017, pp. 2217--2224.

\bibitem{DCFNet}
\BIBentryALTinterwordspacing
Q.~Wang, J.~Gao, J.~Xing, M.~Zhang, and W.~Hu, ``{DCFNet}: Discriminant
  correlation filters network for visual tracking,'' 2017. [Online]. Available:
  \url{http://arxiv.org/abs/1704.04057}
\BIBentrySTDinterwordspacing

\bibitem{DET}
J.~Guo and T.~Xu, ``{Deep ensemble tracking},'' \emph{IEEE Signal Process.
  Lett.}, vol.~24, no.~10, pp. 1562--1566, 2017.

\bibitem{DRN}
J.~Gao, T.~Zhang, X.~Yang, and C.~Xu, ``{Deep relative tracking},'' \emph{IEEE
  Trans. Image Process.}, vol.~26, no.~4, pp. 1845--1858, 2017.

\bibitem{DNT}
Z.~Chi, H.~Li, H.~Lu, and M.~H. Yang, ``{Dual deep network for visual
  tracking},'' \emph{IEEE Trans. Image Process.}, vol.~26, no.~4, pp.
  2005--2015, 2017.

\bibitem{STSGS}
P.~Zhang, T.~Zhuo, W.~Huang, K.~Chen, and M.~Kankanhalli, ``{Online object
  tracking based on CNN with spatial-temporal saliency guided sampling},''
  \emph{Neurocomputing}, vol. 257, pp. 115--127, 2017.

\bibitem{Tripletloss}
X.~Dong and J.~Shen, ``{Triplet loss in {Siamese} network for object
  tracking},'' in \emph{Proc. ECCV}, vol. 11217 LNCS, 2018, pp. 472--488.

\bibitem{DSLT}
X.~Lu, C.~Ma, B.~Ni, X.~Yang, I.~Reid, and M.~H. Yang, ``{Deep regression
  tracking with shrinkage loss},'' in \emph{Proc. ECCV}, 2018, pp. 369--386.

\bibitem{UPDT}
G.~Bhat, J.~Johnander, M.~Danelljan, F.~S. Khan, and M.~Felsberg, ``{Unveiling
  the power of deep tracking},'' in \emph{Proc. ECCV}, 2018, pp. 493--509.

\bibitem{ACT}
B.~Chen, D.~Wang, P.~Li, S.~Wang, and H.~Lu, ``Real-time `actor-critic'
  tracking,'' in \emph{Proc. ECCV}, 2018, pp. 328--345.

\bibitem{DaSiamRPN}
Z.~Zhu, Q.~Wang, B.~Li, W.~Wu, J.~Yan, and W.~Hu, ``{Distractor-aware {Siamese}
  networks for visual object tracking},'' in \emph{Proc. ECCV}, vol. 11213
  LNCS, 2018, pp. 103--119.

\bibitem{RT-MDNet}
I.~Jung, J.~Son, M.~Baek, and B.~Han, ``Real-time {MDNet},'' in \emph{Proc.
  ECCV}, 2018, pp. 89--104.

\bibitem{StructSiam}
Y.~Zhang, L.~Wang, J.~Qi, D.~Wang, M.~Feng, and H.~Lu, ``{Structured {Siamese}
  network for real-time visual tracking},'' in \emph{Proc. ECCV}, 2018, pp.
  355--370.

\bibitem{MMLT}
H.~Lee, S.~Choi, and C.~Kim, ``{A memory model based on the {Siamese} network
  for long-term tracking},'' in \emph{Proc. ECCVW}, 2019, pp. 100--115.

\bibitem{CPT}
M.~Che, R.~Wang, Y.~Lu, Y.~Li, H.~Zhi, and C.~Xiong, ``{Channel pruning for
  visual tracking},'' in \emph{Proc. ECCVW}, 2019, pp. 70--82.

\bibitem{STP}
E.~Burceanu and M.~Leordeanu, ``{Learning a robust society of tracking parts
  using co-occurrence constraints},'' in \emph{Proc. ECCVW}, 2019, pp.
  162--178.

\bibitem{Siam-MCF}
H.~Morimitsu, ``{Multiple context features in {Siamese} networks for visual
  object tracking},'' in \emph{Proc. ECCVW}, 2019, pp. 116--131.

\bibitem{Siam-BM}
A.~He, C.~Luo, X.~Tian, and W.~Zeng, ``{Towards a better match in {Siamese}
  network based visual object tracker},'' in \emph{Proc. ECCVW}, 2019, pp.
  132--147.

\bibitem{WAEF}
L.~Rout, D.~Mishra, and R.~K. S.~S. Gorthi, ``{WAEF: Weighted aggregation with
  enhancement filter for visual object tracking},'' in \emph{Proc. ECCVW},
  2019, pp. 83--99.

\bibitem{TRACA}
J.~Choi, H.~J. Chang, T.~Fischer, S.~Yun, K.~Lee, J.~Jeong, Y.~Demiris, and
  J.~Y. Choi, ``{Context-aware deep feature compression for high-speed visual
  tracking},'' in \emph{Proc. IEEE CVPR}, 2018, pp. 479--488.

\bibitem{VITAL}
Y.~Song, C.~Ma, X.~Wu, L.~Gong, L.~Bao, W.~Zuo, C.~Shen, R.~W. Lau, and M.~H.
  Yang, ``{VITAL: Visual tracking via adversarial learning},'' in \emph{Proc.
  IEEE CVPR}, 2018, pp. 8990--8999.

\bibitem{STRCF}
F.~Li, C.~Tian, W.~Zuo, L.~Zhang, and M.~H. Yang, ``{Learning spatial-temporal
  regularized correlation filters for visual tracking},'' in \emph{Proc. IEEE
  CVPR}, 2018, pp. 4904--4913.

\bibitem{SiamRPN}
B.~Li, J.~Yan, W.~Wu, Z.~Zhu, and X.~Hu, ``High performance visual tracking
  with {Siamese} region proposal network,'' in \emph{Proc. IEEE CVPR}, 2018,
  pp. 8971--8980.

\bibitem{SA-Siam}
A.~He, C.~Luo, X.~Tian, and W.~Zeng, ``{A twofold {Siamese} network for
  real-time object tracking},'' in \emph{Proc. IEEE CVPR}, 2018, pp.
  4834--4843.

\bibitem{FlowTrack}
Z.~Zhu, W.~Wu, W.~Zou, and J.~Yan, ``{End-to-end flow correlation tracking with
  spatial-temporal attention},'' in \emph{Proc. IEEE CVPR}, 2018, pp. 548--557.

\bibitem{DRT}
C.~Sun, D.~Wang, H.~Lu, and M.~H. Yang, ``{Correlation tracking via joint
  discrimination and reliability learning},'' in \emph{Proc. IEEE CVPR}, 2018,
  pp. 489--497.

\bibitem{LSART}
C.~Sun, D.~Wang, H.~Lu, and M.~Yang, ``{Learning spatial-aware regressions for
  visual tracking},'' in \emph{Proc. IEEE CVPR}, 2018, pp. 8962--8970.

\bibitem{RASNet}
Q.~Wang, Z.~Teng, J.~Xing, J.~Gao, W.~Hu, and S.~Maybank, ``{Learning
  attentions: Residual attentional {Siamese} network for high performance
  online visual tracking},'' in \emph{Proc. IEEE CVPR}, 2018, pp. 4854--4863.

\bibitem{MCCT}
N.~Wang, W.~Zhou, Q.~Tian, R.~Hong, M.~Wang, and H.~Li, ``{Multi-cue
  correlation filters for robust visual tracking},'' in \emph{Proc. IEEE CVPR},
  2018, pp. 4844--4853.

\bibitem{DCPF2}
R.~J. Mozhdehi, Y.~Reznichenko, A.~Siddique, and H.~Medeiros, ``{Deep
  convolutional particle filter with adaptive correlation maps for visual
  tracking},'' in \emph{Proc. ICIP}, 2018, pp. 798--802.

\bibitem{VDSR-SRT}
Z.~Lin and C.~Yuan, ``{Robust visual tracking in low-resolution sequence},'' in
  \emph{Proc. ICIP}, 2018, pp. 4103--4107.

\bibitem{FCSFN}
M.~Cen and C.~Jung, ``{Fully convolutional {Siamese} fusion networks for object
  tracking},'' in \emph{Proc. ICIP}, 2018, pp. 3718--3722.

\bibitem{FRPN2T-siam}
G.~Wang, B.~Liu, W.~Li, and N.~Yu, ``{Flow guided {Siamese} network for visual
  tracking},'' in \emph{Proc. ICIP}, 2018, pp. 231--235.

\bibitem{FMFT}
K.~Dai, Y.~Wang, X.~Yan, and Y.~Huo, ``{Fusion of template matching and
  foreground detection for robust visual tracking},'' in \emph{Proc. ICIP},
  2018, pp. 2720--2724.

\bibitem{IMLCF}
G.~Liu and G.~Liu, ``{Integrating multi-level convolutional features for
  correlation filter tracking},'' in \emph{Proc. ICIP}, 2018, pp. 3029--3033.

\bibitem{TGGAN}
J.~Guo, T.~Xu, S.~Jiang, and Z.~Shen, ``{Generating reliable online adaptive
  templates for visual tracking},'' in \emph{Proc. ICIP}, 2018, pp. 226--230.

\bibitem{DAT}
S.~Pu, Y.~Song, C.~Ma, H.~Zhang, and M.~H. Yang, ``{Deep attentive tracking via
  reciprocative learning},'' in \emph{Proc. NIPS}, 2018, pp. 1931--1941.

\bibitem{DCTN}
X.~Jiang, X.~Zhen, B.~Zhang, J.~Yang, and X.~Cao, ``{Deep collaborative
  tracking networks},'' in \emph{Proc. BMVC}, 2018, p.~87.

\bibitem{FPRNet}
D.~Ma, W.~Bu, and X.~Wu, ``{Multi-scale recurrent tracking via pyramid
  recurrent network and optical flow},'' in \emph{Proc. BMVC}, 2018, p. 242.

\bibitem{HCFTs}
C.~Ma, J.~B. Huang, X.~Yang, and M.~H. Yang, ``{Robust visual tracking via
  hierarchical convolutional features},'' \emph{IEEE Trans. Pattern Anal. Mach.
  Intell.}, 2018.

\bibitem{adaDDCF}
Z.~Han, P.~Wang, and Q.~Ye, ``{Adaptive discriminative deep correlation filter
  for visual object tracking},'' \emph{IEEE Trans. Circuits Syst. Video
  Technol.}, 2018.

\bibitem{YCNN}
K.~Chen and W.~Tao, ``{Once for all: A two-flow convolutional neural network
  for visual tracking},'' \emph{IEEE Trans. Circuits Syst. Video Technol.},
  vol.~28, no.~12, pp. 3377--3386, 2018.

\bibitem{DeepHPFT}
S.~Li, S.~Zhao, B.~Cheng, E.~Zhao, and J.~Chen, ``{Robust visual tracking via
  hierarchical particle filter and ensemble deep features},'' \emph{IEEE Trans.
  Circuits Syst. Video Technol.}, 2018.

\bibitem{CFCF}
E.~Gundogdu and A.~A. Alatan, ``{Good features to correlate for visual
  tracking},'' \emph{IEEE Trans. Image Process.}, vol.~27, no.~5, pp.
  2526--2540, 2018.

\bibitem{CFSRL}
Y.~Xie, J.~Xiao, K.~Huang, J.~Thiyagalingam, and Y.~Zhao, ``{Correlation filter
  selection for visual tracking using reinforcement learning},'' \emph{IEEE
  Trans. Circuits Syst. Video Technol.}, 2018.

\bibitem{P2T}
J.~Gao, T.~Zhang, X.~Yang, and C.~Xu, ``{P2T: Part-to-target tracking via deep
  regression learning},'' \emph{IEEE Trans. Image Process.}, vol.~27, no.~6,
  pp. 3074--3086, 2018.

\bibitem{DCDCF}
C.~Peng, F.~Liu, J.~Yang, and N.~Kasabov, ``{Densely connected discriminative
  correlation filters for visual tracking},'' \emph{IEEE Signal Process.
  Lett.}, vol.~25, no.~7, pp. 1019--1023, 2018.

\bibitem{FICFNet}
D.~Li, G.~Wen, Y.~Kuai, and F.~Porikli, ``{End-to-end feature integration for
  correlation filter tracking with channel attention},'' \emph{IEEE Signal
  Process. Lett.}, vol.~25, no.~12, pp. 1815--1819, 2018.

\bibitem{LCTdeep}
C.~Ma, J.~B. Huang, and et~al., ``{Adaptive correlation filters with long-term
  and short-term memory for object tracking},'' \emph{IJCV}, vol. 126, no.~8,
  pp. 771--796, 2018.

\bibitem{HSTC}
Y.~Cao, H.~Ji, W.~Zhang, and F.~Xue, ``{Learning spatio-temporal context via
  hierarchical features for visual tracking},'' \emph{Signal Proc.: Image
  Comm.}, vol.~66, pp. 50--65, 2018.

\bibitem{DeepFWDCF}
F.~Du, P.~Liu, W.~Zhao, and X.~Tang, ``{Spatial–temporal adaptive feature
  weighted correlation filter for visual tracking},'' \emph{Signal Proc.: Image
  Comm.}, vol.~67, pp. 58--70, 2018.

\bibitem{CF-FCSiam}
Y.~Kuai, G.~Wen, and D.~Li, ``{When correlation filters meet
  fully-convolutional {Siamese} networks for distractor-aware tracking},''
  \emph{Signal Proc.: Image Comm.}, vol.~64, pp. 107--117, 2018.

\bibitem{MGNet}
W.~Gan, M.~S. Lee, C.~hao Wu, and C.~C. Kuo, ``{Online object tracking via
  motion-guided convolutional neural network (MGNet)},'' \emph{J. VIS. COMMUN.
  IMAGE R.}, vol.~53, pp. 180--191, 2018.

\bibitem{ORHF}
M.~Liu, C.~B. Jin, B.~Yang, X.~Cui, and H.~Kim, ``{Occlusion-robust object
  tracking based on the confidence of online selected hierarchical features},''
  \emph{IET Image Proc.}, vol.~12, no.~11, pp. 2023--2029, 2018.

\bibitem{ASRCF}
K.~Dai, D.~Wang, H.~Lu, C.~Sun, and J.~Li, ``{Visual tracking via adaptive
  spatially-regularized correlation filters},'' in \emph{Proc. CVPR}, 2019, pp.
  4670--4679.

\bibitem{ATOM}
\BIBentryALTinterwordspacing
M.~Danelljan, G.~Bhat, F.~S. Khan, and M.~Felsberg, ``{ATOM}: Accurate tracking
  by overlap maximization,'' 2018. [Online]. Available:
  \url{http://arxiv.org/abs/1811.07628}
\BIBentrySTDinterwordspacing

\bibitem{CRPN}
\BIBentryALTinterwordspacing
H.~Fan and H.~Ling, ``Siamese cascaded region proposal networks for real-time
  visual tracking,'' 2018. [Online]. Available:
  \url{http://arxiv.org/abs/1812.06148}
\BIBentrySTDinterwordspacing

\bibitem{GCT}
J.~Gao, T.~Zhang, and C.~Xu, ``{Graph convolutional tracking},'' in \emph{Proc.
  CVPR}, 2019, pp. 4649--4659.

\bibitem{RPCF}
Y.~Sun, C.~Sun, D.~Wang, Y.~He, and H.~Lu, ``{ROI pooled correlation filters
  for visual tracking},'' in \emph{Proc. CVPR}, 2019, pp. 5783--5791.

\bibitem{SPM-Tracker}
\BIBentryALTinterwordspacing
G.~Wang, C.~Luo, Z.~Xiong, and W.~Zeng, ``Spm-tracker: Series-parallel matching
  for real-time visual object tracking,'' 2019. [Online]. Available:
  \url{http://arxiv.org/abs/1904.04452}
\BIBentrySTDinterwordspacing

\bibitem{SiamDW}
\BIBentryALTinterwordspacing
Z.~Zhang and H.~Peng, ``Deeper and wider {Siamese} networks for real-time
  visual tracking,'' 2019. [Online]. Available:
  \url{http://arxiv.org/abs/1901.01660}
\BIBentrySTDinterwordspacing

\bibitem{SiamMask}
\BIBentryALTinterwordspacing
Q.~Wang, L.~Zhang, L.~Bertinetto, W.~Hu, and P.~H.~S. Torr, ``Fast online
  object tracking and segmentation: A unifying approach,'' 2018. [Online].
  Available: \url{http://arxiv.org/abs/1812.05050}
\BIBentrySTDinterwordspacing

\bibitem{SiamRPN++}
\BIBentryALTinterwordspacing
B.~Li, W.~Wu, Q.~Wang, F.~Zhang, J.~Xing, and J.~Yan, ``{SiamRPN++}: Evolution
  of {Siamese} visual tracking with very deep networks,'' 2018. [Online].
  Available: \url{http://arxiv.org/abs/1812.11703}
\BIBentrySTDinterwordspacing

\bibitem{TADT}
\BIBentryALTinterwordspacing
X.~Li, C.~Ma, B.~Wu, Z.~He, and M.-H. Yang, ``Target-aware deep tracking,''
  2019. [Online]. Available: \url{http://arxiv.org/abs/1904.01772}
\BIBentrySTDinterwordspacing

\bibitem{UDT}
\BIBentryALTinterwordspacing
N.~Wang, Y.~Song, C.~Ma, W.~Zhou, W.~Liu, and H.~Li, ``Unsupervised deep
  tracking,'' 2019. [Online]. Available: \url{http://arxiv.org/abs/1904.01828}
\BIBentrySTDinterwordspacing

\bibitem{DiMP}
\BIBentryALTinterwordspacing
G.~Bhat, M.~Danelljan, L.~V. Gool, and R.~Timofte, ``Learning discriminative
  model prediction for tracking,'' 2019. [Online]. Available:
  \url{http://arxiv.org/abs/1904.07220}
\BIBentrySTDinterwordspacing

\bibitem{ADT}
F.~Zhao, J.~Wang, Y.~Wu, and M.~Tang, ``{Adversarial deep tracking},''
  \emph{IEEE Trans. Circuits Syst. Video Technol.}, vol.~29, no.~7, pp.
  1998--2011, 2019.

\bibitem{CODA}
H.~Li, X.~Wang, F.~Shen, Y.~Li, F.~Porikli, and M.~Wang, ``{Real-time deep
  tracking via corrective domain adaptation},'' \emph{IEEE Trans. Circuits
  Syst. Video Technol.}, vol. 8215, 2019.

\bibitem{DRRL}
B.~Zhong, B.~Bai, J.~Li, Y.~Zhang, and Y.~Fu, ``{Hierarchical tracking by
  reinforcement learning-based searching and coarse-to-fine verifying},''
  \emph{IEEE Trans. Image Process.}, vol.~28, no.~5, pp. 2331--2341, 2019.

\bibitem{SMART}
J.~Gao, T.~Zhang, and C.~Xu, ``{SMART: Joint sampling and regression for visual
  tracking},'' \emph{IEEE Trans. Image Process.}, vol.~28, no.~8, pp.
  3923--3935, 2019.

\bibitem{MRCNN}
H.~Hu, B.~Ma, J.~Shen, H.~Sun, L.~Shao, and F.~Porikli, ``{Robust object
  tracking using manifold regularized convolutional neural networks},''
  \emph{IEEE Trans. Multimedia}, vol.~21, no.~2, pp. 510--521, 2019.

\bibitem{MM}
L.~Wang, L.~Zhang, J.~Wang, and Z.~Yi, ``{Memory mechanisms for discriminative
  visual tracking algorithms with deep neural networks},'' \emph{IEEE Trans.
  Cogn. Devel. Syst.}, 2019.

\bibitem{MTHCF}
Y.~Kuai, G.~Wen, and D.~Li, ``{Multi-task hierarchical feature learning for
  real-time visual tracking},'' \emph{IEEE Sensors J.}, vol.~19, no.~5, pp.
  1961--1968, 2019.

\bibitem{AEPCF}
X.~Cheng, Y.~Zhang, L.~Zhou, and Y.~Zheng, ``{Visual tracking via Auto-Encoder
  pair correlation filter},'' \emph{IEEE Trans. Ind. Electron.}, 2019.

\bibitem{IMM-DFT}
F.~Tang, X.~Lu, X.~Zhang, S.~Hu, and H.~Zhang, ``{Deep feature tracking based
  on interactive multiple model},'' \emph{Neurocomputing}, vol. 333, pp.
  29--40, 2019.

\bibitem{TAAT}
X.~Lu, B.~Ni, C.~Ma, and X.~Yang, ``{Learning transform-aware attentive network
  for object tracking},'' \emph{Neurocomputing}, vol. 349, pp. 133--144, 2019.

\bibitem{DeepTACF}
D.~Li, G.~Wen, Y.~Kuai, J.~Xiao, and F.~Porikli, ``{Learning target-aware
  correlation filters for visual tracking},'' \emph{J. VIS. COMMUN. IMAGE R.},
  vol.~58, pp. 149--159, 2019.

\bibitem{MAM}
B.~Chen, P.~Li, C.~Sun, D.~Wang, G.~Yang, and H.~Lu, ``{Multi attention module
  for visual tracking},'' \emph{Pattern Recognit.}, vol.~87, pp. 80--93, 2019.

\bibitem{ADNet-CVPR}
S.~Yun, J.~J.~Y. Choi, Y.~Yoo, K.~Yun, and J.~J.~Y. Choi, ``{Action-decision
  networks for visual tracking with deep reinforcement learning},'' in
  \emph{Proc. IEEE CVPR}, 2016, pp. 2--6.

\bibitem{ADNet-TNNLS}
S.~Yun, J.~Choi, Y.~Yoo, K.~Yun, and J.~Y. Choi, ``{Action-driven visual object
  tracking with deep reinforcement learning},'' \emph{IEEE Trans. Neural Netw.
  Learn. Syst.}, vol.~29, no.~6, pp. 2239--2252, 2018.

\bibitem{C2FT}
W.~Zhang, K.~Song, X.~Rong, and Y.~Li, ``{Coarse-to-fine UAV target tracking
  with deep reinforcement learning},'' \emph{IEEE Trans. Autom. Sci. Eng.}, pp.
  1--9, 2018.

\bibitem{DRL-IS}
L.~Ren, X.~Yuan, J.~Lu, M.~Yang, and J.~Zhou, ``{Deep reinforcement learning
  with iterative shift for visual tracking},'' in \emph{Proc. ECCV}, 2018, pp.
  697--713.

\bibitem{DRLT}
\BIBentryALTinterwordspacing
D.~Zhang, H.~Maei, X.~Wang, and Y.-F. Wang, ``Deep reinforcement learning for
  visual object tracking in videos,'' 2017. [Online]. Available:
  \url{http://arxiv.org/abs/1701.08936}
\BIBentrySTDinterwordspacing

\bibitem{EAST}
C.~Huang, S.~Lucey, and D.~Ramanan, ``{Learning policies for adaptive tracking
  with deep feature cascades},'' in \emph{Proc. IEEE ICCV}, 2017, pp. 105--114.

\bibitem{HP}
X.~Dong, J.~Shen, W.~Wang, Y.~Liu, L.~Shao, and F.~Porikli, ``{Hyperparameter
  optimization for tracking with continuous deep Q-learning},'' in \emph{Proc.
  IEEE CVPR}, 2018, pp. 518--527.

\bibitem{P-Track}
J.~Supancic and D.~Ramanan, ``{Tracking as online decision-making: Learning a
  policy from streaming videos with reinforcement learning},'' in \emph{Proc.
  IEEE ICCV}, 2017, pp. 322--331.

\bibitem{RDT}
J.~Choi, J.~Kwon, and K.~M. Lee, ``{Real-time visual tracking by deep
  reinforced decision making},'' \emph{Comput. Vis. Image Und.}, vol. 171, pp.
  10--19, 2018.

\bibitem{SINT++}
X.~Wang, C.~Li, B.~Luo, and J.~Tang, ``{SINT++: Robust visual tracking via
  adversarial positive instance generation},'' in \emph{Proc. IEEE CVPR}, 2018,
  pp. 4864--4873.

\bibitem{Meta-Tracker}
E.~Park and A.~C. Berg, ``Meta-tracker: Fast and robust online adaptation for
  visual object trackers,'' in \emph{Proc. ECCV}, 2018.

\bibitem{CRVFL}
L.~Zhang and P.~N. Suganthan, ``Visual tracking with convolutional random
  vector functional link network,'' \emph{IEEE Trans. Cybernetics}, vol.~47,
  no.~10, pp. 3243--3253, 2017.

\bibitem{VTCNN}
L.~Zhang and N.~Suganthan, ``Visual tracking with convolutional neural
  network,'' in \emph{Proc. IEEE Int. Conf. Syst. Man Cybern.}, 2015.

\bibitem{BGBDT}
L.~Huang, X.~Zhao, and K.~Huang, ``Bridging the gap between detection and
  tracking: A unified approach,'' in \emph{Proc. IEEE ICCV}, 2019.

\bibitem{GFS-DCF}
T.~Xu, Z.-H. Feng, X.-J. Wu, and J.~Kittler, ``Joint group feature selection
  and discriminative filter learning for robust visual object tracking,'' in
  \emph{Proc. IEEE ICCV}, 2019.

\bibitem{GradNet}
P.~Li, B.~Chen, W.~Ouyang, D.~Wang, X.~Yang, and H.~Lu, ``Gradnet:
  Gradient-guided network for visual object tracking,'' in \emph{Proc. IEEE
  ICCV}, 2019.

\bibitem{MLT}
J.~Choi, J.~Kwon, and K.~M. Lee, ``Deep meta learning for real-time
  target-aware visual tracking,'' in \emph{Proc. IEEE ICCV}, 2019.

\bibitem{UpdateNet}
L.~Zhang, A.~Gonzalez-Garcia, J.~v.~d. Weijer, M.~Danelljan, and F.~S. Khan,
  ``Learning the model update for siamese trackers,'' in \emph{Proc. IEEE
  ICCV}, 2019.

\bibitem{CGACD}
F.~Du, P.~Liu, W.~Zhao, and X.~Tang, ``Correlation-guided attention for corner
  detection based visual tracking,'' in \emph{Proc. IEEE CVPR}, 2020.

\bibitem{CSA}
B.~Yan, D.~Wang, H.~Lu, and X.~Yang, ``Cooling-shrinking attack: Blinding the
  tracker with imperceptible noises,'' in \emph{Proc. IEEE CVPR}, 2020.

\bibitem{D3S}
A.~Lukezic, J.~Matas, and M.~Kristan, ``D3s - a discriminative single shot
  segmentation tracker,'' in \emph{Proc. IEEE CVPR}, 2020.

\bibitem{OSAA}
X.~Chen, X.~Yan, F.~Zheng, Y.~Jiang, S.-T. Xia, Y.~Zhao, and R.~Ji, ``One-shot
  adversarial attacks on visual tracking with dual attention,'' in \emph{Proc.
  IEEE CVPR}, 2020.

\bibitem{PrDiMP}
M.~Danelljan, L.~V. Gool, and R.~Timofte, ``Probabilistic regression for visual
  tracking,'' in \emph{Proc. IEEE CVPR}, 2020.

\bibitem{RLS}
J.~Gao, W.~Hu, and Y.~Lu, ``Recursive least-squares estimator-aided online
  learning for visual tracking,'' in \emph{Proc. IEEE CVPR}, 2020.

\bibitem{ROAM}
T.~Yang, P.~Xu, R.~Hu, H.~Chai, and A.~B. Chan, ``Roam: Recurrently optimizing
  tracking model,'' in \emph{Proc. IEEE CVPR}, 2020.

\bibitem{SiamAttn}
Y.~Yu, Y.~Xiong, W.~Huang, and M.~R. Scott, ``Deformable siamese attention
  networks for visual object tracking,'' in \emph{Proc. IEEE CVPR}, 2020.

\bibitem{SiamBAN}
Z.~Chen, B.~Zhong, G.~Li, S.~Zhang, and R.~Ji, ``Siamese box adaptive network
  for visual tracking,'' in \emph{Proc. IEEE CVPR}, 2020.

\bibitem{SiamCAR}
D.~Guo, J.~Wang, Y.~Cui, Z.~Wang, and S.~Chen, ``Siamcar: Siamese fully
  convolutional classification and regression for visual tracking,'' in
  \emph{Proc. IEEE CVPR}, 2020.

\bibitem{SiamRCNN}
P.~Voigtlaender, J.~Luiten, P.~H. Torr, and B.~Leibe, ``Siam r-cnn: Visual
  tracking by re-detection,'' in \emph{Proc. IEEE CVPR}, 2020.

\bibitem{TMAML}
G.~Wang, C.~Luo, X.~Sun, Z.~Xiong, and W.~Zeng, ``Tracking by instance
  detection: A meta-learning approach,'' in \emph{Proc. IEEE CVPR}, 2020.

\bibitem{FGTrack}
K.~Shuang, Y.~Huang, Y.~Sun, Z.~Cai, and H.~Guo, ``Fine-grained motion
  representation for template-free visual tracking,'' in \emph{Proc. IEEE
  WACV}, 2020.

\bibitem{DHT}
H.~Song, D.~Suehiro, and S.~Uchida, ``Adaptive aggregation of arbitrary online
  trackers with a regret bound,'' in \emph{Proc. IEEE WACV}, 2020.

\bibitem{MLCFT}
Y.~Ma, C.~Yuan, P.~Gao, and F.~Wang, ``Efficient multi-level correlating for
  visual tracking,'' in \emph{Proc. ACCV}, 2018.

\bibitem{DSNet}
Q.~WuYan, Y.~LiangYi, and L.~Wang, ``Dsnet: Deep and shallow feature learning
  for efficient visual tracking,'' in \emph{Proc. ACCV}, 2018.

\bibitem{BEVT}
C.~Fu, Z.~Huang, Y.~Li, R.~Duan, and P.~Lu, ``Boundary effect-aware visual
  tracking for {UAV} with online enhanced background learning and multi-frame
  consensus verification,'' in \emph{Proc. IROS}, 2019.

\bibitem{CRAC}
W.~Song, S.~Li, T.~Chang, A.~Hao, Q.~Zhao, and H.~Qin, ``Cross-view contextual
  relation transferred network for unsupervised vehicle tracking in drone
  videos,'' in \emph{Proc. IEEE WACV}, 2020.

\bibitem{KAOT_ICRA}
Y.~Li, C.~Fu, Z.~Huang, Y.~Zhang, and J.~Pan, ``Keyfilter-aware real-time uav
  object tracking,'' in \emph{Proc. ICRA}, 2020.

\bibitem{KAOT_TMM}
Y.~Li, C.~Fu, Z.~Huang, and et~al., ``Intermittent contextual learning for
  keyfilter-aware {UAV} object tracking using deep convolutional feature,''
  \emph{IEEE Trans. Multimedia}, 2020.

\bibitem{MKCT}
C.~Fu, Y.~He, F.~Lin, and W.~Xiong, ``Robust multi-kernelized correlators for
  {UAV} tracking with adaptive context analysis and dynamic weighted filters,''
  \emph{Neural Computing and Applications}, vol.~32, 2020.

\bibitem{SASR}
C.~Fu, W.~Xiong, F.~Lin, and Y.~Yue, ``{Surrounding-aware correlation filter
  for UAV tracking with selective spatial regularization},'' \emph{Signal
  Processing}, vol. 167, 2020.

\bibitem{COMET}
S.~M. Marvasti-Zadeh, J.~Khaghani, H.~Ghanei-Yakhdan, S.~Kasaei, and L.~Cheng,
  ``{COMET: Context-aware IoU-guided network for small object tracking},'' in
  \emph{Proc. ACCV}, 2020.

\bibitem{FGLT}
H.~Wu, X.~Yang, Y.~Yang, and G.~Liu, ``Flow guided short-term trackers with
  cascade detection for long-term tracking,'' in \emph{Proc. IEEE. ICCVW},
  2019.

\bibitem{GlobalTrack}
L.~Huang, X.~Zhao, and K.~Huang, ``Globaltrack: A simple and strong baseline
  for long-term tracking,'' in \emph{Proc. AAAI}, 2020.

\bibitem{i-Siam}
W.~Ren~Tan and S.-H. Lai, ``i-siam: Improving siamese tracker with distractors
  suppression and long-term strategies,'' in \emph{Proc. IEEE. ICCVW}, 2019.

\bibitem{LRVN}
\BIBentryALTinterwordspacing
Y.~Zhang, D.~Wang, L.~Wang, J.~Qi, and H.~Lu, ``Learning regression and
  verification networks for long-term visual tracking,'' 2018. [Online].
  Available: \url{http://arxiv.org/abs/1809.04320}
\BIBentrySTDinterwordspacing

\bibitem{MetaUpdater}
K.~Dai, Y.~Zhang, D.~Wang, J.~Li, H.~Lu, and X.~Yang, ``High-performance
  long-term tracking with meta-updater,'' in \emph{Proc. IEEE CVPR}, 2020.

\bibitem{SPLT}
B.~Yan, H.~Zhao, D.~Wang, H.~Lu, and X.~Yang, ``'skimming-perusal' tracking: A
  framework for real-time and robust long-term tracking,'' in \emph{Proc. IEEE
  ICCV}, 2019.

\bibitem{OTB2013}
Y.~Wu, J.~Lim, and M.~H. Yang, ``{Online object tracking: A benchmark},'' in
  \emph{Proc. IEEE CVPR}, 2013, pp. 2411--2418.

\bibitem{OTB2015}
Y.~Wu, J.~Lim, and M.~Yang, ``{Object tracking benchmark},'' \emph{IEEE Trans.
  Pattern Anal. Mach. Intell.}, vol.~37, no.~9, pp. 1834--1848, 2015.

\bibitem{LaSOT}
\BIBentryALTinterwordspacing
H.~Fan, L.~Lin, F.~Yang, P.~Chu, G.~Deng, S.~Yu, H.~Bai, Y.~Xu, C.~Liao, and
  H.~Ling, ``{LaSOT}: A high-quality benchmark for large-scale single object
  tracking,'' 2018. [Online]. Available: \url{http://arxiv.org/abs/1809.07845}
\BIBentrySTDinterwordspacing

\bibitem{UAV123}
M.~Mueller, N.~Smith, and B.~Ghanem, ``{A benchmark and simulator for UAV
  tracking},'' in \emph{Proc. ECCV}, 2016, pp. 445--461.

\bibitem{UAVDT2018}
D.~Du, Y.~Qi, H.~Yu, Y.~Yang, K.~Duan, G.~Li, W.~Zhang, Q.~Huang, and Q.~Tian,
  ``The unmanned aerial vehicle benchmark: Object detection and tracking,'' in
  \emph{Proc. ECCV}, 2018, pp. 375--391.

\bibitem{VisDrone2019}
D.~Du, P.~Zhu, L.~Wen, X.~Bian, H.~Ling, and et~al., ``{VisDrone-SOT2019: The
  Vision Meets Drone Single Object Tracking Challenge Results},'' in
  \emph{Proc. ICCVW}, 2019.

\bibitem{TC128}
P.~Liang, E.~Blasch, and H.~Ling, ``{Encoding color information for visual
  tracking: Algorithms and benchmark},'' \emph{IEEE Trans. Image Process.},
  vol.~24, no.~12, pp. 5630--5644, 2015.

\bibitem{NUS-PRO}
A.~Li, M.~Lin, Y.~Wu, M.~H. Yang, and S.~Yan, ``{NUS-PRO: A new visual tracking
  challenge},'' \emph{IEEE Trans. Pattern Anal. Mach. Intell.}, vol.~38, no.~2,
  pp. 335--349, 2016.

\bibitem{NfS}
H.~K. Galoogahi, A.~Fagg, C.~Huang, D.~Ramanan, and S.~Lucey, ``{Need for
  speed: A benchmark for higher frame rate object tracking},'' in \emph{Proc.
  IEEE ICCV}, 2017, pp. 1134--1143.

\bibitem{DTB}
S.~Li and D.~Y. Yeung, ``{Visual object tracking for unmanned aerial vehicles:
  A benchmark and new motion models},'' in \emph{Proc. AAAI}, 2017, pp.
  4140--4146.

\bibitem{TrackingNet}
M.~M{\"{u}}ller, A.~Bibi, S.~Giancola, S.~Alsubaihi, and B.~Ghanem,
  ``{TrackingNet: A large-scale dataset and benchmark for object tracking in
  the wild},'' in \emph{Proc. ECCV}, 2018, pp. 310--327.

\bibitem{OxUvA}
J.~Valmadre, L.~Bertinetto, J.~F. Henriques, R.~Tao, A.~Vedaldi, A.~W.
  Smeulders, P.~H. Torr, and E.~Gavves, ``{Long-term tracking in the wild: A
  benchmark},'' in \emph{Proc. ECCV}, vol. 11207 LNCS, 2018, pp. 692--707.

\bibitem{BUAA-PRO}
\BIBentryALTinterwordspacing
A.~Li, Z.~Chen, and Y.~Wang, ``{BUAA-PRO: A tracking dataset with pixel-level
  annotation},'' in \emph{Proc. BMVC}, 2018. [Online]. Available:
  \url{http://bmvc2018.org/contents/papers/0851.pdf}
\BIBentrySTDinterwordspacing

\bibitem{GOT-10k}
\BIBentryALTinterwordspacing
L.~Huang, X.~Zhao, and K.~Huang, ``{GOT-10k}: A large high-diversity benchmark
  for generic object tracking in the wild,'' 2018. [Online]. Available:
  \url{http://arxiv.org/abs/1810.11981}
\BIBentrySTDinterwordspacing

\bibitem{TLP}
A.~Moudgil and V.~Gandhi, ``Long-term visual object tracking benchmark,'' in
  \emph{Proc. ICCV}, 2018, pp. 629--645.

\bibitem{TracKlinic}
H.~Fan, Y.~Fan, P.~Chu, L.~Yuan, and H.~Ling, ``Tracklinic: Diagnosis of
  challenge factors in visual tracking,'' ser. Proc. WACV, 2021.

\bibitem{Long-term_NYSM}
\BIBentryALTinterwordspacing
A.~Lukezic, L.~C. Zajc, T.~Vojir, J.~Matas, and M.~Kristan, ``Now you see me:
  evaluating performance in long-term visual tracking,'' 2018. [Online].
  Available: \url{http://arxiv.org/abs/1804.07056}
\BIBentrySTDinterwordspacing

\bibitem{VisDrone2018}
P.~Zhu, L.~Wen, D.~Du, and et~al., ``{VisDrone-VDT2018: The vision meets drone
  video detection and tracking challenge results},'' in \emph{Proc. ECCVW},
  2018, pp. 496--518.

\bibitem{Small90Dataset}
C.~Liu, W.~Ding, J.~Yang, and et~al., ``Aggregation signature for small object
  tracking,'' \emph{IEEE Trans. Image Processing}, vol.~29, pp. 1738--1747,
  2020.

\bibitem{SSD}
W.~Liu, D.~Anguelov, D.~Erhan, C.~Szegedy, S.~Reed, C.~Y. Fu, and A.~C. Berg,
  ``{SSD: Single shot multibox detector},'' in \emph{Proc. ECCV}, 2016, pp.
  21--37.

\bibitem{Koch2015}
G.~Koch, R.~Zemel, and R.~Salakhutdinov, ``{Siamese neural networks for
  one-shot image recognition},'' in \emph{Proc. ICML Deep Learning Workshop},
  2015.

\bibitem{RefineNet}
G.~Lin, A.~Milan, C.~Shen, and I.~Reid, ``{RefineNet: Multi-path refinement
  networks for high-resolution semantic segmentation},'' in \emph{Proc. IEEE
  CVPR}, 2017, pp. 5168--5177.

\bibitem{Lin2017}
T.~Y. Lin, P.~Doll{\'{a}}r, R.~Girshick, K.~He, B.~Hariharan, and S.~Belongie,
  ``{Feature pyramid networks for object detection},'' in \emph{Proc. IEEE
  CVPR}, 2017, pp. 936--944.

\bibitem{DeepMotionFeatures}
S.~Gladh, M.~Danelljan, F.~S. Khan, and M.~Felsberg, ``{Deep motion features
  for visual tracking},'' in \emph{Proc. ICPR}, 2016, pp. 1243--1248.

\bibitem{YT-VOS}
N.~Xu, L.~Yang, Y.~Fan, D.~Yue, Y.~Liang, J.~Yang, and T.~Huang,
  ``{YouTube-VOS: A} large-scale video object segmentation benchmark,'' in
  \emph{Proc. ECCV}, 2018.

\bibitem{YouTube-BB}
E.~Real, J.~Shlens, S.~Mazzocchi, X.~Pan, and V.~Vanhoucke,
  ``{YouTube-BoundingBoxes: A large high-precision human-annotated data set for
  object detection in video},'' in \emph{Proc. IEEE CVPR}, 2017, pp.
  7464--7473.

\bibitem{KITTI}
A.~Geiger, P.~Lenz, C.~Stiller, and R.~Urtasun, ``Vision meets robotics: {T}he
  {KITTI} dataset,'' vol.~32, no.~11, p. 1231–1237, 2013.

\bibitem{Meta-Survey}
\BIBentryALTinterwordspacing
T.~Hospedales, A.~Antoniou, P.~Micaelli, and A.~Storkey, ``Meta-learning in
  neural networks: {A} survey,'' 2020. [Online]. Available:
  \url{http://arxiv.org/abs/2004.05439}
\BIBentrySTDinterwordspacing

\bibitem{MAML}
C.~Finn, P.~Abbeel, and S.~Levine, ``Model-agnostic meta-learning for fast
  adaptation of deep networks,'' in \emph{Proc. ICML}, 2017, pp. 1126--1135.

\bibitem{SegmentTracker_JOTS}
L.~Wen, D.~Du, Z.~Lei, S.~Z. Li, and M.-H. Yang, ``Jots: Joint online tracking
  and segmentation,'' in \emph{Proc. IEEE CVPR}, 2015.

\bibitem{SegmentTracker_DecisionTree}
J.~Son, I.~Jung, K.~Park, and B.~Han, ``Tracking-by-segmentation with online
  gradient boosting decision tree,'' in \emph{Proc. IEEE ICCV}, 2015.

\bibitem{SegmentTracker_SuperPixel}
D.~Yeo, J.~Son, B.~Han, and J.~Hee~Han, ``Superpixel-based
  tracking-by-segmentation using markov chains,'' in \emph{Proc. IEEE CVPR},
  2017.

\bibitem{Box2Seg1}
J.~Luiten, P.~Voigtlaender, and B.~Leibe, ``Premvos: Proposal-generation,
  refinement and merging for video object segmentation,'' in \emph{Proc. ACCV},
  2018.

\bibitem{Box2Seg2}
V.~Kulharia, S.~Chandra, A.~Agrawal, P.~Torr, and A.~Tyagi, ``Box2seg:
  Attention weighted loss and discriminative feature learning for weakly
  supervised segmentation,'' in \emph{Proc. ECCV}, 2020.

\bibitem{TraX}
L.~{\v{C}}ehovin, ``{TraX: The visual tracking exchange protocol and
  library},'' \emph{Neurocomputing}, vol. 260, pp. 5--8, 2017.

\bibitem{MatConvNet}
A.~Vedaldi and K.~Lenc, ``{MatConvNet: Convolutional neural networks for
  MATLAB},'' in \emph{Proc. ACM Multimedia Conference}, 2015, pp. 689--692.

\bibitem{IoUNet}
B.~Jiang, R.~Luo, J.~Mao, T.~Xiao, and Y.~Jiang, ``Acquisition of localization
  confidence for accurate object detection,'' in \emph{Proc. IEEE ECCV}, 2018,
  pp. 816--832.

\bibitem{Forget_HardAttn}
J.~Serra, D.~Suris, M.~Miron, and A.~Karatzoglou, ``Overcoming catastrophic
  forgetting with hard attention to the task,'' ser. Proc. ICML, 2018, pp.
  4548--4557.

\bibitem{Forget_TransferPenalty}
X.~LI, Y.~Grandvalet, and F.~Davoine, ``Explicit inductive bias for transfer
  learning with convolutional networks,'' ser. Proc. ICML, 2018, pp.
  2825--2834.

\bibitem{Forget_BSS}
X.~Chen, S.~Wang, B.~Fu, M.~Long, and J.~Wang, ``Catastrophic forgetting meets
  negative transfer: Batch spectral shrinkage for safe transfer learning,'' in
  \emph{Proc. NIPS}, 2019, pp. 1908--1918.

\bibitem{Forget_lifelong}
J.~Yoon, E.~Yang, J.~Lee, and S.~J. Hwang, ``Lifelong learning with dynamically
  expandable networks,'' in \emph{Proc. ICLR}, 2018.

\end{thebibliography}

\begin{IEEEbiography}[{\includegraphics[width=1in,height=1.25in,clip,keepaspectratio]{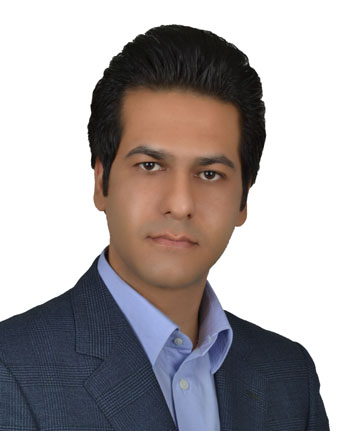}}]{Seyed Mojtaba Marvasti-Zadeh} (Student Member, IEEE) is currently a Ph.D. student in Electrical Engineering, Yazd University (YU), Iran. He was awarded as a distinguished researcher student of the Department of Electrical Engineering, YU, in 2015 and 2017. He is also a member of Vision and Learning Lab, University of Alberta, Canada, where he was a visiting researcher from Dec.~2019 to Sep.~2020. His research interest includes computer vision and machine learning.
\end{IEEEbiography}

\begin{IEEEbiography}[{\includegraphics[width=1in,height=1.25in,clip,keepaspectratio]{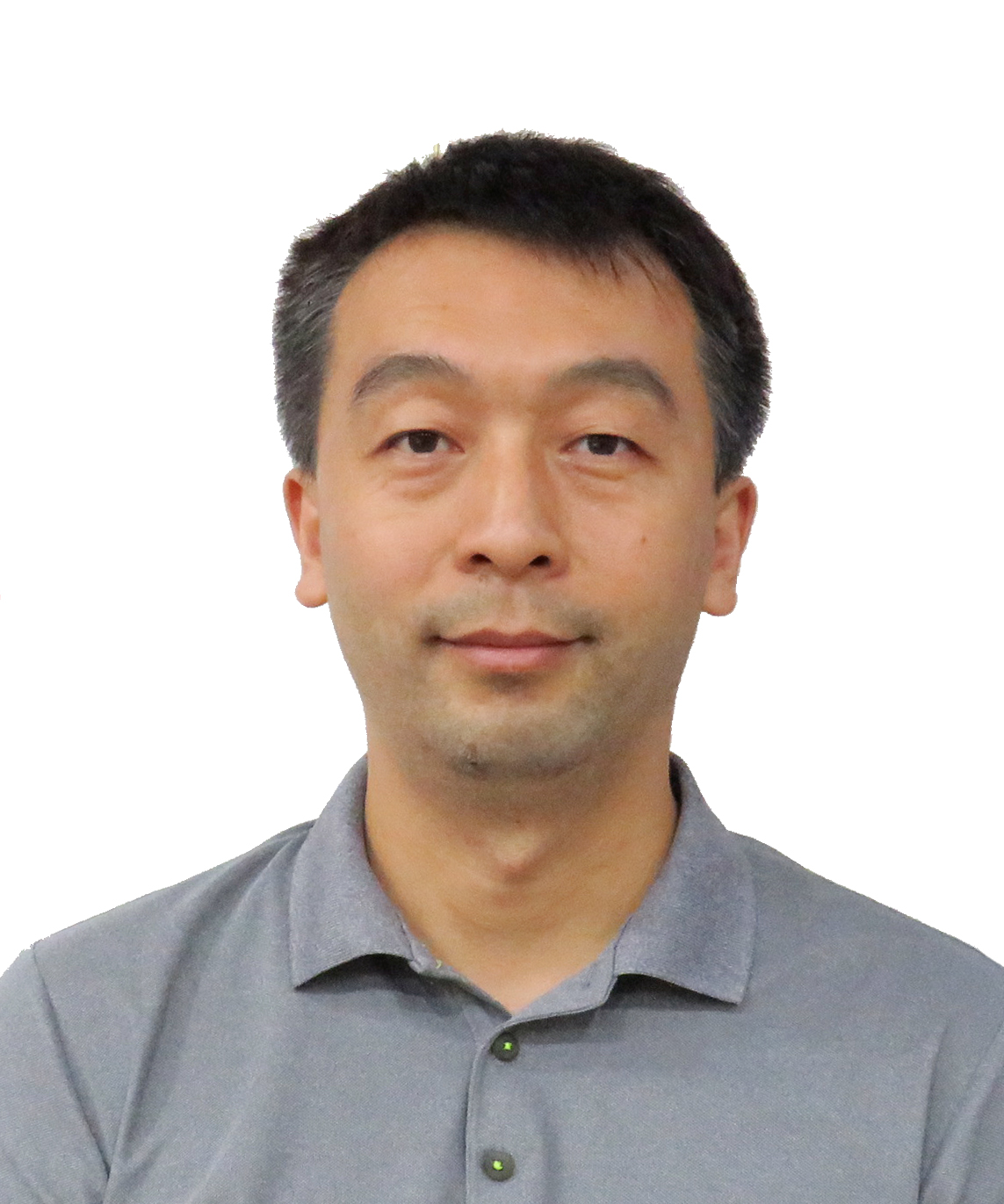}}]{Li Cheng} (Senior Member, IEEE) received the Ph.D. degree in computer science from the University of Alberta, Canada. He is an associate professor with the Department of Electrical and Computer Engineering, University of Alberta, Canada. He is also the Director of the Vision and Learning Lab. Prior to coming back to University of Alberta, he has worked at A*STAR, Singapore, TTIChicago, USA, and NICTA, Australia. His research expertise is mainly on computer vision and machine learning.
\end{IEEEbiography}

\begin{IEEEbiography}[{\includegraphics[width=1in,height=1.25in,clip,keepaspectratio]{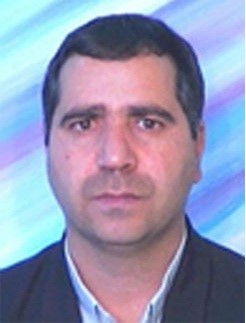}}]{Hossein Ghanei-Yakhdan} received the B.Sc. degree in Electrical Engineering from Isfahan University of Technology, Iran, in 1989, the M.Sc. degree in 1993 from K. N. Toosi University of Technology, Iran, and the Ph.D. degree in 2009 from Ferdowsi University of Mashhad, Iran. Since 2009, he has been an Assistant Professor with the Department of Electrical Engineering, Yazd University, Iran. His research interests are in digital video and image processing, error concealment and error-resilient video coding, and object tracking.
\end{IEEEbiography}

\begin{IEEEbiography}[{\includegraphics[width=1in,height=1.25in,clip,keepaspectratio]{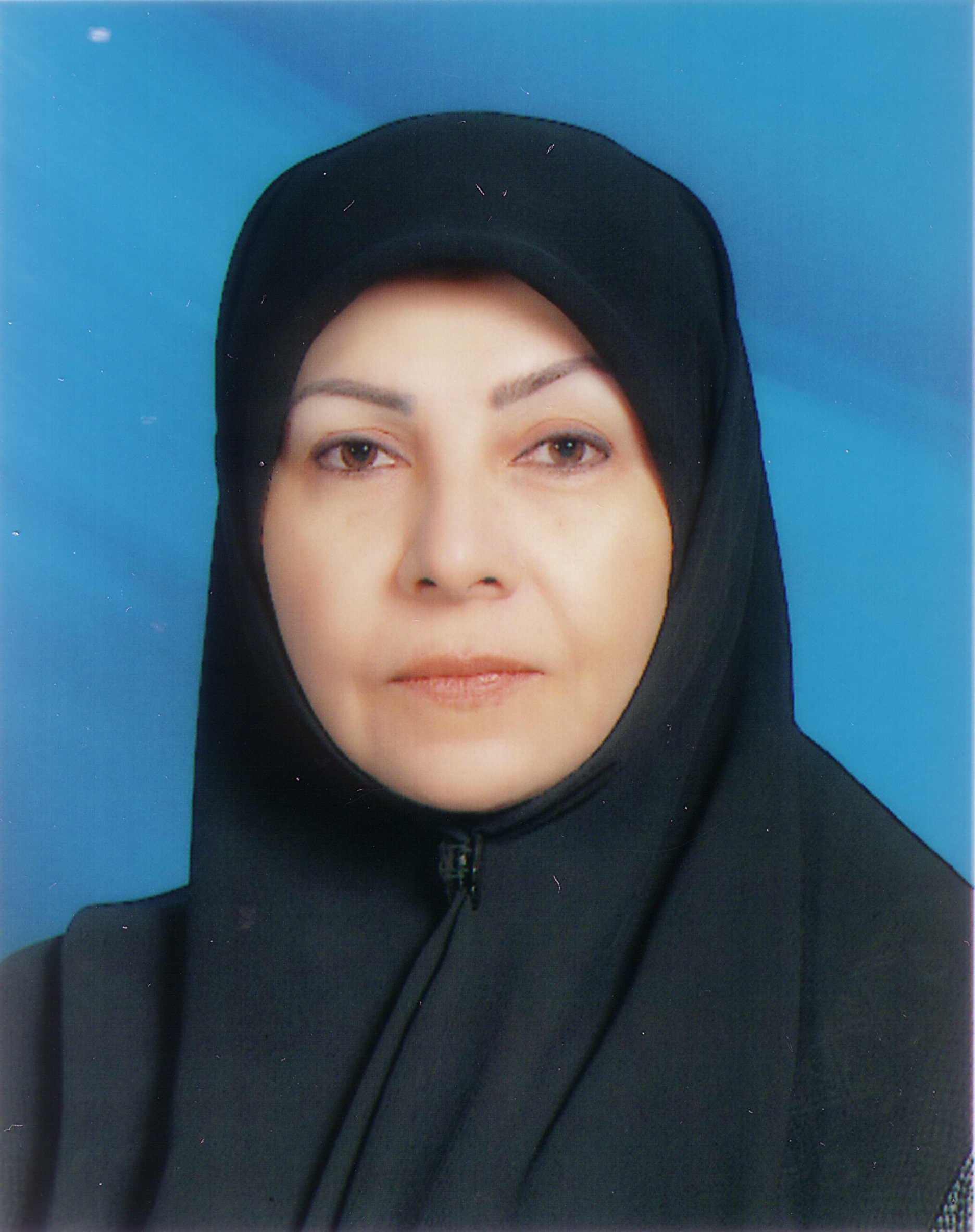}}]{Shohreh Kasaei} (Senior Member, IEEE) received the Ph.D. degree from the Signal Processing Research Center, School of Electrical Engineering and Computer Science, Queensland University of Technology, Australia, in 1998. She was awarded as a distinguished researcher of Sharif University of Technology (SUT), in 2002 and 2010, where she is currently a Full Professor. Since 1999, she has been the Director of the Image Processing Laboratory (IPL). Her research interests include machine learning, computer vision, and image/video processing.
\end{IEEEbiography}

\end{document}